\documentstyle[11pt,jair,twoside,url,theapa,epic]{article}
\jairheading{10}{1999}{117-167}{2/98}{3/99}
\ShortHeadings{Modeling Belief in Dynamic Systems. Part II.}{Friedman \& Halpern}
\firstpageno{117}

\long\def\ifundefined#1#2#3{\expandafter\ifx\csname#1\endcsname\relax
#2\else#3\fi}

\ifundefined{DEFNSTYPRESENT}{\def\thmcolon{\hspace{-.85em} {\bf
:} }}{\def\thmcolon{\relax}} 

\newtheorem{THEOREM}{Theorem}[section]
\newenvironment{theorem}{\begin{THEOREM} \thmcolon }%
                        {\end{THEOREM}}
\newtheorem{LEMMA}[THEOREM]{Lemma}
\newenvironment{lemma}{\begin{LEMMA} \thmcolon }%
                      {\end{LEMMA}}
\newtheorem{COROLLARY}[THEOREM]{Corollary}
\newenvironment{corollary}{\begin{COROLLARY} \thmcolon }%
                          {\end{COROLLARY}}
\newtheorem{PROPOSITION}[THEOREM]{Proposition}
\newenvironment{proposition}{\begin{PROPOSITION} \thmcolon }%
                            {\end{PROPOSITION}}
\newtheorem{DEFINITION}[THEOREM]{Definition}
\newenvironment{definition}{\begin{DEFINITION} \thmcolon \rm}%
                            {\end{DEFINITION}}
\newtheorem{CLAIM}[THEOREM]{Claim}
                            {\end{CLAIM}}
\newtheorem{EXAMPLE}[THEOREM]{Example}
\newenvironment{example}{\begin{EXAMPLE} \thmcolon \rm}%
                            {\end{EXAMPLE}}
\newtheorem{REMARK}[THEOREM]{Remark}
\newenvironment{remark}{\begin{REMARK} \thmcolon \rm}%
                            {\end{REMARK}}

\newcommand{\thm}{\begin{theorem}}
\newcommand{\lem}{\begin{lemma}}
\newcommand{\pro}{\begin{proposition}}
\newcommand{\dfn}{\begin{definition}}
\newcommand{\rem}{\begin{remark}}
\newcommand{\xam}{\begin{example}}
\newcommand{\cor}{\begin{corollary}}
\newcommand{\prf}{\noindent{\bf Proof:} }
\newcommand{\ethm}{\end{theorem}}
\newcommand{\elem}{\end{lemma}}
\newcommand{\epro}{\end{proposition}}
\newcommand{\edfn}{\wbox\end{definition}}
\newcommand{\erem}{\wbox\end{remark}}
\newcommand{\exam}{\wbox\end{example}}
\newcommand{\ecor}{\end{corollary}}
\newcommand{\eprf}{\wbox\vspace{0.1in}}
\newcommand{\beqn}{\begin{equation}}
\newcommand{\eeqn}{\end{equation}}
\newcommand{\wbox}{\mbox{$\sqcap$\llap{$\sqcup$}}}

\newcommand{\sat}{\models}

\newcommand{\rimp}{\Rightarrow}

\newcommand{\dimp}{\Leftrightarrow}

\newcommand{\band}{\bigwedge}
\newcommand{\union}{\cup}
\newcommand{\inter}{\cap}

\newcommand{\natnum}{{\sl N}}

\renewcommand{\phi}{\varphi}
\newcommand{\Circ}{\mbox{{\small $\bigcirc$}}}

\newcommand{\cross}{\times}

\newcommand{\C}{{\cal C}}

\newcommand{\E}{{\cal E}}
\newcommand{\F}{{\cal F}}

\newcommand{\I}{{\cal I}}

\renewcommand{\P}{{\cal P}}

\newcommand{\R}{{\cal R}}

\newcommand{\<}{\langle}
\renewcommand{\>}{\rangle}

\newcommand{\eg}{e.g.,~}
\newcommand{\ie}{i.e.,~}

\newcommand{\cf}{cf.~}

\newcommand{\respc}{resp.,\ }

\newcommand{\ol}{\setlength{\itemsep}{0pt}\begin{enumerate}}
\newcommand{\eol}{\end{enumerate}\setlength{\itemsep}{-\parsep}}
\newcommand{\ul}{\setlength{\itemsep}{0pt}\begin{itemize}}
\newcommand{\dl}{\setlength{\itemsep}{0pt}\begin{description}}
\newcommand{\edl}{\end{description}\setlength{\itemsep}{-\parsep}}
\newcommand{\eul}{\end{itemize}\setlength{\itemsep}{-\parsep}}

\newcommand{\true}{\mbox{{\it true}}}
\newcommand{\false}{\mbox{{\it false}}}

\newcommand{\commentout}[1]{}

\newcommand{\bi}{\begin{itemize}}
\newcommand{\ei}{\end{itemize}}
\newcommand{\be}{\begin{enumerate}}
\newcommand{\ee}{\end{enumerate}}

\newcommand{\denselist}{\itemsep 0pt\partopsep 0pt}

\renewcommand{\L}{{\cal L}}
\renewcommand{\S}{{\cal S}}

\newcommand{\Sys}{\I}

\newcommand{\Next}{\Circ}

\newcommand{\Cond}{\mbox{\boldmath$\rightarrow$\unboldmath}}
\newcommand{\RCond}{>}

\newcommand{\Know}{K}
\newcommand{\Bel}{B}

\newcommand{\True}{\mbox{\it true}}
\newcommand{\False}{\mbox{\it false}}

\newcommand{\intension}[1]{[\![ #1 ]\!]}

\newcommand{\Pl}{\mbox{\rm Pl\/}}
\newcommand{\PL}{\mbox{\it PL}}

\newcommand{\bottom}{\perp}
\newcommand{\sPL}{{\mbox{\scriptsize\it PL}}}

\newcommand{\LCond}{\L^{C}}
\newcommand{\SysP}{system~{\bf P}}

\newcommand{\Tref}[1]{Theorem~\ref{#1}}

\newcommand{\Cref}[1]{Corollary~\ref{#1}}

\newcommand{\Xref}[1]{Example~\ref{#1}}

\newcommand{\Sref}[1]{Section~\ref{#1}}
\newcommand{\BEL}{\mbox{Bel}}

\newcommand{\Card}[1]{\left| #1\right|}

\newenvironment{RETHM}[2]{\it \trivlist \item[\hskip \labelsep{\bf #1 \ref{#2}:}]}{\endtrivlist}
\newcommand{\rethm}[1]{\begin{RETHM}{Theorem}{#1}}
\newcommand{\repro}[1]{\begin{RETHM}{Proposition}{#1}}
\newcommand{\relem}[1]{\begin{RETHM}{Lemma}{#1}}
\newcommand{\recor}[1]{\begin{RETHM}{Corollary}{#1}}

\newcommand{\erethm}{\end{RETHM}}
\newcommand{\erepro}{\end{RETHM}}
\newcommand{\erelem}{\end{RETHM}}
\newcommand{\erecor}{\end{RETHM}}

\renewcommand{\Omega}{W}

\newcommand{\States}{{\mbox{\it States\/}}}

\newcommand{\Plass}{\P}

\renewcommand{\Omega}{W}

\newcommand{\Diag}[1]{D_{#1}}

\newcommand{\Rev}{\circ}
\newcommand{\Upd}{\diamond}

\newcommand{\learn}{\mbox{{\it learn}}}

\newcommand{\obs}[1]{o_{(#1)}}

\newcommand{\io}{\mbox{{\it io}}}
\newcommand{\fault}{\mbox{{\it fault}}}

\newcommand{\LKPT}{\L^{\mbox{\scriptsize{{\it KPT}}}}}

\newcommand{\Phis}{\Phi_e}
\newcommand{\Phil}{\Phi_{obs}}

\newcommand{\Sysclass}{\C}
\newcommand{\SR}{\Sysclass^R}
\newcommand{\SU}{\Sysclass^U}
\newcommand{\SBCS}{\Sysclass^{BCS}}

\newcommand{\Le}{{{\cal L}_e}}

\newcommand{\RP}[1]{[#1]}
\newcommand{\RE}[1]{\R[#1]}

\newcommand{\MapBack}{\mbox{prev}}

\newenvironment{cond}[1]{\begin{quote}{\bf #1} }{\end{quote}}

\newcommand{\diag}{{\mbox{\scriptsize\it diag}}}

\newcommand{\sPl}{{\mbox{\rm\scriptsize Pl}}}

\newcommand{\whiten}{}

\renewcommand{\L}{{\cal L}}
\newcommand{\Ls}{{\cal L}_e}
\renewcommand{\S}{{\cal S}}

\newcommand{\timestamp}{\mbox{\rm timestamp}}

\begin{document}
\title{Modeling Belief in Dynamic Systems\\
Part II: Revision and Update}
\author{%
\name Nir Friedman
\email nir@cs.huji.ac.il\\
\addr  Institute of Computer Science\\
\addr Hebrew University, Jerusalem,
91904, ISRAEL\\
\addr  {\tt http://www.cs.huji.ac.il/}\mbox{$\sim$}{\tt nir}\\
\AND
\name Joseph Y.~Halpern \email halpern@cs.cornell.edu\\
\addr Computer Science Department\\
\addr Cornell University, Ithaca, NY 14853\\
\addr {\tt http://www.cs.cornell.edu/home/halpern}
}
\maketitle
\begin{abstract}
The study of {\em belief change\/} has been an active area in
philosophy and AI. In recent years two special cases of belief change,
{\em belief revision\/} and {\em belief update}, have been studied in
detail. In a companion paper \cite{FrH1Full}, we introduce a new
framework to model belief change. This framework combines temporal and
epistemic modalities with a notion of plausibility, allowing us to
examine the change of beliefs over time.  In this paper, we show how
belief revision and belief update can be captured in our framework.
This allows us to compare the assumptions made by each method, and to
better understand the principles underlying them.
In particular, it
shows that Katsuno and Mendelzon's
notion of belief update \cite{KM91} depends on several strong
assumptions that may limit its applicability in artificial
intelligence.
Finally, our analysis allow us to identify a notion of {\em minimal
change\/} that underlies a broad range of belief change operations
including revision and update.
\end{abstract}

\section{Introduction}
The study of {\em belief change\/} has been an active area in
philosophy and artificial intelligence.  The focus of this
research is to understand how an agent should change her beliefs
as a result of getting new information.
Two instances of this general phenomenon have been studied in detail.
{\em Belief revision\/} \cite{agm:85,Gardenfors1}
focuses on how an agent should change her (set of) beliefs when she
adopts a particular new belief.
{\em Belief update\/} \cite{KM91}, on the other hand,
focuses on how an agent should change her beliefs when she
realizes
that the world has changed.  Both approaches attempt to capture the
intuition that an agent should make minimal changes in her beliefs in
order to accommodate the new belief.  The difference is that belief
revision attempts to decide what beliefs should be discarded to
accommodate a new belief, while belief update attempts to decide
what changes in the world led to the new observation.%
\footnote{Throughout the paper we use ``revision'' to refer to AGM's
proposal for revision \cite{agm:85}
not as a generic term for the general approach initiated by AGM;
similarly, we use ``update'' to refer to
KM's proposal for update \cite{KM91}.}

Belief revision and belief update are two of many possible ways of
modeling belief change.  In \cite{FrH1Full}, we introduce a general
framework for modeling belief change. We start with the framework for
analyzing knowledge in multi-agent systems, introduced in \cite{HF87},
and add to it a measure of plausibility at each situation.  We then
define belief as truth in the most plausible situations.  The
resulting framework is very expressive; it captures both time and
knowledge as well as beliefs.  Having time allows us to reason in the
framework about changes in the beliefs of the agent. It also allows us
to relate the beliefs of the agent about the future with her actual
beliefs in the future.  Knowledge captures in a precise sense the
non-defeasible information the agent has about the world, while belief
captures the defeasible assumptions implied by her plausibility
assessment.  The framework allows us to represent a broad spectrum of
notions of belief change.  In this paper, we focus on how, in
particular, belief revision and update can be represented.

We are certainly not the first to provide semantic models for belief
revision and update.  For example,
\cite{agm:85,Grove,MakGar,Rott91,Boutilier92,derijke:92} deal with
revision, and \cite{KM91,delVal1} deal with update.  In fact, there
are several works in the literature that capture both using the same
machinery \cite{KatsunoSatoh,Goldszmidt+Pearl:1996,Boutilier95Full},
and others that simulate belief revision using belief update
\cite{Grahne2,delVal3}. Our approach is different from most in that we
do not construct a specific framework to capture one or both of these
belief change paradigms.  Instead, we start from a natural framework
to model how an agent's knowledge changes over time and add to it
machinery that captures a defeasible notion of belief.

We believe that our representation offers a number of advantages, and
gives
a deeper understanding of
both revision and update. For one thing, we show
that both revision and update can be viewed as proceeding by {\em
conditioning\/} on initial prior plausibilities.  Thus, our
representation emphasizes the role of conditioning as a way of
understanding {\em minimal change}.  Moreover, it shows that that the
major differences between revision and update can be understood as
corresponding to differences in initial beliefs.  For example,
revision places full belief on the assumption that the propositions
used to describe the world are {\em static}, and do not change their
truth value over time.
By way of contrast, update allows for the possibility that
propositions change their truth value over time.  However, the family
of prior plausibilities that we use to capture update in our
framework have the property that
they prefer sequences of events where abnormal events occur as late as
possible.
Because of this property, conditioning in update always ``explains''
observations by recent changes.
The fact that time appears explicitly in our framework
allows us to make these issues precise.

In the literature, revision has been viewed as dealing with
static worlds (although an agent's beliefs may change, the underlying
world about which the agent is reasoning does not) while update has been
viewed as dealing with dynamic worlds (see, for example, \cite{KM91}).
We believe that the distinction between static and dynamic worlds is
somewhat misleading.
In fact, what is important for
revision is not that the world is static, but that the propositions
used to describe the world are static.  For example, ``At time 0 the
block is on the table'' is a static proposition, while ``The block is
on the table'' is not, since it implicitly references the current
state of affairs.  (Note that the assumption that the propositions are
static is not unique to belief revision. Bayesian updating, for
example, makes similar assumptions.) Because we model time explicitly
in our framework, we can examine this issue in more detail. In fact, in
Section~\ref{sec:synthesis}, we show how we relate these two viewpoints.
More precisely, given a system, we replace each
proposition $p$ used in the system by a family of propositions ``$p$ is
true at time $m$'', one for each time $m$.
The resulting system describes exactly
the same process as the original system,
but from a
different linguistic perspective. As we show, if the original system
corresponds to KM update, then the resulting system is very close to
satisfying the requirements of AGM revision. The only requirement
that is not met is that the prior is totally ordered, or ranked. This
requirement, however, has been relaxed in several variants of revision
\cite{KM92,Rott92}. Thus, a large part of the difference between
revision and update can be understood as a difference
in the language used to describe what is happening.

The generality of our framework forces us to be clear about the
assumptions we make in the process of capturing revision and
update. As a consequence, we have to deal with issues that have been
largely ignored by previous semantic accounts. One of these issues is
the status of observations. As we show below, to capture
either revision or update, we have to assume that observations are
minimally informative---the only information
carried by an observation of
$\phi$ is that $\phi$ should be believed. This is a strong assumption,
since most observations carry additional information.
For example, when trekking in Nepal, one does not expect to observe the
weather in Boston.
If an agent observes that it is in fact raining in Boston, then this
``observation'' might well provide extra information about the world
(for example, that cable television is available in Nepal).
We remark that in
\cite{BFH1} there is a treatment of revision in our framework where
observations are allowed to convey additional information.

Finally, our representation makes it clear how the intuitions of
revision and update can be applied in settings where the postulates used
to describe them are not sound.  For example, we consider situations
where they may be irreversible changes (such as death, or breaking a
glass vase), and where the agent may perform actions beyond just making
observations.
Revision and update, as they stand, cannot handle such situations.
As we show, our framework allows us
to extend them in a natural way so they do.

The rest of the paper is organized as follows. In
Section~\ref{sec:framework}, we give an overview of the framework we
introduced in \cite{FrH1Full}.  In Section~\ref{sec:review}, we give a
brief review of belief revision and belief
update.  In Section~\ref{bcs}, we define a specific class
of structures
that
embody assumptions that are common to both update and
revision.  In Section~\ref{sec:revision}, we describe additional
assumptions that are required to capture revision.  In
Section~\ref{sec:update}, we describe the assumptions that are required
to capture update. In Section~\ref{sec:synthesis}, we reexamine the
differences and similarities between belief revision and update.
In Section~\ref{sec:extensions}, we consider possible extensions to the
setup of revision and update, and discuss how these extensions can be
handled in our framework.
Finally, in
Section~\ref{sec:discussion}, we conclude with a discussion of related
and  future work.

\section{The Framework}\label{sec:framework}

We now review the framework of Halpern and Fagin \citeyear{HF87} for
modeling knowledge,
and our extension of it
for dealing with belief change.
The reader is encouraged to consult \cite{FHMV} for further details and
motivation.

\subsection{Modeling Knowledge}

The framework of Halpern and Fagin was developed to model knowledge in
distributed (\ie multi-agent) systems \cite{HF87,FHMV}. In this paper,
we restrict our attention to the single agent case.
The key assumption in this framework is that we can characterize the
system by describing it in terms of a {\em state\/} that changes over
time.  Formally, we assume that at each point in time, the agent is in
one of a possibly infinite set of (local) states.
At this point, we do not put any further structure on
these states
(although, as we shall see from our examples, when we model situations in
a natural way, states typically do have a great deal of meaningful
structure).
Intuitively, this local state encodes the
information the agent has observed thus far. There is also an {\em
environment}, whose state encodes relevant aspects of the system that
are not part of the agent's local state.

A {\em global state\/} is a pair $(s_e, s_a)$ consisting of the
environment state $s_e$ and the local state $s_a$ of the agent.
A {\em
run\/} of the system is a function from time (which, for ease of
exposition, we assume ranges over the natural numbers) to global
states.
Thus, if $r$ is a run, then $r(0), r(1), \ldots$ is a sequence
of global states that, roughly speaking, is a complete description of
what happens over time in one possible execution of the system.
Given a run $r$, we can define two functions $r_e$ and $r_a$ that map
from time to states of the environment and the agent, respectively, by
taking $r_e(m)$ to be the state of the environment in the global state
$r(m)$ and $r_a(m)$ to be the agent's local state in $r(m)$.  We can
thus identify run $r$ with the pair of functions
$\<r_e,r_a\>$.
We
take a {\em system\/} to consist of a set of runs.  Intuitively, these
runs describe all the possible behaviors of the system, that is, all
the possible sequences of events that could occur in the system over
time.

Given a system $\R$, we refer to a pair $(r,m)$ consisting of a run $r
\in \R$ and a time $m$ as a {\em point}.
We say two points $(r,m)$
and $(r',m')$ are {\em indistinguishable\/} to the agent, and write
$(r,m) \sim_a (r',m')$, if $r_a(m) = r'_a(m')$, \ie if the agent has
the same local state at both points.  Finally, an {\em interpreted\/}
system $\Sys$ is a tuple $(\R,\pi)$
consisting of a system $\R$ together with a mapping $\pi$ that
associates with each point a truth assignment to a set $\Phi$ of primitive
propositions.  In an interpreted system we can talk about an agent's
knowledge: the agent knows $\phi$ at a point $(r,m)$ if $\phi$ holds
in all points $(r',m')$ such that $(r,m) \sim_a (r',m')$.
Intuitively, an agent knows $\phi$ at $(r,m)$ if $\phi$ is implied by
the information in the local state $r_a(m)$.
We give formal semantics for a language of knowledge (and time and
plausibility) in \Sref{sec:plaus-knowledge}.

\xam
\label{xam:diag-sys}
The circuit diagnosis problem has been well studied in the literature
(see \cite{davis} for an overview). Consider a circuit that contains
$n$ logical components $c_1,\ldots,c_n$ and $k$ lines $l_1, \ldots,
l_k$. The agent can set the values
on the input lines of the circuit and observe the values on the
output lines. The agent then compares the actual output values to the
expected output values and attempts to locate faulty components. Since
a single test is usually insufficient to locate the problem, the agent
might perform a sequence of such tests.

We want to model diagnosis
using an interpreted system.
To do so, we need to describe
the agent's local state, the state of the environment,
and some appropriate propositions for reasoning about diagnosis.
Intuitively,
the agent's
state is the sequence of input-output relations observed, while the
environment's state describes
the current state of the circuit. This
consists of the {\em failure
set\/}, that is, the set of faulty components of the circuit
and the values on all the lines in the circuit.
Each run describes the results of a specific series of tests the agent
performs and the results she observes.  We make two additional
assumptions: (1) the agent does not forget what tests were performed
and their results, and (2) the faults are persistent and do not change
over time.

To make this precise,
we define the environment state at a point $(r,m)$ to consist
of the
failure set at
$(r,m)$,
which we denote $\fault(r,m)$, as well as the values of all the lines
in the circuit. We require that the environment state be
consistent with the description of the circuit.
Thus, for example, if $c_1$ is an AND gate with input lines $l_1$ and
$l_2$ and output line $l_3$, then if $r_e(m)$ says that $c_1$ is not
faulty, then we require that
there is a 1 on $l_3$ if and only if there is a 1 on both $l_1$ and
$l_2$.%
\footnote{Note that this means that we can recover the behavior of the
circuit (although not necessarily its exact description)
by simply looking at the environment state at a point where there are no
failures.  Of course, if we could have a yet richer environment state
that encodes the actual description of the circuit, but this is
unnecessary for the analysis we do here.}
We capture the assumption
that faults are persistent by requiring that $\fault(r,m) =
\fault(r,0)$.
For our later results, it is useful to describe the agent's observations
using our logical language.
Consider the set $\Phi_\diag = \{ f_1,
\ldots, f_n, h_1, \ldots, h_k \}$
of primitive propositions,
where $f_i$ denotes that component $i$ is faulty and $h_i$
denotes that there is a 1 on line $i$ (that is,
line $i$ in a ``high'' state).
An observation is a conjunction of literals of the form $h_i$ and
$\neg h_i$.
The agent's state at time $m$ is a sequence of $m$ such observations.
Formally, we define the agent's state $r_a(m)$ to be $\<
o_1, \ldots, o_m\>$, where, intuitively, $o_k$ is the formula describing
the input-output relation observed at time $k$.
We use the notation $\io(r,k)$ to denote
the
formula describing the
observation made by the agent at the point $(r,k)$.
Given this language, we can define the interpretation $\pi_\diag$
in the obvious way.
We say that an
observation $o$ is {\em
consistent\/} with an environment state $r_e(m)$ if the states of
the input/output lines in $r_e(m)$ agree with these in $o$.  The
system $\R_\diag$ consists of all runs $r$ satisfying these
requirements in which $\io(r,m)$ is consistent with $r_e(m)$ for all
times $m$.

Given the system $(\R_\diag,\pi_\diag)$, we can examine the agent's
knowledge after making a sequence of observations $o_1,\ldots,o_m$. It
is easy to see that the agent knows that the fault set must be one
with which all the observations are consistent. However, the agent
cannot rule out any of these fault sets. Thus, even if all the
observations are consistent with the circuit being fault-free, the
agent does not know that the circuit is fault-free, since there might
be a fault that manifests itself only in configurations that have not
yet been tested.
Of course, the agent might strongly believe that the circuit is
fault-free, but we cannot (yet) express this fact in our formalism.
The next section rectifies this problem.
\exam

\subsection{Plausibility Measures}

Most non-probabilistic approaches to belief change require (explicitly
or implicitly) that the agent has some ordering over possible
alternatives.  For example, the agent might have a preference ordering
over possible worlds \cite{Boutilier94AIJ2,Grove,KM92} or an
entrenchment ordering over formulas \cite{MakGar}.  This ordering
dictates how the agent's beliefs change. For example, in \cite{Grove},
the new beliefs are characterized by the most preferred worlds that
are consistent with the new observation, while in \cite{MakGar},
beliefs are discarded according to their degree of entrenchment until
it is consistent to add the new observation to the resulting set of
beliefs.
We represent this ordering using {\em plausibility measures\/}, which
were introduced in \cite{FrH7,FrH5Full}.
We briefly review the relevant definitions and results
here.

Recall that a probability space is a tuple $(W,\F,\Pr)$, where $W$ is
a set of worlds, $\F$ is an algebra of {\em measurable\/} subsets of
$W$ (that is, a set of subsets closed under union and complementation
to which we assign probability), and $\Pr$ is a {\em probability
measure}, that is, a function mapping each set in $\F$ to a number in
$[0,1]$ satisfying the well-known probability axioms ($\Pr(\emptyset)
= 0$, $\Pr(W) = 1$, and $\Pr(A \union B) = \Pr(A) + \Pr(B)$, if $A$
and $B$ are disjoint).

Plausibility spaces are a direct generalization of probability
spaces. We simply replace the probability measure $\Pr$ by a {\em
plausibility measure\/} $\Pl$, which, rather than mapping sets in $\F$
to numbers in $[0,1]$, maps them to elements in some arbitrary
partially ordered set. We read $\Pl(A)$ as ``the plausibility of set
$A$''.  If $\Pl(A) \le \Pl(B)$, then $B$ is at least as plausible as
$A$. Formally, a {\em plausibility space\/} is a tuple $S =
(W,\F,\Pl)$, where $W$ is a set of worlds, $\F$ is an algebra of
subsets of $W$, and $\Pl$ maps sets in $\F$ to some domain $D$ of {\em
plausibility values\/} partially ordered by a relation $\le_D$ (so
that $\le_D$ is reflexive, transitive, and anti-symmetric).  We assume
that $D$ is {\em pointed\/}: that is, it contains two special elements
$\top_D$, and $\bottom_D$ such that $\bottom_D \le_D d \le_D \top_D$
for all $d \in D$; we further assume that $\Pl(\Omega) = \top_D$ and
$\Pl(\emptyset) = \bottom_D$.  As usual, we define the ordering $<_D$
by taking $d_1 <_D d_2$ if $d_1 \le_D d_2$ and $d_1 \neq d_2$.  We
omit the subscript $D$ from $\le_D$, $<_D$, $\top_D$, and $\bottom_D$
whenever it is clear from context.

Since we want a set to be at least as plausible as any of its subsets,
we require
\begin{cond}{A1}
If $A \subseteq B$, then $\Pl(A) \le \Pl(B)$.
\end{cond}

Some brief remarks on this definition: We have deliberately suppressed
the domain $D$ from the tuple $S$, since for the purposes of this
paper, only the ordering induced by $\le$ on the subsets in $\F$ is
relevant.  The algebra $\F$ also does not play a significant role in
this paper.  Unless we say otherwise, we assume $\F$ contains all
subsets of interest and suppress mention of $\F$, denoting a
plausibility space as a pair $(W,\Pl)$.

Clearly plausibility spaces generalize probability spaces. In
\cite{FrH5Full,FrH7} we show that they also generalize {\em belief
function\/} \cite{Shaf}, {\em fuzzy measures\/} \cite{WangKlir}, {\em
possibility measures\/} \cite{DuboisPrade88}, {\em ordinal ranking\/}
(or {\em $\kappa$-ranking\/}) \cite{Goldszmidt+Pearl:1996,spohn:88}, {\em
preference orderings\/} \cite{KLM,Shoham87}, and {\em parameterized
probability distributions \/} \cite{GMPFull} that are used  as a
basis for Pearl's {\em $\epsilon$-semantics\/} for defaults \cite{Pearl90}.

Our goal is to describe the agent's beliefs in terms of plausibility.
    To do this, we describe how to evaluate statements of the form
    $\Bel\phi$ given a plausibility space. In fact, we use a
    richer logical language that also allows us to describe how the agent
    compares different alternatives. This is the
    logic of conditionals.
Conditionals are statements of the form $\phi\Cond\psi$,
    read ``given
    $\phi$, $\psi$ is plausible'' or ``given $\phi$, then by default
    $\psi$''. The syntax of the logic of
    conditionals is simple: we start with primitive propositions and
    close off under conjunction, negation and the modal operator
    $\Cond$. The resulting language is denoted $\LCond$.

A {\em  plausibility structure\/} is a tuple $\PL = (W, $\Pl$, \pi)$,
    where $W$ is a set of possible worlds, $\Pl$ is a plausibility
measure on $W$, and $\pi(w)$ is a truth assignment to primitive
propositions.
Given a plausibility structure $\PL = (W, $\Pl$, \pi)$, we define
$\intension{\phi}_\sPL = \{ w \in W : \pi(w) \sat \phi \}$ to be the
set of worlds that satisfy $\phi$. We omit the subscript $\PL$, when
it is clear from the context.
Conditionals are evaluated according to a rule that is essentially
the same as the one used by Dubois and Prade
\citeyear{DuboisPrade:Defaults91} to evaluate conditionals using
possibility measures:
\begin{itemize}\denselist
\item $\PL \sat \phi\Cond\psi$ if either
    $\Pl(\intension{\phi}) = \bottom$ or
    $\Pl(\intension{\phi\land\psi}) >
    \Pl(\intension{\phi\land\neg\psi})$.
\end{itemize}
Intuitively, $\phi \Cond \psi$ holds vacuously if $\phi$ is
impossible; otherwise, it holds if $\phi \land \psi$ is more plausible
than $\phi \land \neg \psi$.  As we show in \cite{FrH5Full},
this semantics of conditionals also
generalizes the semantics of conditionals in $\kappa$-ranking
\cite{Goldszmidt+Pearl:1996},
and PPD structures \cite{GMPFull}.
As we also show in \cite{FrH5Full}, this semantics for conditionals
generalizes the semantics of preferential structures. As this
relationship plays a role in the discussion below, we review the
necessary definitions here.  A {\em preferential structure\/} is a
tuple $(W,\prec,\pi)$, where $\prec$ is a partial order on $W$.
Roughly speaking, $w \prec w'$ holds if $w$ is {\em preferred\/} to $w'$.%
\footnote{We follow the standard notation for preference here
\cite{KLM}, which uses the (perhaps confusing) convention
of placing the more likely (or less abnormal) world on the left of the
$\prec$ operator.
Unfortunately, when translated to plausibility, this will mean $w \prec
w'$ holds iff $\Pl(\{w\} > \Pl(\{w'\})$.}
The intuition \cite{Shoham87} is that a preferential structure
satisfies a
conditional $\phi\Cond\psi$ if all the
most preferred worlds (\ie
the minimal worlds according to $\prec$) in $\intension{\phi}$ satisfy
$\psi$.
However, there may be no minimal worlds in
$\intension{\phi}$. This can happen if $\intension{\phi}$ contains an
infinite descending sequence $\ldots \prec w_2 \prec w_1$. What do we
do in these structures?  There are a number of options: the first is
to assume that, for each formula $\phi$, there are minimal worlds in
$\intension{\phi}$; this is the assumption actually made in
\cite{KLM}, where it is called the {\em smoothness\/} assumption.  A
yet more general definition---one that works even if $\prec$ is not
smooth---is given in \cite{Lewis73,Boutilier94AIJ1}.  Roughly
speaking, $\phi \Cond \psi$ is true if, from a certain point on,
whenever $\phi$ is true, so is $\psi$.  More formally,
\begin{quote}
$(W,\prec,\pi)$ satisfies $\phi\Cond\psi$, if for every world $w_1 \in
\intension{\phi}$, there is a world $w_2$ such that (a) $w_2 \preceq
w_1$ (so that $w_2$ is at least as normal as $w_1$), (b) $w_2 \in
\intension{\phi\land\psi}$, and (c) for all worlds $w_3 \prec w_2$, we
have $w_3 \in \intension{\phi \rimp \psi}$ (so any world more normal
than $w_2$ that satisfies $\phi$ also satisfies $\psi$).
\end{quote}
It is easy to verify that this definition is equivalent to the
earlier one if $\prec$ is smooth.

\pro{\rm\cite{FrH5Full}}\label{pro:prec}
If $\prec$ is a preference ordering on $W$, then there is a
plausibility measure $\Pl_\prec$ on $W$ such that
$(W,\prec,\pi) \sat \phi\Cond\psi$ if and only if $(W,\Pl_\prec,\pi) \sat
    \phi\Cond\psi$.
\epro

We briefly describe the construction of $\Pl_\prec$ here,
since we use it in the sequel.  Given a preference order $\prec$ on
$W$,
let $D_0$ be the domain of plausibility values consisting of
one element $d_w$ for every element $w \in W$.
We define a partial order on $D_0$ using $\prec$:
$d_v < d_w$
if $w \prec v$. (Recall that $w \prec
    w'$ denotes that $w$ is preferred to $w'$.)  We then take $D$ to
    be the smallest set containing
$D_0$
that is
closed
under
least upper bounds (so that every set
    of elements in $D$
has a least upper bound in $D$).
For a subset $A$ of $W$, we can then define $\Pl_\prec(A)$ to be the
least upper bound of $\{d_w: w \in A\}$.  Since $D$ is closed under least
upper bounds, $\Pl(A)$ is well defined.
As we show in \cite{FrH5Full},
this choice of $\Pl_\prec$ satisfies Proposition~\ref{pro:prec}.

The results of \cite{FrH5Full} show that this semantics for
conditionals generalizes previous semantics for conditionals.
Does this semantics capture our
intuitions about conditionals? In the AI literature, there has been
little consensus on
the ``right'' properties for defaults (which are essentially
conditionals). However, there has been
some consensus on a reasonable ``core'' of inference rules for default
reasoning.
This core is usually known as the KLM properties \cite{KLM}, and
includes such properties as
\begin{cond}{AND}
{From} $\phi\Cond\psi_1$ and $\phi\Cond\psi_2$ infer
    $\phi\Cond \psi_1 \land \psi_2$
\end{cond}
\begin{cond}{OR}
{From} $\phi_1\Cond\psi$ and $\phi_2\Cond\psi$ infer
    $\phi_1\lor\phi_2\Cond \psi$
\end{cond}
What constraints on plausibility spaces gives us the KLM properties?
Consider the following two conditions:
\begin{cond}{A2}
If $A$, $B$, and $C$ are pairwise disjoint sets,
$\Pl(A \union B) > \Pl(C)$, and $\Pl(A \union C) > \Pl(B)$, then
$\Pl(A) > \Pl(B \union C)$.
\end{cond}
\begin{cond}{A3}
If $\Pl(A) = \Pl(B) = \bottom$, then $\Pl(A \union B) = \bottom$.
\end{cond}

A plausibility space $(W,\Pl)$ is {\em qualitative\/} if it satisfies
A2 and A3.  A plausibility structure $(W,\Pl,\pi)$ is qualitative if
$(W,\Pl)$ is a qualitative plausibility space.  In \cite{FrH5Full}, we show
that, in a very general sense, qualitative plausibility structures
capture default reasoning. More precisely, we show that the KLM
properties are sound with respect to a class of plausibility
structures if and only if the class consists of qualitative
plausibility structures. (We also provide a weak condition that we
show
is necessary and sufficient for the KLM properties to be
complete.) These results show that plausibility
structures provide a unifying framework for the characterization of
default entailment in these different logics.

\subsection{Plausibility and Knowledge}\label{sec:plaus-knowledge}

In \cite{FrH1Full} we show how plausibility measures can be incorporated
into the multi-agent system framework of \cite{HF87}.
This allows us to describe the agent's assessment of the
possible states the system is in at each point in time. At the same
time we also introduce conditionals into the logical language in order
to reason about these plausibility assessments.
We now review the relevant details.

An {\em (interpreted) plausibility system\/} is a tuple $(\R,\pi,\P)$
where, as before, $\R$ is a set of runs and $\pi$ maps each point to a
truth assignment, and where $\P$ is a {\em plausibility assignment
function\/} mapping each point $(r,m)$ to a qualitative plausibility
space $\P(r,m) = (\Omega_{(r,m)},\Pl_{(r,m)})$. Intuitively, the
plausibility space $\P(r,m)$ describes the relative plausibility of
events from the point of view of the agent at $(r,m)$.  In this paper,
we restrict our attention to plausibility spaces that satisfy two additional
assumptions:
\begin{itemize}
\item $\Omega_{(r,m)} = \{ (r',m') | (r,m) \sim_a (r',m')
\}$.  Thus, the agent considers plausible only situations that are
possible according to her knowledge.
\item if
$(r,m) \sim_a (r',m')$ then $\P(r,m) = \P(r',m')$.
This means
that the
plausibility space is a function of the agent's local state.%
\footnote{The framework presented in \cite{FrH1Full} is more general than
this, dealing with multiple agents and allowing the agent to consider
several plausibility spaces in each local state. The simplified
version we present here suffices to capture belief revision and
update.}
\end{itemize}

We define a logical language to reason about interpreted systems. The
syntax of the logic is simple; we start with primitive propositions
and close off under conjunction, negation, the $\Know$ modal operator
($\Know\phi$ says that the agent knows $\phi$), the $\Next$ modal
operator  ($\Next\phi$ says that $\phi$ is true at the next time
step), and the $\Cond$ modal operator. The resulting language is
denoted $\LKPT$.%
\footnote{It is easy to add other temporal modalities such as {\em until,
    eventually, since\/}, etc. These do not play a role in this paper.}
We recursively assign truth values to formulas in $\LKPT$ at a
point $(r,m)$ in a plausibility system $\Sys$.  The truth of primitive
propositions is determined by $\pi$, so that
\begin{quote}
$(\Sys,r,m) \sat p$ if $\pi(r,m)(p) = {\bf true}$.
\end{quote}
Conjunction and negation are treated in the standard way, as is
knowledge:
The agent knows $\phi$
at $(r,m)$ if $\phi$ holds
at all points that she cannot distinguish from $(r,m)$.  Thus,
\begin{quote}
$(\Sys,r,m)\sat
    \Know\phi$ if $(\Sys,r',m')\sat \phi$ for all $(r',m') \sim_a (r,m)$.
\end{quote}
    $\Next\phi$ is true at $(r,m)$ if $\phi$ is true at $(r,m+1)$. Thus,
\begin{quote}
    $(\Sys,r,m) \sat \Next\phi$ if $(\Sys,r,m+1)\sat\phi$.
\end{quote}
Finally, we define the conditional operator $\Cond$ to describe the
    agent's plausibility assessment at the current time.
    Let $\intension{\phi}_{(r,m)} = \{ (r',m') \in
    \Omega_{(r,m)} : (\Sys,r,m) \sat \phi \}$.
$$\mbox{$(\Sys,r,m) \sat \phi \Cond \psi$ if either
   $\Pl_{(r,m)}(\intension{\phi}_{(r,m)}) = \bot$ or\linebreak[3] $\Pl_{(r,m)}(
    \intension{\phi\land\psi}_{(r,m)}) > \Pl_{(r,m)}(
    \intension{\phi\land\neg\psi}_{(r,m)})$.}$$

We now define a notion of {\em belief}.  Intuitively, the agent
    believes $\phi$ if $\phi$ is more plausible than not. Formally, we
    define $\Bel\phi \dimp (\True \Cond\phi)$.

In \cite{FrH1Full} we prove that, in this framework, knowledge is an S5
operator, the conditional operator $\Cond$ satisfies the usual axioms
of conditional logic \cite{Burgess81}, and $\Next$ satisfies the
usual properties of temporal logic \cite{MP1}.
In addition,
these properties imply that belief is a K45 operator,
and the interactions between knowledge and belief are captured by the
axioms $\Know\phi \rimp \Bel\phi$ and $\Bel\phi \rimp \Know\Bel\phi$.

\xam
\label{xam:diag-plaus}
\cite{FrH1Full}
We add a plausibility measure to the system defined in
\Xref{xam:diag-sys}. We define $\Sys_\diag =
(\R_\diag,\pi_\diag,\P_\diag)$, where $\P_\diag$ is the plausibility
assignment we now describe. We assume that failures of individual
components are independent of one another. If we also assume that the
likelihood of each component failing is the same, and also that this
likelihood is small (\ie failures are exceptional), then we can
construct
a plausibility measure as follows.  If $(r',m)$ and $(r'',m)$ are two
points in $\Omega_{(r,m)}$, we say that $(r',m)$ is more plausible
than $(r'',m)$ if $\Card{\fault(r',m)} < \Card{\fault(r'',m)}$, that
is, if the failure set at $(r',m)$ consists of fewer faulty components
than at
$(r'',m)$. We extend these comparisons to sets: $\Pl_{(r,m)}(A) \le
\Pl_{(r,m)}(B)$ if $\min_{(r',m) \in A}(\Card{\fault(r',m)}) \ge \min_{(r',m)
\in B}(\Card{\fault(r',m)})$; that is, $A$ is less plausible if all the
points in $A$ have failure sets of larger cardinality then the minimal
one in $B$.
With this plausibility measure,
if all of the agent's observations
up to time $m$
are consistent with there being no failures,
then the agent believes that all
components are functioning correctly. On the other hand, if the
observations do not match the expected output of the circuit, then
the agent considers minimal failure sets that are consistent with her
observations. Thus, if the observations are consistent with a failure
of $c_1$, or a failure of $c_3$, or the combined failure of $c_2$ and
$c_7$, then the agent believes that either $c_1$ or $c_3$ is faulty,
but not both.

We now make this more precise.
A {\em failure set\/} (\ie a diagnosis)
is characterized by a complete formula over $f_1,\ldots,f_n$---that
is, one  that determines the truth values all these propositions.
For example, if $n=3$, then $f_1 \land \neg f_2 \land \neg f_3$
characterizes the failure set $\{c_1\}$.   We define $\Diag{(r,m)}$ to be
the set of failure sets
(\ie diagnoses) that the agent considers possible at $(r,m)$; that is
$\Diag{(r,m)} = \{ f \in F : (\Sys_\diag,r,m) \sat
\neg \Bel \neg f \}$ where $F$
is the set of all possible failure sets.

Belief change in
$\Sys_\diag$ is characterized by the following proposition.

\pro\label{pro:diag-sys}
If there is some $f \in \Diag{(r,m)}$ that is consistent with
the new observation $\io(r,m+1)$, then $\Diag{(r,m+1)}$
consists of all the failure sets in $\Diag{(r,m)}$ that are
consistent with $\io(r,m+1)$.  If all $f \in \Diag{(r,m)}$ are
inconsistent with $\io(r,m+1)$, then $\Diag{(r,m+1)}$ consists
of all failure sets of cardinality $j$ that are consistent with
$\io(r,1), \ldots, \io(r,m+1)$, where $j$ is the least cardinality for
which there is at least one failure set consistent with these
observations.
\epro
Thus, in $\Sys_\diag$, a new observation consistent with the current
set of most likely explanations reduces this set (to those consistent
with the new observation).  On the other hand, a surprising
observation (one inconsistent with the current set of most likely
explanations) has a rather drastic effect.  It easily follows from
Proposition~\ref{pro:diag-sys} that if $\io(r,m+1)$ is surprising,
then $\Diag{(r,m)} \inter \Diag{(r,m+1)} = \emptyset$,
so the agent discards all her current explanations in this case.
Moreover, an easy induction on $m$ shows that if $\Diag{(r,m)}
\inter \Diag{(r,m+1)} = \emptyset$, then the cardinality of
the failure sets in $\Diag{(r,m+1)}$ is greater than the
cardinality of failure sets in $\Diag{(r,m)}$.
Thus, in this case, the explanations in $\Diag{(r,m+1)}$ are more
complicated than those in $\Diag{(r,m)}$.
\exam

\subsection{Conditioning}\label{sec:conditioning}
In an interpreted system, the agent's beliefs change from point to
point as her plausibility space changes.
The general framework
does not put any constraints on
how the plausibility space changes.
If we were thinking
probabilistically, we could imagine the agent starting with a prior on
the runs in the system.  Since a run describes a complete history over
time, this means that the agent puts a prior probability on the
possible sequences of events that could happen.  We would then expect
the agent to modify her prior by conditioning on whatever information
she has learned. As we  show below, this notion of conditioning is
closely related to belief revision and update.
We remark that we
are not the first to applying conditioning in the context of belief
change (\cf
\cite{Goldszmidt+Pearl:1996,spohn:88});
the details are a little more complex in our framework, because we model
time explicitly.

We start by making the simplifying assumption that we are dealing with
{\em synchronous\/} systems where agents have {\em perfect recall\/}
\cite{HV2}. Intuitively, this means that the agent knows what the time
is and does not forget the observations she has made.  Formally, a
system is synchronous if $(r,m) \sim_a (r',m')$ only if $m = m'$.  In
synchronous systems, the agent has perfect recall if $(r',m+1) \sim_a
(r,m+1)$ implies $(r',m) \sim_a (r,m)$.  Thus, the agent considers run
$r$ possible at the point $(r,m+1)$ only if she also considers it
possible at $(r,m)$.  This means that any runs considered impossible
at $(r,m)$ are also considered impossible at $(r,m+1)$: the agent does
not forget what she knew.

Just as with probability, we assume that the agent has a prior
plausibility measure on runs that describes her prior assessment on
the possible executions of the system. As the agent gains knowledge,
she updates her prior by conditioning.  More precisely, at each point
$(r,m)$, the agent conditions her previous assessment on the set of
runs considered possible at $(r,m)$.  This results in an updated
assessment (posterior) of the plausibility of runs.  This posterior
induces, via a projection from runs to points, a plausibility measure
on points.  We can think of the agent's posterior at time $m$ as
simply her prior conditioned on her knowledge at time $m$.

Formally, the prior plausibility of the agent is a plausibility
measure $\P_a = (\R,\Pl_a)$ over the runs in the system. If $A$ is a
set of points, we define $\R(A) =
\{ r : \exists m ((r,m) \in A)\}$ to
be the set of runs on which the points in $A$ lie.  The agent updates
plausibilities
by conditioning in $\Sys$ if the following condition is met:
\begin{cond}{PRIOR}
There is prior $\P_a = (\R,\Pl_a)$ such that
for all runs $r \in \R$, times $m$, and sets $A,B \subseteq
\Omega_{(r,m)}$,
$\Pl_{(r,m)}(A) \le
    \Pl_{(r,m)}(B)$ if and only if $\Pl_a(\R(A)) \le \Pl_a(\R(B))$.
\end{cond}
This definition implies that the agent's plausibility assessment at each point
    is determined, in a straightforward fashion, by her prior.

As shown in \cite{FrH1Full}, in synchronous systems that satisfy PRIOR
where agent have perfect recall, we can say even more:
the agent's plausibility measure at time $m+1$ is determined by her
plausibility measure at time $m$. To make this precise, if $A$ is a
set of points, let $\MapBack(A) = \{ (r,m) : (r,m+1) \in A \}$.

\thm\label{thm:prior}{\rm \cite{FrH1Full}}.
Let $\Sys$ be a synchronous system satisfying
PRIOR where agents have perfect recall.
Then $\Pl_{(r,m+1)}(A) \le \Pl_{(r,m+1)}(B)$ if and only if
$\Pl_{(r,m)}(\MapBack(A)) \le \Pl_{(r,m)}(\MapBack(B))$, for all runs
$r$, times $m$, and sets $A,B \subseteq \Omega_{(r,m+1)}$.
\ethm

Thus, in synchronous systems
where agents have perfect recall
PRIOR implies
a ``local'' rule for update that incrementally changes the agent's
plausibility at each step. This local rule consists of two steps.
First, the agent's plausibility at time $m$ is projected to time $m+1$
points. Second, time $m+1$ points that are inconsistent with the agent
knowledge at $(r,m+1)$ are discarded. This procedure implies that the
relative plausibility of  two sets of runs does not change unless
one of them is incompatible with the new knowledge.

\xam
It is easy to verify that the system $\Sys_\diag$ we consider in
Example~\ref{xam:diag-plaus} satisfies PRIOR.  The prior $\P_a$ is
determined by the failure set in each run in a manner similar to the
construction of $\Pl_{(r,m)}$. That is, $R_1$ is more plausible than
$R_2$ if there is a run in $R_1$ with a smaller failure set than all
the runs in $R_2$.
\exam

\section{Review of Revision and Update}\label{sec:review}

We now present a brief review of belief revision and update.

{\em Belief revision\/} attempts to describe how a rational agent
incorporates new beliefs.  As we said earlier, the main intuition is
that as few changes as possible should be made.  Thus, when something
is learned that is consistent with earlier beliefs, it is just added
to the set of beliefs.  The more interesting situation is when the
agent learns something inconsistent with her current beliefs. She must
then discard some of her old beliefs in order to incorporate the new
belief and remain consistent. The question is which ones?

The most widely accepted notion of belief revision is defined by the
AGM theory \cite{agm:85,Gardenfors1}. This theory was originally
developed in philosophy of science, where one attempts to understand
when a scientist changes her beliefs (\eg theory of physical laws) in
a rational manner. In this context, it seems reasonable to assume
that the world is {\em static\/}; that is, the laws of physics do not
change while the scientist is performing experiments.

Formally, this theory assumes a logical
language
$\Le$ over a set $\Phis$ of primitive propositions with a
consequence relation $\vdash_\Le$ that contains
the propositional calculus and satisfies the deduction theorem.
The AGM approach assumes that an agent's epistemic state is represented
by a {\em belief set}, that is, a set $K$ of formulas in the
language $\Le$.%
\footnote{For example, G\"ardenfors~\citeyear[p.~21]{Gardenfors1} says
``A simple way of modeling the epistemic state of an individual is to
represent it by a {\em set} of sentences.''}
There is
also assumed to be a revision operator $\Rev$ that takes a belief set
$A$ and a formula $\phi$ and returns a new belief set $A \Rev \phi$,
intuitively, the result of revising $A$ by $\phi$. The following AGM
postulates are an attempt to characterize the intuition of ``minimal
change'':
\begin{itemize}\denselist
\item[(R1)] $A \Rev \phi$ is a belief set
\item[(R2)] $\phi \in A\Rev\phi$
\item[(R3)] $A \Rev\phi \subseteq Cl(A \cup \{\phi\})$%
\footnote{
$Cl(A) = \{ \phi | A \vdash_\Le \phi \}$
is the deductive closure
of a set of formulas $A$.
}
\item[(R4)] If $\neg\phi \not\in A$ then $Cl(A \cup \{\phi\})
    \subseteq A \Rev\phi$
\item[(R5)] $A\Rev\phi = Cl(\False)$ if and only if
$\vdash_{\Le} \neg\phi$
\item[(R6)] If
$\vdash_{\Le} \phi \dimp \psi$
then $A\Rev\phi = A\Rev\psi$
\item[(R7)] $A \Rev (\phi\land\psi) \subseteq Cl(A\Rev\phi \cup \{\psi\})$
\item[(R8)] If $\neg\psi \not\in A\Rev\phi$ then $Cl(A\Rev\phi \cup
    \{\psi\}) \subseteq A \Rev (\phi\land\psi)$.
\end{itemize}

The essence of these postulates is the following. After a revision by
$\phi$ the belief set should include $\phi$ (postulates R1 and R2). If
the new belief is consistent with the belief set, then the revision
should not remove any of the old beliefs and should not add any new
beliefs except these implied by the combination of the old beliefs
with the new belief (postulates R3 and R4). This condition is called
{\em persistence\/}. The next two conditions discuss the coherence of
beliefs. Postulate R5 states that the agent is capable of
incorporating any consistent belief and postulate R6 states that the
syntactic form of the new belief does not affect the revision process.
The last two postulates enforce a certain coherency on the outcome of
revisions by related beliefs.  Basically, they state that if $\psi$ is
consistent with $A\Rev\phi$ then $A\Rev(\phi\land\psi)$ is just
$A \Rev \phi \Rev \psi$.

The notion of {\em belief  update\/} originated in the database community
\cite{KellerWinslett,Winslett}. The problem is how a knowledge base
should change when something is learned about the world. For example,
suppose that a transaction adds to the knowledge base the fact ``Table
7 is in Office 2'', which contradicts the previous belief that ``Table
7 is in Office 1''. What else should change? The intuition that update
attempts to capture is that such a transaction describes a change that
has occurred in the world. Thus, in our example,
by applying update we might conclude that the reason that the table is
in Office 2 is that it was moved, not that our earlier beliefs were
false.
This example shows that, unlike revision, update does not assume
that the world is static.

Katsuno and Mendelzon \citeyear{KM91} suggest a set of postulates that
an update operator should satisfy. The update postulates are expressed
in terms of formulas, not belief sets. That is, an update operator
$\Upd$ maps a pair of formulas, one describing the agent's current
beliefs and the other describing the new observation, to a new formula
that describes the agent's updated beliefs. This is not unreasonable,
since we can identify a formula $\phi$ with the belief set $Cl(\phi)$.
Indeed, if $\Phi$ is finite (which is what Katsuno and Mendelzon
assume) every belief set $A$ can be associated with some formula
$\phi_A$ such that $Cl(\phi_A) = A$, and every formula $\phi$
corresponds to a belief set $Cl(\phi)$. Thus, any update operator
induces an operator that maps a belief state and an observation to a
new belief state. We slightly abuse notation and use the same symbol
to denote both types of mappings. We say that a belief set $A$ is {\em
complete\/} if, for every $\phi \in \Le$, either $\phi\in A$ or
$\neg\phi\in A$. A formula $\mu$ is {\em complete\/} if $Cl(\mu)$ is
complete.

The KM postulates are:
\begin{itemize}\denselist
\item[(U1)] $\vdash_\Le \mu \Upd \phi \rimp \phi$
\item[(U2)] If $\vdash_\Le\mu\rimp\phi$, then $\vdash_\Le \mu \Upd\phi
    \dimp \mu$
\item[(U3)] $\vdash_\Le \neg \mu\Upd\phi$ if and only if
    $\vdash_\Le \neg\mu$ or $\vdash_\Le \neg\phi$
\item[(U4)] If $\vdash_\Le \mu_1 \dimp \mu_2$ and $\vdash_\Le \phi_1
    \dimp \phi_2$ then
    $\vdash_\Le \mu_1\Upd\phi_1 \dimp \mu_2\Upd\phi_2$
\item[(U5)] $\vdash_\Le (\mu\Upd\phi)\land\psi \rimp
    \mu\Upd(\phi\land\psi)$
\item[(U6)] If $\vdash_\Le \mu\Upd\phi_1 \rimp \phi_2$ and $\vdash_\Le
    \mu\Upd\phi_2 \rimp \phi_1$, then $\vdash_\Le\mu \Upd \phi_1 \dimp
    \mu\Upd\phi_2$
\item[(U7)] If $\mu$ is complete then
    $\vdash_\Le (\mu\Upd\phi_1)\land(\mu\Upd\phi_2) \rimp \mu\Upd(\phi_1 \lor
    \phi_2)$%
\item[(U8)] $\vdash_\Le (\mu_1 \lor \mu_2) \Upd \phi \dimp (\mu_1\Upd\phi)
    \lor (\mu_2 \Upd \phi)$.
\end{itemize}

The essence of these postulates is as following. After learning
$\phi$, the agent believes $\phi$ (postulate U1, which is
analogous to R2). If $\phi$ is already
believed, then updating by $\phi$ does not change the agent's beliefs
(postulate U2, which is a weaker version of R3 and R4). The next two
postulates (U3 and U4) deal with coherence of the belief change
process. They are analogous to R5 and R6, respectively, with minor
differences. Postulates U5 and U6 deal with observations that are
related to each other. U5 states that beliefs after learning $\phi$ that
are consistent with $\psi$ are also believed after learning $\phi\land\psi$.
U6 states that if $\phi_2$ is believed after learning $\phi_1$ and
$\phi_1$ is believed after learning $\phi_2$, then learning either
$\phi_1$ or $\phi_2$ leads to the same belief set.
Finally, U7 and U8 deal with decomposition properties of the update
operation. U7 states that if $\mu$ is essentially a truth assignment
to $\L$, then if $\psi$ is believed after learning $\phi_1$ and is
also believed after learning $\phi_2$ then it is believed after
learning $\phi_1\lor\phi_2$. U8 states that the update of the
knowledge base can be computed by independent updates on each
sub-part of the knowledge. That is, if $\mu = \mu_1 \lor\mu_2$, then we
can apply update to each of $\mu_1$ and $\mu_2$, and then combine the results.

\section{Belief Change Systems}\label{bcs}
\label{sec:bcs}
\label{SEC:BCS}

We want to model belief change---particularly belief revision and
belief update---in the framework of systems.  To do so, we consider a
particular class of systems that we call {\em belief change systems}.
In belief change systems, the agent makes observations about an
external environment.  Just as is (implicitly) assumed in both
revision and update, we assume that these observations are described by
formulas in some logical language.  We then make other assumptions
regarding the plausibility measure used by the agent.  We formalize
our assumptions as conditions BCS1--BCS5, described below, and say
that a system $\Sys = (\R,\pi,\P)$ is a {\em belief change system\/}
if it satisfies these conditions.
We denote by $\SBCS$ the set of belief change systems.

Assumption BCS1 formalizes the intuition that our language includes
propositions for reasoning about the environment, whose truth depends
only on the environment state.
\begin{cond}{BCS1}
The language $\L$ includes a propositional sublanguage $\Le$ over a set
$\Phis$ of primitive
propositions.  $\Le$ contains the usual propositional connectives and
comes equipped with a consequence relation $\vdash_\Le$.  The
interpretation
$\pi(r,m)$ assigns truth to propositions in $\Phis$ in such a way that
\begin{itemize}\denselist
\item[(a)] $\pi(r,m)$ is consistent with $\vdash_\Le$, that is,
$\{ p : p \in \Phis,
\pi(r,m)(p) = $ {\bf true}$ \} \union \{ \neg p : p \in \Phis,
\pi(r,m)(p) = $ {\bf false}$ \}$ is $\vdash_\Le$ consistent,
and
\item[(b)] $\pi(r,m)(p)$ depends only on $r_e(m)$ for propositions in
$\Phis$; that is, $\pi(r,m)(p) = \pi(r',m')(p)$ whenever $r_e(m) =
r'_e(m')$.
\end{itemize}
\end{cond}
Part (b) of BCS1 implies that we can evaluate formulas in $\Le$ with
respect
to environment states; that is, if $\phi \in \Le$ and $r_e(m) =
r'_e(m')$, then $(\Sys,r,m) \sat \phi$ if and only if $(\Sys,r',m')
\sat \phi$.
Since the environment is all that is relevant for formulas
in $\Le$, if $\phi \in \Le$, we write $s_e \sat
\phi$ if $(\Sys, r,m) \sat \phi$ for some
point $(r,m)$ such that $r_e(m) = s_e$.

BCS2 is concerned with the form of the agent's local state.
Recall that, in our framework, the local state captures the relevant
aspects of the agent's epistemic state.
The functional form of the revision and update operators suggests
that all that matters regarding how an agent changes her beliefs are
the agent's current epistemic state
(which is taken by both AGM and KM to be a belief set)
and what is learned.
In terms of our framework, this suggests that
agent's local state at time $m+1$ should be a function of her local
state of time $m$ and the observation made at time $m$.
We in fact make the stronger assumption here that the agent's state
consists of the sequence of observations made by the agent.  This
means that the agent remembers all her past observations.   Note that
this surely implies that the agent's local state at time $m+1$ is
determined by her state at time $m$ and the observation made at time
$m$.
We make the further assumption that
the observations made by the agent can be described by formulas
in $\Le$.
Although
this is quite a strong assumption on the expressive power of
$\Le$,
it is standard in the literature:
both revision and update assume that observations can be expressed
as formulas in the language (see Section~\ref{sec:review}).
These assumptions are
formalized
in BCS2:
\begin{cond}{BCS2}
For all $r \in R$ and for all $m$, we have $r_a(m) =
\<\obs{r,1},\ldots,\obs{r,m}\>$ where $\obs{r,k} \in \Le$ for
$1 \le k \le m$.
\end{cond}
Intuitively, $\obs{r,k}$ is the observation the agent makes immediately
after the transition from time $k-1$ to time $k$ in run $r$. Thus, it
represents what the agent observes about the new state of the system at
time $k$.
Note that BCS2
implies that the agent's state at time $0$ is the empty sequence
in all runs. Moreover, it implies that $r_a(m+1) = r_a(m)
\cdot \obs{r,m+1}$, where $\cdot$ is the append operation on sequences.
That is, the agent's state at $(r,m+1)$ is the result
of appending to her previous state the latest observation she has made about
the system. It is not too hard to show that
belief change systems
are synchronous and agents in them have perfect recall.
(We remark that
the agents' local states are modeled in a similar way in the model of
knowledge bases presented in \cite{FHMV}.)

Clearly we want to reason in our language about the observations the
agent makes. Thus, we
assume that the language includes propositions that describe the
observations made by the agent.
\begin{cond}{BCS3}
The language $\L$ includes a set $\Phil$ of primitive propositions
disjoint from $\Phis$ such that $\Phil = \{ \learn(\phi) : \phi \in
\Le \}$.  Moreover,
$\pi(r,m)(\learn(\phi)) = $ {\bf true} if and only if $\obs{r,m} =
\phi$ for all runs $r$ and times $m$.
\end{cond}

In a system satisfying BCS1--BCS3, we can talk about belief change.
The agent's state encodes observations, and we have propositions that
allow us to talk about what is observed.  The next assumption is
somewhat more geared to situations where observations are always
``accepted'', so that after the agent observes $\phi$, she believes
$\phi$.  While this is not a necessary assumption, it is made by both
belief revision and belief update.  We capture this assumption here in
what is perhaps the simplest possible way: by assuming that
observations are reliable, so that the agent observes $\phi$ only if the
current state of the environment satisfies $\phi$.
This is certainly not the only way of enforcing the assumption that
observations are accepted, but it is perhaps the simplest, so we focus
on it here.
As we shall see, this assumption is consistent with both
revision and update, in the sense that we can capture both in systems
satisfying it.
\begin{cond}{BCS4}
$(\Sys,r,m) \sat \obs{r,m}$ for all runs $r$ and times $m$.
\end{cond}
Note that BCS4 implies that the agent
never observes $\False$. Moreover, it
implies that after observing $\phi$, the agent knows that $\phi$
is true.
In \cite{BFH1}, we consider an instance of our framework in which
observations are unreliable (so that BCS4 does not hold in general),
and examine the status of R2, the acceptance postulate, in this case.

Finally, we assume that belief change proceeds by conditioning.  While
there are certainly other assumptions that can be made, as we have tried
to argue, conditioning is a principled approach that captures the
intuitions of minimal change, given the observations.  And, as we shall
see, conditioning (as captured by PRIOR) is consistent with both revision
and update.
\begin{cond}{BCS5}
$\Sys$ satisfies PRIOR.
\end{cond}

Many interesting systems can be viewed as BCS's.
\xam
\label{xam:diag-BCS}
Consider the systems $\Sys_{\diag,1}$ and $\Sys_{\diag,2}$ of
Example~\ref{xam:diag-sys}.
Are these systems BCSs?
Not quite, since $\pi_\diag$ is not defined on primitive propositions
of the form $\learn(\phi)$, but we can easily embed
both systems in a BCS.
Let $\L_\diag$ the
propositional language defined over $\Phi_\diag$, and let $\Phi_\diag^+$
consist
of $\Phi_\diag$ together with all the primitive propositions of the form
$\learn(\phi)$ for $\phi \in \L_\diag$. Let $\pi_\diag^+$ be the obvious
extension of $\pi_\diag$ to $\Phi_\diag^+$, defined so that BCS3 holds.
Then in it is easy to see that $(\R_\diag,\pi_\diag^+,\P_{\diag,i})$ is a
BCS:
we take the $\Phis$ of BCS1 to be $\Phi_\diag$, and define
$\vdash_{\L_\diag}$ so that it enforces the relationships determined by
the circuit layout.  Thus, for example,
if $c_1$ is an AND gate with input lines $l_1$ and
$l_2$ and output line $l_3$, then we would have
$\vdash_{\L_\diag} \neg f_1
\rimp (h_3 \dimp h_1 \land h_2)$.  It is then easy to see that
BCS2--BCS5 hold by our construction.
\exam

These definitions set the
background for our presentation of belief revision and belief update.

\section{Capturing Revision}
\label{sec:revision}
\label{SEC:REVISION}

Revision can be captured by restricting to BCSs that satisfy several
additional assumptions. Before describing these assumptions, we briefly
review a well-known representation of revision that will help
motivate them.

While there are several representation theorems for belief revision,
the clearest is perhaps the following \cite{Grove,KM92}.  We associate
with each belief set $A$ a set $W_A$ of possible worlds that consists
of those worlds where $A$ is true.  Thus, an agent whose belief set is
$A$ believes that one of the worlds in $W_A$ is the real world.  An
agent that performs belief revision behaves as though in each belief
state $A$ she has a {\em ranking\/}, \ie a total preorder, over all
possible worlds such that the minimal (\ie most plausible) worlds in
the ranking are exactly those in $W_A$.
When revising by $\phi$, the agent
chooses the minimal worlds satisfying $\phi$ in the ranking and
constructs a belief set from them. It is easy to see that this
procedure for belief revision satisfies the AGM postulates.  Moreover,
in \cite{Grove,KM92}, it is shown that any belief revision operator can
be described in terms of such a ranking.

This representation suggests how we can capture belief revision in our
framework. We define $\SR \subseteq \SBCS$ to be the set of
belief change
systems $\Sys =
(\R,\pi,\P)$ that satisfy the conditions REV1--REV4 that we define
below.

Revision assumes that the world does not change during the revision
process. Formally this implies that propositions in $\Phis$ do not
change their truth value along a run, \ie $(\Sys,r,m) \sat p$ if and
only if $(\Sys,r,m+1) \sat p$ for all $p \in \Phis$. This says that
the state of the world is the same with respect to the properties that
the agent reasons about (\ie the propositions in $\Phis$).
\begin{cond}{REV1}
$\pi(r,m)(p) = \pi(r,0)(p)$ for all $p \in \Phis$ and points $(r,m)$.
\end{cond}
Note that REV1 does not necessarily imply that $r_e(m) =
r_e(m+1)$. That is, REV1 allows for a changing environment. The only
restriction is that
the truth value of propositions that describe the environment does not
change.
We return to this issue in \Sref{sec:synthesis}.

The representation of \cite{Grove,KM91} requires the agent to totally
order possible worlds. We put a similar requirement on the agent's
plausibility assessment. Recall that BCS5 says that the agent's
plausibility is induced by a prior $\Pl_a$; REV2 strengthens this
assumption.
\begin{cond}{REV2}
The prior $\Pl_a$ of BCS5 is
ranked; that is, for all
    $A,B \subseteq \R$, either $\Pl_a(A) \le \Pl_a(B)$ or
    $\Pl_a(B) \le \Pl_a(A)$, and $\Pl(A \union B) = \max(\Pl(A), \Pl(B))$.
\end{cond}

The representation of \cite{Grove,KM91} also requires that the agent
considers all truth assignments possible.  We need a similar condition,
except that we want not only that all truth assignments be considered
possible,
but that they have nontrivial plausibility
(\ie
are
more plausible than $\bot$)
as well.

To make this precise, it is helpful to introduce some notation that will
be useful for our later definitions as well.
Given a system $\Sys$ and two sequences
$\phi_1, \ldots, \phi_k$ and  $o_1, \ldots, o_{k'}$ of formulas in
$\Le$,
let $\RE{\phi_1, \ldots, \phi_k ; o_1,
\ldots, o_{k'}}$ consist of all runs $r$ where for each $i$ with $1
\le i \le k$, the formula $\phi_i$ is true at $(r,i)$ and the agent
observes $o_1, \ldots, o_{k'}$.  That is,
$\RE{\phi_0, \ldots, \phi_k ; o_1, \ldots, o_{k'}} = \{r \in\Sys:
(\Sys,r,i) \sat \phi_i, i= 0, \ldots,k, \mbox{ and } r_a(k') =
\<o_1,\ldots,
o_{k'}\>\}$.  We allow either sequence of formulas to be empty, so, for
example,
$\RE{\phi; \cdot }$ consists of all runs for which $\phi$ is true at the
initial state.  (Note that if REV1 holds, this means that $\phi$ is true
in all subsequent states as well.) We use the notation
$\RE{\phi_1,\ldots,\phi_m}$ as an abbreviation for
$\RE{\phi_1,\ldots,\phi_m; \cdot}$.

\begin{cond}{REV3}
If $\phi\in \Le$ is consistent, then $\Pl_a(\RE{\phi }) > \bot$.
\end{cond}

It might seem that REV1--REV3 capture all of the assumptions made by
the representation of \cite{Grove,KM91}.
However, there is another assumption implicit in the way revision is
performed in these representations that we must make explicit in our
representation, because of the way we have distinguished observing
$\phi$ (captured by the formula $\learn(\phi)$) from $\phi$ itself.
Intuitively, when the agent observes $\phi$, she updates
her plausibility assessment by conditioning on $\phi$.  This is
essentially what we can think of the earlier representations as
doing.  However, in our representation, the agent does {\em
not\/} condition on $\phi$, but on the fact that she has observed
$\phi$.  Although we do require that $\phi$ must be true if the agent
observes it (BCS4), the agent may in general gain extra information by
observing $\phi$.

To understand this issue, consider the following example.
Suppose that $\R$ is such that the agent observes $p_1$ at time $(r,m)$
only if  $p_2$ and $q$ are also true at $(r,m)$, and she observes $p_1
\land p_2$ at $(r,m)$ only if $q$ is false. It is easy to construct
a BCS satisfying REV1--REV3 that also satisfies these requirements.  In
this system, after
observing $p_1$, the agent believes $p_2$ and $q$. According to AGM's
postulate R7 (and also KM's postulate U5) the agent must believe $q$
after observing $p_1 \land p_2$.
To see this, note that our assumptions about $\R$ can be
phrased in the AGM language as $p_2 \land q \in K \Rev p_1$ and
$\neg q \in K \Rev(p_1\land p_2)$. Postulate R7 states that
$K\Rev(p_1\land p_2) \subseteq Cl(K \Rev p_1 \union \{ p_2 \})$. Since
$p_2 \in K \Rev p_1$, we have that $Cl(K \Rev p_1 \union \{ p_2 \}) =
K \Rev p_1$. Thus, R7 implies in this case that $q \in K \Rev
(p_1\land p_2)$.
However, in $\R$, the agent believes
(indeed knows) $\neg q$ after observing $p_1\land p_2$.%
\footnote{We stress this does not mean that $p_1 \land p_2$ {\em
implies\/} $\neg q$ in $\R$.  There may well be points in $\R$ at which
$p_1 \land p_2 \land q$ is true.  However, at such points, the agent
would not observe $p_1 \land p_2$, since the agent observes $p_1
\land p_2$ only if $q$ is false.}
Thus,
revision and update both are implicitly assuming that the observation of
$\phi$ does
not provide such additional knowledge.
The following assumption ensures that this is the case for revision (a
more general version will be required for update; see
Section~\ref{sec:update}).
\begin{cond}{REV4}
$\Pl_a(\RE{\phi;  o_1,\ldots,o_m}) \ge \Pl_a(\RE{\psi;
o_1,\ldots,o_m})$ if and only if $\Pl_a(\RE{\phi \land o_1 \land
\ldots\land o_m }) \ge \Pl_a(\RE{\psi \land o_1 \land
\ldots\land o_m })$.
\end{cond}

This assumption
captures the intuition that observing $o_1, \ldots, o_k$
provides no more information than just the fact that $o_1 \land\ldots \land
o_m$ is true. That is, the agent compares the plausibility of $\phi$
and $\psi$ in the same way after conditioning by the observations
$o_1,\ldots,o_m$ as after conditioning by the fact that
$o_1\land\ldots\land o_m$ is true.
It easily follows from REV4 and PRIOR that the agent
believes $\psi$ after observing $o_1 \land \ldots \land o_m$ exactly if
$o_1 \land \ldots \land o_m \land\psi$ was initially considered more
plausible than $o_1 \land \ldots \land o_m \land \neg\psi$.  Thus, the
agent believes $\psi$ after observing $o_1 \land \ldots \land o_m$
exactly if initially, she believed $\psi$ conditional on $o_1 \land
\ldots \land o_m$: the observations provide no extra information
beyond the fact that each of the $o_i$'s are true.

REV4 is quite a strong assumption.  Not only does it say that
observations do not give the agent any additional information (beyond
the fact that they are true), it also says that all consistent
observations can be made (since if $\phi \land o$ is consistent,
we must have $\Pl_a(\RE{\phi;o}) = \Pl_a(\RE{\phi \land o}) > \bot$, by
REV3 and REV4). We might instead consider using a weaker version
of REV4
that
says that, provided an observation can be made, it gives no additional
information.  Formally, this would be captured as
\begin{cond}{REV4$'$}
If $\Pl_a(\RE{\phi; o_1,\ldots,o_m}) > 0$, then
$\Pl_a(\RE{\phi;  o_1,\ldots,o_m}) \ge \Pl_a(\RE{\psi;
o_1,\ldots,o_m})$ if and only if $\Pl_a(\RE{\phi \land o_1 \land
\ldots\land o_m }) \ge \Pl_a(\RE{\psi \land o_1 \land
\ldots\land o_m })$.
\end{cond}
The following examples suggests that
REV4$'$ may be more reasonable in practice than
REV4.  We used REV4 only because it comes closer to the spirit of
the requirement of revision that all observations are possible.

\xam\label{xam:diag-rev}
Consider the system $\Sys_{\diag,1}$ described in
Example~\ref{xam:diag-sys}.
As discussed in \Xref{xam:diag-BCS}, this system can be viewed as a BCS.
Is it a revision system?
It is easy to see that $\Sys_{\diag,1}$ satisfies REV2 and REV3.  It
clearly does
not satisfy REV1, since propositions that describe input/output lines
can change their values from one point to the next.  However, as we are
about to show, a slight variant of $\Sys_{\diag,1}$ does satisfy REV1.
A more fundamental problem is that
$\Sys_{\diag,1}$ does not satisfy REV4.  This is inherent in our assumption
that the agent never directly
observes faults, so that, for example, we have
$\Pl_{\diag,1}(\RE{\cdot; f_1}) = \bot$, while
$\Pl_{\diag,1}(\RE{f_1}) > \bot$.  It does, however, satisfy REV4$'$.

To see how to modify $\Sys_{\diag,1}$ so as to satisfy REV1, recall
that in the diagnosis task, the agent is mainly interested in her
beliefs about faults. Since faults are static in $\Sys_{\diag,1}$, we can
satisfy REV1 if we ignore all propositions except $f_1,\ldots, f_n$.
Let $\Phi'_\diag = \{ f_1, \ldots, f_n \}$ and let $\L'_\diag$ be the
propositional language over $\Phi'_\diag$.  For every observation $o$
made by the agent regarding the value of the lines, there corresponds
a formula in $\L'_\diag$ that characterizes all the fault sets that
are consistent with $o$.  Thus, for every run $r$ in $\Sys_{\diag,1}$, we
can construct a run $r'$ where the agent's local state is a sequence
of formulas in $\L'_\diag$.  Let $\Sys'_\diag$ be the system
consisting of all such runs $r'$.  We can clearly put a plausibility
assignment on these runs so that $\Sys_{\diag,1}$ and $\Sys'_\diag$ are
isomorphic in an obvious sense.  In particular, the agent has the same
beliefs about formulas in $\L'_\diag$ at corresponding points in the
two systems.  More precisely, if $\phi \in \L'_\diag$, then
$(\Sys_\diag',r,m) \sat
\phi$ if and only if $(\Sys_{\diag,1},r,m) \sat \phi$ for all
points $(r,m)$ in $\Sys_{\diag,1}$.
It is easy to verify that $\Sys'_\diag$ satisfies REV1--REV3 and
REV4$'$, although it still does not satisfy REV4.

We are not advocating here here using $\Sys'_\diag$
instead of $\Sys_\diag$---$\Sys_\diag$ seems to us a perfectly
reasonable way of modeling the situation.  Rather, the point is that if
we want a BCS to satisfy properties that validate the AGM postulates, we
must make some strong, and not always natural, assumptions.
\exam

We want to show that a revision operator corresponds to a system in $\SR$
and vice versa. To do so, we need to examine the beliefs of the agent
at each point $(r,m)$.
First we note that if
$(r,m)
\sim_a (r',m')$ then $(\Sys,r,m) \sat \Bel
\phi$ if and only if $(\Sys,r',m') \sat \Bel \phi$; this is a
consequence of the requirement that,
as we have defined interpreted systems,
the agent's plausibility assessment
is a function of her local state. Thus, we think of
the agent's beliefs as a function of her local state.
We use the notation $(\Sys,s_a) \sat \Bel\phi$ as shorthand for
$(\Sys,r,m) \sat \Bel\phi$ for some $(r,m)$ such that $r_a(m) = s_a$.
Let $s_a$ be some local state of
the agent. We define the agent's {\em belief state\/} at $s_a$ as
$$\BEL(\Sys,s_a) = \{\phi \in \Le: (\Sys,s_a) \sat B\phi\}.$$
Since the agent's state is a sequence of observations, the agent's state after
observing $\phi$ is simply $s_a \cdot \phi$, where $\cdot$ is the
append operation. Thus, $\BEL(\Sys, s_a \cdot \phi)$ is the belief
state after observing $\phi$.
We adopt the convention that if the agent can never attain the local
state  $s_a$ in $\Sys$, then $\BEL(\Sys,s_a) = \Le$.
With these definitions, we can compare the agent's belief state before
and after observing $\phi$, that is $\BEL(\Sys,s_a)$ and
$\BEL(\Sys,s_a\cdot\phi)$.

We start by showing that every AGM revision operator can be
represented in $\SR$.

\thm\label{rep1}
Let $\Rev$ be an AGM revision operator and let $K \subseteq \Le$ be a
consistent belief state. Then there is a system $\Sys_{\Rev,K} \in
\SR$
such that
$\BEL(\Sys_{\Rev,K},\<\>) = K$ and
$$
\BEL(\Sys_{\Rev,K},\<\>) \Rev \phi = \BEL(\Sys_{\Rev,K}, \<\phi\>)
$$
for all
$\phi \in \Le$.
\ethm
\prf
See Appendix~\ref{prf:revision}.
\eprf

Thus, Theorem~\ref{rep1} says that we can represent a revision operator
$\Rev$ in the sense that we have a family of systems $\Sys_{\Rev,K}
\in \SR$, one for each consistent belief state $K$, such that $K$ is
the agent's initial belief state in $\Sys_{\Rev,K}$, and for each
formula $\phi$ in $\Le$, the agent's belief state after learning
$\phi$ is $K \Rev \phi$.  Notice that we restrict attention to
consistent belief states $K$.  The AGM postulates allow the agent to
``escape'' from an inconsistent state, so that $K \Rev \phi$ may be
consistent even if $K$ is inconsistent.  We might thus hope to extend
the theorem so that it also applies to the inconsistent belief state,
but this is impossible in our framework.  If
$\False \in
\BEL(\Sys_{\Rev,K},s_a)$
for some state $s_a$, and $r_a(m) = s_a$, then
$\Pl_{(r,m)}(\Omega_{(r,m)}) = \bot$.  Since we update by
conditioning, we must have $\Pl_{(r,m+1)}(\Omega_{(r,m+1)}) = \bot$, so
the agent's belief state will remain inconsistent no matter what she
learns.  Although we could modify our framework to allow the agent to
escape from inconsistent states, we actually consider this to be a
defect in the AGM postulates, not in our framework.  To see why,
suppose that the agent's belief set is inconsistent at $s_a$,
and $r_a(m) = s_a$.  Thus, the agent considers all states in
$\Omega_{(r,m)}$ to be completely implausible (since
$\Pl_{(r,m)}(\Omega_{(r,m)}) = \bot$).  On the other hand, to escape
inconsistency, she must have a plausibility ordering over the worlds
in $\Omega_{(r,m)}$.  These two requirements seem somewhat
inconsistent.%
\footnote{One strength of the AGM framework is that it can deal with an
inconsistent sequence of observations, that is, it can cope with an
observation sequence of the form $\langle p, \neg p, p, \neg p, \ldots
\rangle$.   We stress that being able to cope with such an inconsistent
sequence of observations does not require allowing the agent to escape
from inconsistent belief sets.   These are two orthogonal issues.}

Not surprisingly, this inconsistency creates problems
for  other semantic representations in
the literature. For example, Boutilier's representation theorem
\citeyear{Boutilier92} states that for every revision operator $\Rev$
and belief state $K$, there is a ranking $R$ such that $\psi \in K
\Rev\phi$ if and only if $\psi$ is believed in the minimal
$\phi$-worlds according to $R$. If we examine this theorem, we note
that he does not state that the minimal (\ie most preferred) worlds in
$R$ correspond to the belief state $K$ (in the sense that the minimal
worlds are precisely those where the formulas in $K$ hold); this would
be the analogue of our requiring that $\BEL(\Sys_{\Rev,K},\<\>) = K$.
In fact,
if $K$ is $\vdash_\Le$-consistent, the minimal worlds do correspond
to $K$. However, if $K$ is inconsistent, they cannot, since any
nonempty ranking induces a consistent set of beliefs.
We could state a weaker version of \Tref{rep1} that would correspond
exactly to Boutilier's theorem.  We presented the stronger result (that
does not apply to inconsistent belief states) to bring out what we
believe to be a problem with the AGM postulates.
See \cite{FrH8Full} for further discussion of this issue.

\Tref{rep1} shows that, in a precise sense, we can map AGM revision
operations to $\SR$.
What about the other direction? The next theorem shows that the first
belief change step in systems in $\SR$ satisfies the AGM postulates.
\thm\label{rep2}
Let $\Sys$ be a system in $\SR$.  Then there is an AGM revision
operator $\Rev_{\Sys}$ such that
$$
\BEL(\Sys,\<\>) \Rev_{\Sys} \phi = \BEL(\Sys, \<\phi\>)
$$
for all $\phi \in \Le$.
\ethm
\prf
See Appendix~\ref{prf:revision}.
\eprf

We remark that if we used REV4$'$ instead of REV4, then we would  be
able to prove this result only for those formulas $\phi$ that are observable
(\ie for which $\Pl(\RE{\phi}) > \bot$).

Both Theorems~\ref{rep1} and~\ref{rep2} apply to one-step revision,
starting from the initial (empty) state.  What happens once we allow
iterated revision?  In our framework, observations are taken to be
known, so if the agent makes an inconsistent sequence of observations,
then her belief state will be inconsistent, and (as we observed above)
will remain inconsistent from then on, no matter what she observes.
This creates a problem if we try to get analogues to Theorems~\ref{rep1}
and~\ref{rep2} for iterated revision.
As the following theorem demonstrates, we can already see the problem
if we consider one-step revisions from a state other than the initial
state.
\thm\label{rep4}
Let $\Sys$ be a system in $\SR$ and let $s_a = \<\phi_1, \ldots, \phi_k\>$
be a local state in $\Sys$.
Then there is an AGM revision
operator $\Rev_{\Sys,s_a}$ such that
$$
\BEL(\Sys,s_a) \Rev_{\Sys,s_a} \phi = \BEL(\Sys, s_a \cdot \phi)
$$
for all formulas $\phi \in \Le$ such that $\phi_1 \land \ldots
\land
\phi_k
\land \phi$ is consistent.
\ethm
\prf
See Appendix~\ref{prf:revision}.
\eprf

We cannot do better than this.  If $\phi_1 \land \ldots
\land
\phi_k \land
\phi$ is inconsistent then, because of our requirements that all
observations must be true of the current state of the environment
(BCS4) and that propositions are static (REV1), there cannot be any
global state in $\Sys$ where the agent's local state in $s_a \cdot
\phi$.  Thus, $\BEL(\Sys, s_a \cdot \phi)$ is inconsistent,
contradicting R5.

There is another problem with trying to get an analogue of
Theorem~\ref{rep2} for iterated revision, a problem that seems inherent
in the AGM framework.
Our framework makes a
clear distinction between the agent's {\em epistemic state\/} at a point
$(r,m)$ in $\Sys$, which we can identify with her local state $s_a =
r_a(m)$, and the agent's belief set at $(r,m)$,
$\BEL(\Sys,s_a)$, which is the set of formulas she believes.
In a system in $\SR$, the agent's belief set does not in general
determine how the agent's beliefs will be revised; her epistemic state
does. On the other hand, the AGM postulates assume that revision is a
function of the agent's belief set and observations.
Now suppose we have a system $\Sys$ and two points $(r,m)$ and $(r,m')$
on some run $r \in \Sys$ such that (1) the agent's belief set is the
same at $(r,m)$ and $(r,m')$, that is $\BEL(\Sys,r_a(m)) =
\BEL(\Sys,r_a(m'))$, (2) the agent observes $\phi$ at both $(r,m)$ and
$(r,m')$, (3) $\BEL(\Sys,r_a(m+1))\ne \BEL(\Sys,r_a(m'+1)$.  It is not
hard to construct such a system $\Sys$.  However, there cannot be an
analogue of Theorem~\ref{rep2} for $\Sys$, even if we restrict to
consistent sequences of observations.  For suppose there were a revision
operator $\Rev$ such $\BEL(\Sys,\<\>)) \Rev \phi_1 \Rev \cdots \Rev \phi_k
= \BEL(\Sys,\<\phi_1, \ldots, \phi_k\>)$ for all $\phi_1, \ldots,
\phi_k$ such that $\phi_1 \land \ldots \land \phi_k$ is consistent.
Then we would have
$\BEL(\Sys,r_a(m+1)) = \BEL(\Sys,r_a(m)) \Rev \phi = \BEL(\Sys,r_a(m'))
\Rev \phi = \BEL(\Sys,r_a(m'+1))$, contradicting our assumption.

The culprit here is the assumption that revision depends only on the
agent's belief set.  To see why this is an unreasonable assumption,
consider a situation where at time 0 the agent
believes both $p$ and $q$, but her belief in $q$ is stronger than her
belief in $p$ (\ie the plausibility of $q$ is greater than that of
$p$).  We can well imagine that after observing $\neg p \lor
\neg q$ at time 1, she would believe $\neg p$ and $q$.  However,
if she first observed $p$ at time 1 and then $\neg p \lor \neg q$ at
time 2, she would believe $p$ and $\neg q$, because, as a result of
observing $p$, she would assign $p$
greater plausibility than $q$.
Note, however, that the AGM postulates dictate that after an
observation that is already believed, the agent does not change her
beliefs. Thus, the AGM
setup would force the agent to have the same beliefs after learning
$\neg p \lor \neg q$ in both situations.

There has been a great deal of work on the problem of
{\em iterated belief revision\/}
\cite{Boutilier:Iterated,Darwiche94Full,FreundLehmann,Lehmann95,Levi88,Nayak:1994,Williams94}).
Much of the recent work moves away from the assumption that belief
revision depends solely on the agent's belief set.
For example the approaches of Boutilier
\citeyear{Boutilier:Iterated} and Darwiche and Pearl \citeyear{Darwiche94Full}
define revision operators that map (rankings $\cross$ formulas) to
rankings.
Because our framework makes such a clear distinction between epistemic
states and belief states, it gives us a natural way of
maintaining the spirit of the AGM postulates while assuming that
revision is a function of epistemic states.
Rather than taking $\Rev$ to be a function from (belief states
$\cross$ formulas) to belief states, we take it $\Rev$ to be a function
from (epistemic states $\cross$ formulas) to epistemic states.

This leaves open the question of how to represent epistemic
states. Boutilier and Darwiche and Pearl use rankings to represent
epistemic states. In our framework, we represent epistemic states by
local states in interpreted systems. That is, a pair $(\Sys,s_a)$
denotes the agent's state in an interpreted system, and
the pair determines
the agent's relevant epistemic attitudes, such as her
beliefs, how her beliefs changed given particular observations, her
plausibility assessment over runs, and so on.
When the system is understood, we simply
use $s_a$ as a
shorthand representation of an epistemic state.

We can
easily modify the AGM postulates to deal with such revision operators
on epistemic states. We start by assuming that there is a set of
epistemic states and a function $\BEL(\cdot)$ that maps epistemic states
to belief states.
We then have analogues to each of the AGM
postulates, obtained by replacing each
belief set by the beliefs in the corresponding epistemic state.  For
example, we have
\begin{description}\denselist
\item[(R1$'$)] $E \Rev \phi$ is an epistemic state
\item[(R2$'$)] $\phi \in \BEL(E\Rev\phi)$
\item[(R3$'$)] $\BEL(E \Rev\phi) \subseteq Cl(\BEL(E) \cup \{\phi\})$
\end{description}
and so on, with the obvious transformation.%
\footnote{The only problematic postulate is R6.  The question is whether
R6$'$ should be ``If $\vdash_\Le \phi \dimp \psi$ then
$\BEL(E\Rev\phi) = \BEL(E\Rev\psi)$'' or ``If $\vdash_\Le \phi \dimp
\psi$ then $E\Rev\phi = E\Rev\psi$''.  Dealing with either version is
straightforward.  For definiteness, we adopt the first alternative
here.}

We can get strong representation theorems if we work at the level of
epistemic states.
Given a language $\Le$ (with an associated consequence relation
$\vdash_{\Le}$), let $\E_{\Le}$ consist of all
finite sequences
of formulas in $\Le$.
Note that we allow $\E_{\Le}$ to include sequences of formulas
whose conjunction is inconsistent.
We define revision in $\E_{\Le}$
in the obvious way: if $E\in \E_{\Le}$, then $E \Rev \phi = E
\cdot\phi$.
\thm\label{rep6}
Let $\Sys$ be a system in $\SR$ whose local states are $\E_{\Le}$.
There is a function
$\BEL_{\Sys}$
that maps epistemic states to belief states such that
\begin{itemize}\denselist
\item if $s_a$ is a local state of the agent in $\Sys$, then
$\BEL(\Sys,s_a) = \BEL_{\Sys}(s_a)$, and
\item $(\Rev,\BEL_{\Sys})$ satisfies R1$'$--R8$'$.
\end{itemize}
\ethm
\prf
Roughly speaking, we define $\BEL_\Sys(s_a)
= \BEL(\Sys,s_a)$ when $s_a$ is a local state in $\Sys$. If $s_a$
is not in $\Sys$, then we set $\BEL_\Sys(s_a) = \BEL(\Sys,s')$,
where $s'$ is the longest consistent suffix of $s_a$.
See Appendix~\ref{prf:revision} for details.
\eprf

Notice that, by definition, we have $\BEL_{\Sys}(\Sys,\<\> \Rev_{\Sys}
\phi_1 \Rev_{\Sys} \ldots \Rev_{\Sys} \phi_k) =
\BEL_{\Sys}(\Sys,\<\phi_1, \ldots, \phi_k\>)$, so, at the level of
epistemic states, we get an analogue to Theorem~\ref{rep2}.
We remark that to ensure that R5$'$ holds for $(\Rev,\BEL_{\Sys})$, we
need to define $\BEL_{\Sys}(E)$ appropriately for sequences $E \in
\E_{\Sys}$ whose conjunction  is inconsistent.

Theorem~\ref{rep6} shows that any system in $\SR$ corresponds to a
revision operator over epistemic states that satisfies the generalized
AGM postulates.
We would hope that the converse also holds. Unfortunately, this is not
quite the case. There are revision operators on epistemic states that
satisfy the generalized AGM postulates but do not correspond to a
system in $\SR$. This is because systems
in $\SR$ satisfy an additional postulate:
\begin{description}\denselist
\item[(R9$'$)] If $\not\vdash_\Le \neg(\phi \land \psi)$ then $\BEL(E \Rev
\phi \Rev\psi) = \BEL(E\Rev \phi\land\psi)$.
\end{description}

We show that R9$'$ is sound in $\SR$ by proving the
following strengthening of Theorem~\ref{rep6}.
\pro\label{pro:R9'}
Let $\Sys$ be a system in $\SR$ whose local states are $\E_{\Le}$.
There is a function
$\BEL_{\Sys}$
that maps epistemic states to belief states such that
\begin{itemize}\denselist
\item if $s_a$ is a local state of the agent in $\Sys$, then
$\BEL(\Sys,s_a) = \BEL_{\Sys}(s_a)$, and
\item $(\Rev,\BEL_{\Sys})$ satisfies R1$'$--R9$'$.
\end{itemize}
\epro
\prf
We show that the function $\BEL_\Sys$ defined in the proof of
Theorem~\ref{rep6} satisfies R9$'$.
See Appendix~\ref{prf:revision} for details. \eprf

We can prove the converse to Proposition~\ref{pro:R9'}: a revision
system on epistemic states that satisfies the generalized AGM
postulates {\em and R9$'$\/} does correspond to a system in $\SR$.
\thm\label{rep5}
Given a function $\BEL_{\Le}$ mapping epistemic states in $\E_{\Le}$ to
belief sets over $\Le$ such that $\BEL_{\Le}(\<\>)$ is consistent and
$(\BEL_{\Le},\Rev)$ satisfies R1$'$--R9$'$, there is a system
$\Sys \in \SR$ whose local states are in $\E_{\Le}$ such that
$\BEL_{\Le}(s_a) = \BEL(s_a)$ for each local state $s_a$in
$\Sys$.
\ethm
\prf
According to Theorem~\ref{rep1}, there is a system
$\Sys$ such that $\BEL(\Sys,\<\>) = \BEL_{\Le}(\<\>)$ and
$\BEL(\Sys,\<\phi\>) = \BEL_{\Le}(\<\phi\>)$ for all $\phi\in\Le$.
We show that $\BEL(\Sys,s_a) = \BEL_{\Le}(s_a)$ for local states $s_a$
in $\Sys$.  See Appendix~\ref{prf:revision}.
\eprf

Notice that, by definition, for the system $\Sys$ of Theorem~\ref{rep5},
we have  $\BEL(\<\> \Rev \phi_1 \Rev \ldots \Rev \phi_k) =
\BEL(\<\phi_1, \ldots, \phi_k\>)$ as long as $\phi_1 \land \ldots \land
\phi_k$ is consistent.

\section{Capturing Update}
\label{sec:update}
\label{SEC:UPDATE}

Update tries to capture the intuition that there is a preference for
runs where all the observations made are true, and where changes
{f}rom one point to the next along the run are minimized.

We start by reviewing Katsuno and Mendelzon's semantic representation
of update. To characterize an agent beliefs, Katsuno and Mendelzon
consider the set of ``worlds'' the agent considers possible. In their
representation, they associate a world with a truth assignment to the
primitive propositions. (In our terminology, we can think of a world
as an environment state.)  To capture the notion of ``minimal change
from world to world'', Katsuno and Mendelzon use a {\em distance
function\/} $d$ on worlds.  Given two worlds $w$ and $w'$, $d(w,w')$
measures the distance between them. Intuitively, the larger the
distance, the larger the change required to get from world $w$ to
$w'$. (Note that that distances are not necessarily symmetric, that
is, it might require a smaller change to get from $w$ to $w'$, than
from $w'$ to $w$.)
Distances might be incomparable, so we require that $d$ map pairs of
worlds into a {\em partially ordered\/} domain with a unique minimal
element $0$ and that $d(w,w') = 0$ if and only if $w = w'$.

Katsuno and Mendelzon show that there is a 
close relationship between
update operators and distance functions. To make this relationship
precise, we need to introduce some definitions. An {\em update
structure\/} is a tuple $U = (W,d,\pi)$, where $W$ is a finite set
of worlds, $d$ is a distance function on $W$, and $\pi$ is a mapping
from worlds to truth assignments for $\Le$ such that
\begin{itemize}\denselist
\item $\pi(w)$ is $\vdash_\Le$ consistent,
\item if $\not\vdash_\Le\neg\phi$, then there is some $w \in W$ with
$\pi(w)(\phi) = $ {\bf true}, and
\item if $w \neq w'$ then $\pi(w) \neq \pi(w')$ for all $w,w' \in W$.
\end{itemize}
Given an update structure $U = (W,d,\pi)$, we define
$\intension{ \phi }_U = \{ w : \pi(w)(\phi) = $ {\bf
true}$\}$. Katsuno and Mendelzon use update structures as semantic
representations of update operators.
Given an update structure $U = (W,d,\pi)$ and sets $A, B \subseteq
W$, Katsuno and Mendelzon define $\min_U(A,B)$ to be the set of worlds
in $B$ that are closest to worlds in
$A$, according to $d$. Formally, $\min_U(A,B) = \{ w \in B : \exists
w_0 \in A \forall w'\in B\, d(w_0,w') \not< d(w_0,w) \}$.

\thm\label{thm:KM}{\rm \cite{KM92}}
A belief change operator $\Upd$ satisfies U1--U8 if and only if
there is an update structure $U = (W,\pi,d)$ such that
$$
\intension{\phi \Upd \psi}_U = {\min}_U(\intension{\phi}_U,\intension{\psi}_U).
$$
\ethm
Thus the worlds the agent believes possible after updating with
$\psi$ are these worlds that are closest to
some world considered possible before learning $\psi$.

Katsuno and Mendelzon's account of update is ``static'' in the sense
that it describes a single belief change. Nevertheless, there is
a clear intuition that each world $w' \in \intension{\phi\Upd
\psi}_U$ is the result of considering a minimal change from some world
$w \in \intension{\phi}_U$. However, in Katsuno and Mendelzon's
representation, we do not keep track of the worlds 
that ``lead to'' the worlds  in the current belief set.

We now try to capture behavior similar to Katsuno and Mendelzon's
semantics in our framework. We define systems where each run describes
the sequence of changes, so that the most plausible runs, given a set
of observations, correspond the worlds that define the belief set in
Katsuno and Mendelzon's semantics. More precisely, given a sequence of
observations $\psi_1,\ldots,\psi_n$, each world in
$\intension{\phi\Upd\psi_1\Upd\ldots\Upd\psi_n}_U$ can be ``traced''
back through a series of minimal changes to a world in
$\intension{\phi}_U$. In our model, each such trace corresponds to one
of the most plausible runs, where the environment state at time $m$ is
the $m$th world in the trace. 
We can capture this intuition by using a family of priors with a
particular form. 

We start with some preliminary definitions.  Let $\Sys$ be a BCS, and
let $s_0,\ldots,s_n$ be a set of environment states in $\Sys$.  We
define $\RP{s_0,\ldots,s_n}$ as the set of runs where $r_e(i) = s_i$
for all $0 \le i \le n$. Thus, $\RP{s_0,\ldots,s_n}$ describes a set
of runs that share a common prefix of environment states. A prior
plausibility space $\P_a = (\R,\Pl_a)$ is {\em consistent\/} with a
distance measure $d$ if the following holds:
\begin{quote}
$\Pl_a(\RP{s_0,\ldots,s_n}) < \Pl_a(\RP{s_0', \ldots,
s_n'})$ if and only if there is some $j < n$ such that $s_k = s'_k$
for all $0 \le k \le j$, $s_{j+1} \neq s'_{j+1}$, and $d(s_{j},s_{j+1})
< d(s_{j},s'_{j+1})$.
\end{quote}
Intuitively,
we compare events of the form
$\RP{s_0,\ldots,s_n}$ using a lexicographic ordering based on
$d$. Notice that this ordering focuses on the {\em first\/} point of
difference.
Runs with a smaller change at this point are
preferred, even if later there are abnormal changes. This point is
emphasized in the borrowed car example
below.

$\Pl_a$ is {\em prefix-defined\/} if the plausibility of an event is
uniquely defined by the plausibility of run-prefixes that are contained in
it, so that
\begin{quote}
$\Pl_a(\RE{\phi_0,\ldots,\phi_m}) \ge
\Pl_a(\RE{\psi_0,\ldots,\psi_m})$ if and
only if for all $\RP{s_0,\ldots,s_m} \subseteq
\RE{\psi_0,\ldots,\psi_m} - \RE{\phi_0,\ldots,\phi_m}$
there is
some $\RP{s'_0,\ldots,s'_m} \subseteq
\RE{\phi_0,\ldots,\phi_m}$ such that $\Pl_a(\RP{s'_0,\ldots,s'_m}) >
\Pl_a(\RP{s_0,\ldots,s_m})$.
\end{quote}
Roughly speaking, this requirement states that we compare events by
properties of dominance. This property is similar to one satisfied by
the plausibility measures that we get from preference ordering using
the construction of Proposition~\ref{pro:prec}.

We define the set $\SU$ to consist of BCSs $\Sys = (\R,\pi,\P)$ that
satisfy the following four requirements UPD1--UPD4.  UPD1 says that
there are only finitely many possible truth assignments, and that there
is a one-to-one map between environment states and truth assignments.
\begin{cond}{UPD1}
The set $\Phis$ of propositions (of BCS1) is finite and $\pi$ is such
that for all environment states $s$, $s'$, if $s \ne s'$, then there is a
formula $\phi \in \Le$ such that $s \sat \phi$ and $s' \sat \neg \phi$.
\end{cond}

UPD2--UPD4 are analogues to REV2--REV4. Like REV2, UPD2 puts constraints
on the form of the prior, but now we consider lexicographic priors of
the form described above.
\begin{cond}{UPD2}
The prior of BCS5 is prefix defined and consistent with some distance measure.
\end{cond}

Recall that REV3 requires only
that all truth assignments initially have nontrivial plausibility.  In
the case of
revision, the truth assignment does not change over time, since we are
dealing with static propositions.  In the case of update, the truth
assignment may change over time, so UPD3 requires that all consistent
sequences of truth assignments have nontrivial plausibility.
\begin{cond}{UPD3}
If $\phi_i\in\Le$, $i = 0,\ldots,k$, are consistent
formulas, then
$\Pl(\RE{\phi_0, \ldots,\phi_k}) > \bot$.
\end{cond}

Finally, like REV4, UPD4 requires that the agent gain no information
from her observations beyond the fact that they are true.
\begin{cond}{UPD4}
$\Pl_a(\RE{\phi_0,\ldots, \phi_{k+1};  o_1,\ldots,o_k}) \ge
\Pl_a(\RE{\psi_0,\ldots, \psi_{m+1};  o_1,\ldots,o_m})$ if and only
if
$\Pl_a(\RE{\phi_0, \phi_1 \land o_1, \ldots, \phi_m \land o_m,
\phi_{m+1}}) \ge
\Pl_a(\RE{\psi_0, \psi_1 \land o_1, \ldots, \psi_m \land o_m,
\psi_{m+1}})$
\end{cond}
We remark that in the presence of REV1, UPD4 is equivalent to REV4.
We might consider generalized versions of UPD4, where the
two sequences of formulas can have arbitrary relative lengths; this
version suffices for our purposes.
We can also define an
analogue UPD4$'$ in the spirit of REV4$'$, which applies only if
$\Pl(\RE{\phi_0, \ldots, \phi_{m+1};  o_1,\ldots,o_m}) > \bot$.

We now show that $\SU$ corresponds to (KM)
update. Recall that Katsuno and Mendelzon define an update operator as
mapping a pair of formulas $(\mu,\phi)$, where $\mu$ describes the
agent's beliefs and $\phi$ describes the observation, to a new formula
$\mu\Upd \phi$ that describes the agent's new beliefs. However, as we
discussed in Section~\ref{sec:review}, when $\Phis$ is finite, we can
also treat $\Upd$
mapping a belief state and a formula to a new belief state. Also
recall that $\BEL(\Sys,s_a)$ is the agent's belief set when
her local state is  $s_a$.
\thm
\label{thm:update-rep}
A belief change operator $\Upd$ satisfies U1--U8 if and only if there
is a system $\Sys \in \SU$ such that
$$\BEL(\Sys,s_a) \Upd \psi = \BEL(\Sys, s_a \cdot \psi)$$
for all epistemic states $s_a$ and formulas $\psi \in \Le$.
\ethm
\prf
Roughly speaking, we show that any system in $\SU$ corresponds to a
Katsuno and Mendelzon update structure. Suppose that $\Sys =\in \SU$
is such that the set of environment states is $\S_e$ and the prior of
BCS5 is consistent with distance function $d$. We define an update
structure $U_\Sys$. We then show that belief change in $\Sys$
corresponds to belief change in $U_\Sys$ in the sense of
Theorem~\ref{thm:KM}. Since Theorem~\ref{thm:KM} states that any
belief change operation defined by an update structure satisfies
U1--U8, this will suffice to prove the ``if'' direction of the
theorem. To prove the ``only if'' direction of the theorem, we show
that that for any update structure $U$, there is a system $\Sys \in
\SU$ such that $U_\Sys = U$.

See Appendix~\ref{prf:update} for details.
\eprf

This result immediately generalizes to sequences of updates.
\cor
A belief change operator $\Upd$ satisfies U1--U8 if and only if there
is a system
$\Sys_\Upd \in \SU$ such that for all $\psi_1, \ldots, \psi_k
\in \Le$, we have
$$
\BEL(\Sys_\Upd,s_a) \Upd \psi_1 \Upd \ldots \Upd \psi_k =
\BEL(\Sys_\Upd, s_a \cdot \psi_1 \cdot \ldots \cdot \psi_k).
$$
\ecor
These results show
that for update, unlike revision, the systems we consider are
such that the belief state does determine the result of the update,
    \ie if $\BEL(\Sys,s_a) = \BEL(\Sys,s_a')$, then for any $\phi$ we
    get that $\BEL(\Sys,s_a\cdot\phi) = \BEL(\Sys,s_a'\cdot\phi)$.
Roughly speaking, the reason is that the distance measure
that determines the prior does not change over time.
While this allows us to get an elegant representation theorem, it also
causes problems for the applicability of update, as we shall see below.

Note that, since the world is allowed to change, there is no problem
if we update by a sequence $\psi_1, \ldots, \psi_k$ of consistent
formulas such that $\psi_1 \land \ldots \land \psi_k$ is inconsistent.
There is no requirement that the formulas $\psi_1, \ldots, \psi_k$ be
true simultaneously. All that matters is that $\psi_i$ is true at time
$i$.  Also note that an update by an inconsistent formula does not
pose a problem for our framework.
It follows from postulates U1 and U2 that once the agent
learns an inconsistent formula (\ie $\False$),
she believes $\False$ from then on.

How reasonable is the notion of update?  As the discussion of UPD2
above suggests, it has a preference for deferring abnormal events.
This makes it quite similar to Shoham's {\em chronological
ignorance\/} \citeyear{Shoham88},
and it suffers from some of the same
problems. Consider the following story, that we call the {\em
borrowed-car example}.%
\footnote{This example is based on Kautz's stolen car story
\citeyear{Kautz}, and is due to Boutilier, who independently observed
this problem [private communication, 1993].}  At time 1, the agent
parks her car in front of her house with a full fuel tank. At time 2,
she is in her house.  At time 3, she returns outside to find the car
still parked where she left it.  Since the agent does not observe the
car while she is inside the house, there is no reason for her to
revise her beliefs regarding the car's location.  Since she finds it
parked at time 3, she still has no reason to change her beliefs.  Now,
what should the agent believe when, at time 4, she notices that the
fuel tank is no longer full?  The agent may want to consider a number
of possible explanations for her time-4 observation, depending on what
she considers to be the most likely sequence(s) of events between time
1 and time 4.  For example, if she has had previous gas leaks, then
she may consider leakage to be the most plausible explanation.  On the
other hand, if her spouse also has the car keys, she may consider it
possible that he used the car in her absence.  Update, however,
prefers to defer abnormalities, so it will conclude that the fuel must
have disappeared, for inexplicable reasons, between times~3 and~4. To
see this, note that runs where the car has been taken on a ride have
an abnormality at time~2, while runs where the car did not move at
time~2 but the fuel suddenly disappeared, have their first abnormality
at time~4, and thus are preferred!

Suppose we formalize the example using
propositions such as {\it car-parked-outside, fuel-tank-full}, etc.
Let the agent's belief set at time $i$ be $\mu_i$, $i = 1, \ldots, 4$.
Notice that $\mu_1$ includes the belief that the car is parked in front
of the house with a full fuel tank.  (That is, $\vdash_\Le \mu_1 \rimp
\mbox{{\it fuel-tank-full}} \land \mbox{{\it car-parked-outside}}$.)
At time 2 the agent
makes no observations since she is in her house, so $\mu_2 = \mu_1 \Upd
\true = \mu_1$ by U2.  At time 3 the agent observes the car outside her
house, so $\mu_3 = \mu_2 \Upd \mbox{{\it car-parked-outside}} =
\mu_1$, again by U2.  Finally, $\mu_4 = \mu_3 \Upd \neg \mbox{{\it
fuel-tank-full}}$. The observation of $\neg${\it fuel-tank-full} at
time~4 must be explained by some means. 
In our semantics, the answer is
clear. The most plausible runs are these where the car was parked
until time $3$, and somewhere between time $3$ and $4$ some change 
occurred.

Is this counterintuitive conclusion an artifact of our representation?
To some extent it is. This issue cannot be formally addressed within
Katsuno and Mendelzon's semantic framework, since that framework does
not provide an account of sequences of changes.  Moreover, one might
argue that within out framework there might be other families of
priors that satisfy U1--U8, which will offer alternative
explanations of the surprising observation at time 4. Nevertheless, we claim
that our semantics captures, in what we believe to be the most straightforward
way, the intuition embedded in the Katsuno and Mendelzon's
representation. In particular, condition UPD2, which enforces the
delay of abnormal events, was needed in order to capture the
``pointwise'' nature of the update.
It would be interesting to know whether there is a natural way of
capturing update in our framework that does not suffer from these problems.

Does this way of capturing update semantically ever lead to reasonable results?
Of course, that
depends on how we interpret ``reasonable''.  We briefly consider one
approach here.

In a world $w$, the agent has some beliefs that are described by, say,
the formula $\phi$.  These beliefs may or may not be {\em correct\/}
(where we say a belief $\phi$ is correct in a world $w$ if $\phi$ is
true of $w$).  Suppose something happens and the world changes to
$w'$.  As a result of the agent's observations, she has some new
beliefs, described by $\phi'$.  Again, there is no reason to believe
that $\phi'$ is correct.  Indeed, it may be quite unreasonable to
expect $\phi'$ to be correct, even if $\phi$ is correct.  Consider the
borrowed-car example.  Suppose that while the agent was sitting inside
the house, the car was, in fact, taken for a ride.  Nevertheless, the
most reasonable belief for the agent to hold when she observes that
the car is still in the parked after she leaves the house is that it
was there all along.

The problem here is that the information the agent obtains
at times~2 and~3 is insufficient to determine what happened.
We cannot expect all the agent's beliefs to be correct at this point.
On the other hand, if she does obtain sufficient information about the
change and her beliefs were initially correct, then it seems reasonable
to expect that her new beliefs will be correct.
But what counts as {\em sufficient\/} information?

We say that $\phi$ provides {\em sufficient information\/} about the
change from $w$ to $w'$ if there is no world $w''$ satisfying $\phi$
such that $d(w,w'') < d(w,w')$. In other words, $\phi$ is sufficient
information if, after observing $\phi$ in world $w$, the agent will
consider the real world ($w'$) one of the most likely worlds.  Note
that this definition is monotonic, in that if $\phi$ is sufficient
information about the change, then so is any formula $\psi$ that
implies $\phi$ (as long as it holds at $w'$).  Moreover, this
definition depends on the agent's distance function $d$.  What
constitutes sufficient information for one agent might not for
another.  We would hope that the function $d$ is realistic in the
sense that the worlds judged closest according to $d$ really are the
most likely to occur.

We can now show that update has the property that if the agent has
correct beliefs and receives sufficient information about a change,
then she will continue to have correct beliefs.
\thm
\label{thm:update}
Let $\Sys \in \SU$. If the agent's beliefs at $(r,m)$ are correct and
$\obs{r,m}$ provides sufficient information about the change from
$r_e(m)$ to $r_e(m+1)$, then the agent's beliefs at $(r,m+1)$ are
correct.
\ethm
\prf
Straightforward; left to the reader.
\eprf

As we observed earlier, we cannot expect the agent to always have
correct beliefs.  Nevertheless,
we might hope that if
the agent does (eventually) receive sufficiently detailed information,
then she should realize that her beliefs were incorrect.  But this is
precisely what does {\em not\/} happen in the borrowed-car example.
Intuitively, once the agent observes that the fuel tank is not full,
this should be sufficient information to eliminate the possibility
that the car remained in the parking lot.  However, it is not. Roughly
speaking, this is because
update focuses only on the current state of the world, and thus
cannot go back and revise beliefs about the past.

The problem here is again due to the fact that belief update is
determined only by the agent's belief state and not her epistemic
state. Thus, update can only take into account the agent's current
beliefs and not other information, such as the sequence of
observations that led to these beliefs.  In our example, if we limit
our attention to beliefs about the car's whereabouts and the fuel
tank, then since the agent has the same belief state at time~1 and~3,
she must change her beliefs in the same manner at both times.  This
implies that the observation the fuel tank is not full at time~4
cannot be sufficient information about the past, since a fuel leak might
be the most plausible explanation of missing fuel at time~2.%
\footnote{In this example the usual intuition is that, given the
observation that
the tank is not full, the agent should revise her belief in some
manner instead of performing update. This immediately raises the
question of how the agent knows what the right belief change operation
should be here.  We return to this issue below.}

Our discussion of update shows that update is guaranteed to be safe
only in situations where there is always enough information to
characterize the change that has occurred.  While this may be a plausible
assumption in database applications, it seems somewhat less reasonable
in AI examples, particularly in cases involving reasoning about action.%
\footnote{Similar observations were independently made by Boutilier
    \citeyear{Boutilier94}, although his representation is quite different
    from ours.}

\section{Synthesis}\label{sec:synthesis}

In previous sections we analyzed belief revision and belief update
separately. We provided representation theorems for both notions and
discussed issues specific to each notion. In this section, we
try to identify some common themes and points of difference.

Katsuno and Mendelzon \citeyear{KM91} focused on the following three
differences between AGM revision and KM update:
\begin{enumerate}
\item Revision deals with static propositions, while update allows
propositions that are not static.

\item \label{pt:inconsistent}
Revision and update treat inconsistent belief states differently.
Revision allows an agent to ``recover'' from an inconsistent state
after observing a consistent formula. Update dictates that once the
agent has inconsistent beliefs, she will continue to have inconsistent
beliefs.
As we noted above, it seems that revision's ability
to recover from an inconsistent belief set leads to several technical
anomalies
in iterated revision.

\item Revision considers only total preorders, while
    update allows partial preorders.
\end{enumerate}

Our framework suggests a different
approach to categorizing the differences between revision and update
(and other approaches to belief change): focusing on the restrictions
that have to be added to basic BCSs to obtain systems in $\SR$ and
$\SU$, respectively.
In particular, we
focus on three aspects of a system:
\begin{itemize}\denselist
\item How does the environment state change?
\item How does the agent form her initial beliefs? What regularities
appear in the agent's beliefs at the initial state?
\item How does the agent change her beliefs?
\end{itemize}

\begin{table}
\begin{center}
\begin{tabular}{|l||c|c|}
\hline
Restriction on                 & Revision      & Update\\
\hline
\hline
\rule[-0.7em]{0cm}{2em}
%%%Environment changes & \parbox[c]{1.8in}{\centerline{No change}\\
%%%\centerline{(Static propositions)}}         & All possible sequences \\
Environment changes & \parbox[c]{1.8in}{\begin{center}No change\\
(Static propositions)\end{center}}   & All possible sequences \\
\hline
\rule[-0.7em]{0cm}{2em}
Initial plausibility  & Total preorder  & Lexicographic \\
\hline
\rule[-0.7em]{0cm}{2em}
Belief change        & Conditioning  & Conditioning \\
\hline
\end{tabular}
\end{center}
\caption{A summary of the restrictions we impose to capture
revision and update.}
\label{tab:summary}
\end{table}

Table~\ref{tab:summary} summarizes the answers to these questions for
revision and update; it highlights the different restrictions
imposed by each.
Revision puts a severe restriction on changes of the
environment (more precisely, on how we describe the environment in the
language) and a
rather mild restriction on the agent's prior beliefs
(they must form a
total preorder). On the other hand, update allows all sequences of
environment states, but requires the agent's prior beliefs to have a
specific form.
These formal properties match the intuitive description of revision
and update given in \cite{agm:85,KM92}. However, the explicit
representation of time in our framework allows us to make these
intuitions precise. Moreover, our framework makes explicit other
assumptions made by revision and update. For example, the
lexicographic nature of update is not immediately evident from the
presentation in \cite{KM92}.

The key point to notice in this table is that belief change in both
revision and update is done by conditioning. This observation, and the
naturalness of conditioning as a notion of change, support our claim
that conditioning should be adopted as semantic foundations for
minimal change.

How significant are the differences between revision and update?  We
claim that some of these differences are a result of different ways
of modeling the
same underlying process. Recall that in the
introduction we noted that the restriction to static propositions
is not such
a serious limitation of belief revision, since we can
always convert a dynamic proposition to a static one by adding timestamps.
More precisely, we can replace a proposition $p$ by a family
of propositions $p^m$ that stand for ``$p$ is true at time
$m$''. This  makes it possible to use revision to reason about a changing
world.
We now show how revision and update can be related under this viewpoint.

To make this discussion precise,  we need to introduce some formal definitions.
Let $\Sys = (\R,\pi, \P)$ be a BCS.
We ``statify''
$\Sys$ into a system $\Sys^* = (\R^*,\pi^*,\P^*)$
by replacing the underlying language with static propositions.

Let $\Phi^*_e = \{ p^m
: p \in \Phi_e, m \in \natnum\}$ be a set of timestamped
propositions and let $\Ls^*$ be the logical language based on these
propositions. We can easily ``timestamp'' every formula in $\L$. We
define $\timestamp(\phi,m)$ recursively as follows. The base case is
$\timestamp(p,m) = p^m$ for $p \in \Phi_e$. For standard logical
connectives, we simply apply the transformation recursively, for
example $\timestamp(\phi\land\psi) =
\timestamp(\phi,m)\land\timestamp(\psi,m)$.

Next, we define the set of runs in the ``statified'' system. For each run $r
\in \R$, we define a $r^*$ in $\R^*$ as follows. The environment
states in $r^*$ are defined to be the whole sequence of environment
states in $r$, that is,  $r^*_e(m) = r_e$. If $r_a(m) =
\<\obs{r,1},\ldots,\obs{r,m}\>$, we define $r_a(m) = \<
\timestamp(\obs{r,1},1), \ldots, \timestamp(\obs{r,m},m) \>$. We
define the interpretation $\pi^*$ in the
obvious way:
$\pi^*(r^*,m)(p^{m'}) =$ {\bf true} if and only if $\pi(r,m')(p) =$ {\bf true}
and $\pi^*(r^*,m)(\learn(\phi)) = $ {\bf true}  if and only if $\obs{r^*,m} =
\phi$.

Finally, we need to define the prior plausibility $\Pl_a^*$. We define
this prior to be isomorphic to $\Pl_a$ under the transformation
$r^* \mapsto r$. That is, for each set of runs $R^*\subseteq \R^*$, we
define $\Pl^*_a(R^*) =\Pl_a ( \{ r \in \R : r^* \in R^* \} )$.

It is clear that the two systems $\Sys$ and $\Sys^*$ describe the
same underlying process.
Perhaps the most significant difference is that
the environment state in a run of $\Sys^*$ encodes the future of the
run.  This was necessary so that the environment state could determine
the truth of all propositions of the form $p^m$, so as to satisfy BCS1.
Without this requirement, we could have
simply changed $\pi$ and left $\R$ and $\P$ unchanged.

Because different base languages are used in $\Sys$ and $\Sys^*$, the
agent has different beliefs in the two systems.
It is easy to show that, for all $\phi \in \Ls$, we have
$(\Sys,r,m) \sat
\Bel \phi$ iff $(\Sys^*,r^*,m) \sat \Bel (\timestamp(\phi,m))$. However,
at $(r^*,m)$ the agent also has beliefs about propositions
that describe past and future times. Thus, the set of beliefs of the
agent in $\Sys^*$
can be viewed as
a superset of her beliefs in $\Sys$ at the
corresponding points.

The following result makes precise the relationship between $\Sys$ and
$\Sys^*$ in terms of the properties we have been considering.
\pro
Let $\Sys$ be a BCS and let $\Sys^*$ the transformed system defined
above. Then
\begin{itemize}\denselist
\item $\Sys^*$ is a BCS, that is, it satisfies BCS1--BCS5.
\item $\Sys^*$  satisfies REV1.
\item If $\Sys$ satisfies UPD3, then $\Sys^*$ satisfies REV3.
\item If $\Sys$ satisfies UPD4, then $\Sys^*$ satisfies REV4$'$.
\end{itemize}
\epro
\prf
Straightforward; left to the reader.
\eprf

Thus, if $\Sys$ is a BCS, so is $\Sys^*$. Moreover, if $\Sys \in \SU$,
then $\Sys^*$ satisfies all but two of the requirements for $\SR$.
First, $\Sys^*$ does not necessarily satisfy REV2, since
the prior of systems in $\SU$ is, in general, not ranked. Second,
$\Sys^*$ satisfies REV4$'$, the weaker version of REV4. The reason for
this is that runs $\Sys^*$ do not allow all sequences of possible
observations. Remember that in the language of $\Ls^*$, the agent can
observe the proposition $p^2$ (\ie that $p$ is true at time $2$) at time
$1$. However, in the original system, the agent only observes
properties of the {\em current\/} time. Thus, $\obs{r^*,m}$ involves
only propositions that deal with time $m$.

Neither of these shortcomings is serious.
First, variants of AGM
revision that involve partial orders were discussed in the literature
\cite{KM92,Rott92}. It is fairly straightforward to show that these
can captured in our systems using BCSs that satisfy REV1, REV3, and
REV4. Second, it is easy to add to $\Sys^*$
runs so as to get a system that satisfies REV4.  Moreover, we can do
this is a way that does not change the agent's
beliefs for sequences of observations that can be observed in $\Sys$.
Thus, the ``statified'' version of a system
in $\SU$ displays behavior much in the spirit of belief revision.

This result may seem somewhat surprising in light of the significant
differences between the AGM postulates and KM postulates.
In part, it shows how much is bound up in our choice of language.
(Recall that similar issues arose in Example~\ref{xam:diag-rev}.)
This highlights the sensitivity of the
postulate approach to the modeling assumptions we make. Unfortunately,
these modeling assumptions are rarely discussed in
the belief change literature.
(See \cite{FrH8Full} for a more
detailed
discussion of this point.)

Table~\ref{tab:summary} emphasizes that, despite the well-known
differences between revision and update, they can be viewed as sharing
one very important feature: they both use conditioning to do belief
change. Thus, we have a common mechanism both
for understanding and extending them.
To a certain extent, our results show that
revision is more general than update,
in the sense that
we can view the 
statified version of any system in $\SU$ as performing revision
(possibly with unranked prior) over runs.

\section{Extensions}\label{sec:extensions}

In the preceding sections, we introduced several assumptions that
were needed to capture revision and update. Of
course, there are other ways of capturing these notions that
require somewhat different assumptions. Nevertheless, these
assumptions give insight into the underlying choices
made, either explicitly or implicitly, in the definition of revision
and update.
In addition, thinking in terms of such restrictions makes it
straightforward to extend the intuitions of revision and update beyond
the context where they were originally applied.  In this section, we
consider a number of such extensions, to illustrate our point.

\subsection{Knowledge}
In many domains of interest, the agent
knows that some sequences of observations are impossible.
We already saw in the circuit-diagnosis problem that observing failures
was impossible.  In the context of update, we know that we cannot
observe a person die and then be alive, despite the fact that both
being dead and being alive are consistent states.

We can easily maintain what we regard as the defining properties of
revision and update, as discussed in the previous section: no change in
the environment state and a
ranked
prior in the case of
revision, and a lexicographic prior in the case of update, with belief
change proceeding by conditioning in both cases.
We simply drop REV3 and
replace REV4 by REV4$'$ (\respc drop UPD3 and replace UPD4 and UPD4$'$).
We remark that this change affects the postulates. For example,
consider
update.
Suppose that the agent considers the possibility that Mr.~Bond
is dead. If she then observes Mr.~Bond alive and well then, according
to update, she must account for the new observation by some change from
the worlds she previously considered possible. However, there is  no
transition from worlds in which Mr.~Bond is
dead that can account for the new observation. Thus, once the agent knows
that certain transitions are impossible, some observations (\eg
observing that Mr.~Bond is alive) require her to remove from
consideration some of the worlds that she previously considered
possible. As a consequence, postulate U8 does not hold, since the
agent's new beliefs are not determined by a pointwise update at each of
the worlds she previously considered possible.
(Boutilier~\citeyear{Boutilier95Full}
uses a related semantic framework to draw similar conclusions in his
analysis of update.) 

\subsection{Language of Beliefs}
In our analysis of revision and update, we focused on the agent's
beliefs about the current state of the environment. Often we are also
interested in how the agent changes her beliefs about other types of
statements, such as beliefs about future states of the environment,
beliefs about other agents' beliefs, and introspective beliefs about
her own beliefs.
Again, it is
straightforward in our framework
to deal with an enriched language
that lets us express such statements.   For
example, in \cite{FrH4}
we examine {\em  Ramsey conditionals}.
These are formulas of the form $\phi \RCond \psi$, which can be read as
saying ``after learning $\phi$, the agent believes $\psi$''. This
formula can be expressed as $\learn(\phi) \rimp B\psi$ in the language
$\LKPT$.  As is well known,
if belief sets include Ramsey conditionals (and not just propositional
formulas), then the AGM postulates become inconsistent (at least,
provided we have at least three mutually exclusive consistent formulas
in the language)
\cite{Gardenfors86}. Similar inconsistency results arise when one
tries to add other forms of introspective beliefs \cite{Fuhrmann}.
In our setting, it is easy to see why the problem arises.  Even if we
allow belief sets to include nonpropositional formulas, it still seems
quite clear that we want to distinguish the propositional formulas from
formulas that talk explicitly about an agent's beliefs.  For example, it
is not clear that we should allow an observation of a formula such as
$\phi \RCond \psi$. What would it mean to observe such a formula? It
clearly seems quite different from observing a propositional formula.
Nor does it make  sense to extend an assumption such as REV1
to arbitrary formulas.  While it may be reasonable to restrict to static
propositions if we are viewing these as making statements about a
relatively stable environment, it seems far less reasonable to assume
that formulas that talk about an agent's beliefs will be static,
especially when we are trying to model belief change!

Of course, if we allow only propositional formulas to be learned (or
observed), and restrict REV1 to propositional formulas, then it is easy
to see that all of our results still hold, even if the full language is
quite rich; we avoid the triviality result completely.

\subsection{Observations}
One of the strongest assumptions made by revision and update involves
the treatment of observations.  This assumption seems unreasonable in
most domains.  REV4 and UPD4 essentially assume that the observation
that the agent makes is chosen randomly among all formulas consistent
with the current state of the world.
Suppose that $\phi$ says that the agent is outdoors,  $\psi$ says that
the agent is in the basement, and $o_1$ says that the basement light is
on.  We may well have $\Pl_a(\RE{\phi \land o_1}) > \Pl_a(\RE{\psi
\land o_1})$.  For example, the agent may hardly ever go to the basement
and frequently go outdoors, but her children may often leave the
basement light on.  Nevertheless, we may also have $\Pl_a(\RE{\phi;
o_1}) < \Pl_a(\RE{\psi;o_1})$, contradicting REV4.  Indeed, it may well
be impossible for the agent to observe that the basement light is on
when she is outdoors, so that $\Pl_a(\RE{\phi; o_1}) = \bot$, but this
is not permitted according to REV4 or UPD4.

In many domains it is
useful to reason about hidden quantities that simply cannot be
observed. For example, the event that  component $c_i$ is faulty in
\Xref{xam:diag-rev} is a basic event in our description of the
problem, yet it cannot be observed. Similarly, the event where a
patient has a disease X or the opponent is planning to
capture the queen are useful in reasoning about medical diagnosis and
game strategy, yet are not directly observable in practice.
Thus, the requirement that all formulas in the language can be
observed seems quite unnatural.
We note that explicitly modeling sensory input is a standard practice in
control theory and stochastic processes (\eg in hidden Markov chains).
In these fields, one models the probability of an observation in
various situations. Making an observation increases the probability of
situations where that observation is likely to be observation and
decreases the probability of situations where it is unlikely.
Again, it is straightforward to consider a more detailed model of the
observation process in our framework; see \cite[Chapter 6]{FrThesis}
and \cite{BFH1}.

\subsection{Actions} Our definition of belief change systems
essentially assumes that the agent is {\em passive.\/} The situation
is more complex when the agent can influence the environment.
The agent's choice of action interacts with her beliefs. It is clear
that after performing an action, the agent should change her beliefs.%
\footnote{Indeed, an alternative interpretation of the update
postulates is that they describe how the agent should update her
beliefs after doing the action ``achieve $\phi$''
\cite{Goldszmidt+Pearl:1996,delVal1,delVal2}. However, as these works show,
the update postulates are problematic under this interpretation.}
Moreover, the information content of observations depends on the
action the agent has just performed. For example, the agent might
consider hearing a loud noise to be surprising. However,
it would be expected after the agent pulls the trigger of her
gun.

\subsection{Summary}

This list of possible extensions is clearly not exhaustive; there are
many others that we may want to consider. Nevertheless, these are
extensions that seem to be of interest. The main points we want to make
here are (1) it is easy to accommodate these extensions in our
framework while still maintaining the main characteristics of revision
and update,
and (2) it is difficult to deal with such extensions if we
focus on postulates.

\section{Conclusion}
\label{sec:discussion}

We have shown how the framework introduced in \cite{FrH1Full} can be
used to capture belief revision and update.  Modeling revision and
update in the framework also gives us a great deal of further insight
into their properties, and emphasizes the role of conditioning as a way
of capturing minimal change.

Of course, revision and update
are but two points in a wide spectrum of possible types of belief
change.  Our ultimate
goal is to use this framework to understand the whole
spectrum better and to help us design belief change
operations that
overcome some of the difficulties
we have observed with revision and update.  In particular, we want
belief change operations that can handle dynamic propositions,
while still being able to revise information about the past.

Our framework suggests how to construct such belief change operations.
In this framework, belief change operations can be determined by
choosing a plausibility measure that captures the agent's preferences
among sequences of worlds.  This is the agent's prior plausibility,
and captures her initial beliefs about the relative likelihood of
runs. As the agent receives information, she changes her beliefs using
conditioning. In this paper we show that revision and update
correspond to two specific families of priors. Clearly, however, there
are prior plausibilities that, when conditioned on a surprising
observation, allow the agent to revise some earlier beliefs and to
assume that some change has occurred.  One obvious problem is that,
even if there are only two possible states, there are uncountably many
possible runs.  How can an agent describe a prior plausibility over
such a complex space?

One approach to doing this is based on intuition from the
probabilistic settings. In these settings, the standard solution to
this problem is to assume that state transitions are independent of
when they occur, that is, that the probability of the system going
from state $s$ to state $s'$ is independent of the sequence of
transitions that brought the system to state $s$.  This {\em Markov
assumption\/} significantly reduces the complexity of the problem.
All that is necessary is to describe the probability of state
transitions. In \cite{FrH6,FrThesis} we define a notion of plausibilistic
independence, and show how to describe priors that satisfy the Markov
assumption and the consequences for belief change.
See also \cite{Boutilier95Full,BFH1} for  recent proposals along these lines.

Whether or not this particular approach turns out to be a useful one,
it is clear that these are the types of questions we should be asking.
As these works show, our framework provides a useful basis for answering
them.

Finally, we note that our approach is quite different from the
traditional approach to belief change \cite{agm:85,Gardenfors1,KM91}.
Traditionally, belief change was
viewed as an abstract process.
Our framework, on the other hand, models the agent and the environment
she is situated in, and how both change in time. This allows us to
model concrete agents in concrete settings (for example,
diagnostic systems are analyzed in
\cite{FrH1Full} and throughout this paper),
and to reason about the
beliefs and knowledge of such agents.
    We can then
    investigate what plausibility ordering
    induces beliefs that match our intuitions.
By gaining a better understanding of such concrete situations, we can
better investigate more abstract notions of belief change.
More generally, we believe that, when studying belief change, it is
important to specify the underlying {\em ontology}: that is, exactly
what scenario underlies the belief-change process.  We have
specified one such scenario here.  While others are certainly possible,
we view it as a defect in the literature on belief change that the
underlying scenario is so rarely discussed.  The framework we have
introduced here provides a way of making formal what the scenario is.
(See \cite{FrH8Full} for further discussion of this issue.)

\section*{Acknowledgments}
The authors are grateful to Craig Boutilier, Ronen Brafman, Adnan
Darwiche, Moises Goldszmidt, Adam Grove, Alberto Mendelzon, Alvaro
del~Val, and particularly Daphne Koller and Moshe Vardi, for comments
on drafts of this paper and useful discussions relating to this work.
Some of this work was done while both authors were at the IBM Almaden
Research Center.
The first author was also at Stanford while much of
the work was done.  IBM and Stanford's support are gratefully
acknowledged.   The work
was also supported in part by the Air Force Office of Scientific
Research (AFSC), under Contract F49620-91-C-0080 and grant
F94620-96-1-0323 and by NSF under grants
IRI-95-03109 and IRI-96-25901.  The first author was also supported in
part by an IBM Graduate Fellowship and by Rockwell Science Center.
A preliminary version of this paper
appears in J.~Doyle, E.~Sandewall, and P.~Torasso (Eds.), {\em
Principles of Knowledge Representation and Reasoning:
Proc.~Fourth International Conference\/},  1994, pp. 190--201,
under the title ``A knowledge-based
framework for belief change, Part II: revision and update.''
\appendix
\section{Proofs}

\subsection{Proofs for Section~\protect\ref{sec:revision}}
\label{prf:revision}

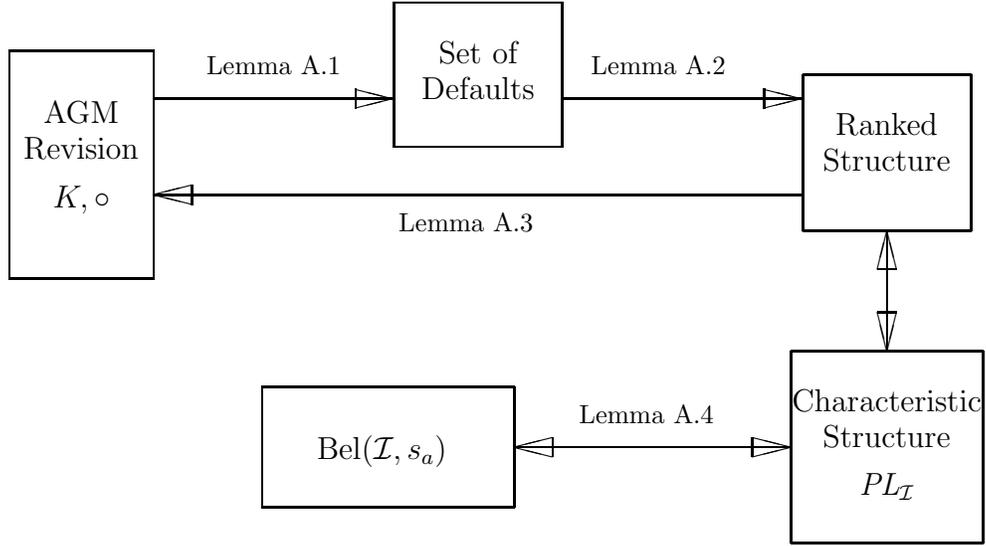
\begin{figure}
\begin{center}
\setlength{\unitlength}{0.00083333in}
\begingroup\makeatletter\ifx\SetFigFont\undefined
\def\x#1#2#3#4#5#6#7\relax{\def\x{#1#2#3#4#5#6}}%
\expandafter\x\fmtname xxxxxx\relax \def\y{splain}%
\ifx\x\y   % LaTeX or SliTeX?
\gdef\SetFigFont#1#2#3{%
  \ifnum #1<17\tiny\else \ifnum #1<20\small\else
  \ifnum #1<24\normalsize\else \ifnum #1<29\large\else
  \ifnum #1<34\Large\else \ifnum #1<41\LARGE\else
     \huge\fi\fi\fi\fi\fi\fi
  \csname #3\endcsname}%
\else
\gdef\SetFigFont#1#2#3{\begingroup
  \count@#1\relax \ifnum 25<\count@\count@25\fi
  \def\x{\endgroup\@setsize\SetFigFont{#2pt}}%
  \expandafter\x
    \csname \romannumeral\the\count@ pt\expandafter\endcsname
    \csname @\romannumeral\the\count@ pt\endcsname
  \csname #3\endcsname}%
\fi
\fi\endgroup
{\renewcommand{\dashlinestretch}{30}
\begin{picture}(6099,3414)(0,-10)
\put(462,2637){\makebox(0,0)[b]{\smash{{{\SetFigFont{12}{14.4}{rm}AGM}}}}}
\put(462,2412){\makebox(0,0)[b]{\smash{{{\SetFigFont{12}{14.4}{rm}Revision}}}}}
\put(462,2112){\makebox(0,0)[b]{\smash{{{\SetFigFont{12}{14.4}{rm}$K,\Rev$}}}}}
\put(2937,3012){\makebox(0,0)[b]{\smash{{{\SetFigFont{12}{14.4}{rm}Set of}}}}}
\put(2937,2787){\makebox(0,0)[b]{\smash{{{\SetFigFont{12}{14.4}{rm}Defaults}}}}}
\drawline(2412,3387)(3462,3387)(3462,2487)
	(2412,2487)(2412,3387)
\drawline(1587,987)(3162,987)(3162,237)
	(1587,237)(1587,987)
\put(2337,537){\makebox(0,0)[b]{\smash{{{\SetFigFont{12}{14.4}{rm}$\BEL(\Sys,s_a)$}}}}}
\drawline(5487,1962)(5487,1962)
\drawline(5487,1962)(5487,1212)
\whiten\drawline(5547.000,1722.000)(5487.000,1962.000)(5427.000,1722.000)(5547.000,1722.000)
\whiten\drawline(5427.000,1452.000)(5487.000,1212.000)(5547.000,1452.000)(5427.000,1452.000)
\drawline(4887,612)(3162,612)
\whiten\drawline(4647.000,552.000)(4887.000,612.000)(4647.000,672.000)(4647.000,552.000)
\whiten\drawline(3402.000,672.000)(3162.000,612.000)(3402.000,552.000)(3402.000,672.000)
\drawline(4962,2187)(912,2187)
\whiten\drawline(1152.000,2247.000)(912.000,2187.000)(1152.000,2127.000)(1152.000,2247.000)
\drawline(12,3087)(912,3087)(912,1662)
	(12,1662)(12,3087)
\drawline(3462,2787)(4962,2787)
\whiten\drawline(4722.000,2727.000)(4962.000,2787.000)(4722.000,2847.000)(4722.000,2727.000)
\put(5487,312){\makebox(0,0)[b]{\smash{{{\SetFigFont{12}{14.4}{rm}$\PL_\Sys$}}}}}
\drawline(912,2787)(2412,2787)
\whiten\drawline(2172.000,2727.000)(2412.000,2787.000)(2172.000,2847.000)(2172.000,2727.000)
\drawline(4962,2937)(6012,2937)(6012,1962)
	(4962,1962)(4962,2937)
\drawline(4887,1212)(6087,1212)(6087,12)
	(4887,12)(4887,1212)
\put(3987,762){\makebox(0,0)[b]{\smash{{{\SetFigFont{10}{12.0}{rm}Lemma \ref{lem:Bel-in-SR}}}}}}
\put(1662,2937){\makebox(0,0)[b]{\smash{{{\SetFigFont{10}{12.0}{rm}Lemma \ref{lem:Delta_K}}}}}}
\put(2862,1962){\makebox(0,0)[b]{\smash{{{\SetFigFont{10}{12.0}{rm}Lemma \ref{lem:Rank-to-Rev}}}}}}
\put(4062,2937){\makebox(0,0)[b]{\smash{{{\SetFigFont{10}{12.0}{rm}Lemma \ref{lem:Exists-Ranking}}}}}}
\put(5487,2562){\makebox(0,0)[b]{\smash{{{\SetFigFont{12}{14.4}{rm}Ranked}}}}}
\put(5487,2337){\makebox(0,0)[b]{\smash{{{\SetFigFont{12}{14.4}{rm}Structure}}}}}
\put(5487,837){\makebox(0,0)[b]{\smash{{{\SetFigFont{12}{14.4}{rm}Characteristic}}}}}
\put(5487,612){\makebox(0,0)[b]{\smash{{{\SetFigFont{12}{14.4}{rm}Structure}}}}}
\end{picture}
}

\end{center}
\caption{Schematic description of the entities and lemmas involved
in the proof of Theorems~\protect\ref{rep1} and~\protect\ref{rep2}.}
\label{fig:revision-proof}
\end{figure}

We start with the proof of Theorems~\ref{rep1} and~\ref{rep2}. To do
this, we need some preliminary definitions and
lemmas. Figure~\ref{fig:revision-proof} shows the general outline of
the intermediate representations we use in these proofs. Roughly
speaking, we show how to map from a revision operator $\Rev$ and a
consistent belief set $K$ to a ranking, and similarly how to map from
a ranking to an AGM revision operator. These rankings correspond,
in a direct way, to priors in systems in $\SR$, and thus have close
connection to the beliefs of the agent in various states.

These mapping between AGM revision operators and rankings are related
to the representation theorems of
Boutilier~\citeyear{Boutilier94AIJ2}, Grove~\citeyear{Grove}, and
Katsuno and Mendelzon~\citeyear{KM91}. However, the exact details of our
representations are different than those of Boutilier, Grove, and
Katsuno and Mendelzon. Thus, for completeness we provide the full
proofs here.

We start with the mapping from revision operator applied to a specific
belief set to a ranking. As an intermediate step we construct a set of
defaults as follows. We then will use the results from
\cite{FrH5Full} to construct a ranked plausibility structure
that satisfies these defaults.

\lem\label{lem:Delta_K}
Let $\Rev$ be an AGM revision operator, let $K \subseteq \Le$ be a
consistent belief set, and  let
$$\Delta_{(\Rev,K)} = \{ \phi \Cond \psi : \phi,\psi \in \Le,\, \psi \in
K\Rev\phi \}.$$ Then the following is true:
\begin{itemize}\denselist
\item[(a)] $\Delta_{(\Rev,K)}$ is closed under the rules of $\SysP$,
\item[(b)] $\phi \Cond \False \not\in \Delta_{(\Rev,K)}$ for all consistent $\phi \in
\Le$, and
\item[(c)] $\Delta_{(\Rev,K)}$ satisfies rational monotonicity; that is, if
$\phi\Cond\psi \in \Delta_{(\Rev,K)}$ and $\phi\Cond\neg\xi \not\in
\Delta_{(\Rev,K)}$, then $\phi\land\xi \Cond \psi \in \Delta_{(\Rev,K)}$.
\end{itemize}
\elem

\prf
We start with part (a):
\begin{itemize}\denselist
\item[LLE] Assume that $\vdash_{\Le} \phi \equiv \phi'$ and that
$\phi \Cond \psi \in \Delta_{(\Rev,K)}$. Thus, $\psi \in K \Rev \phi$. {From}
R5, it follows that $\psi \in K \Rev \phi'$, and thus $\phi' \Cond \psi
\in \Delta_{(\Rev,K)}$.

\item[RW] Assume that $\vdash_{\Le} \psi \rimp \psi'$ and that
$\phi \Cond \psi \in \Delta_{(\Rev,K)}$. Thus, $\psi \in K \Rev \phi$. Since
$K \Rev \phi$ is a belief set, it is closed under logical
consequence. In particular, $\psi' \in K \Rev\phi$, and hence $\phi
\Cond \psi' \in \Delta_{(\Rev,K)}$.

\item[REF] By R2, $\phi \in K \Rev \phi$, and thus, $\phi
\Cond \phi \in \Delta_{(\Rev,K)}$.

\item[AND] Assume that $\phi \Cond \psi_1, \phi \Cond \psi_2 \in
\Delta_{(\Rev,K)}$. Thus, $\psi_1, \psi_2 \in K \Rev\phi$. Since $K \Rev
\phi$ is a belief set, $\psi_1 \land \psi_2 \in K
\Rev\phi$. Thus, $\phi \Cond \psi_1\land\psi_2 \in \Delta_{(\Rev,K)}$.

\item[OR] Assume that $\phi_1 \Cond \psi, \phi_2 \Cond \psi \in
\Delta_{(\Rev,K)}$. There are two cases. If $K \Rev (\phi_1 \lor
\phi_2)$ is inconsistent, then $\psi \in K \Rev (\phi_1 \lor \phi_2)$
and thus $\phi_1\lor\phi_2 \Cond \psi \in \Delta_{(\Rev,K)}$.
If $K  \Rev (\phi_1\lor \phi_2)$ is consistent, then, by R2,
$\phi_1 \lor \phi_2 \in K \Rev (\phi_1 \lor \phi_2)$. Thus, we cannot
have both $\neg\phi_1$ and $\neg\phi_2$ in $K
\Rev (\phi_1\lor\phi_2)$. Without loss of generality, assume that
$\neg\phi_1 \not\in K \Rev (\phi_1\lor\phi_2)$. Using R7 and R8, we
get that  $K \Rev((\phi_1\lor\phi_2)\land\phi_1) =
Cl(K\Rev(\phi_1\lor\phi_2) \union \{ \phi_1 \} )$. Using R6, we get that $K
\Rev((\phi_1\lor\phi_2)\land\phi_1) = K \Rev \phi_1$. Thus, we
conclude that $K \Rev \phi_1 = Cl(K\Rev(\phi_1\lor\phi_2) \union \{
\phi_1 \} )$.
Since $\phi_1 \Cond \psi \in \Delta_{(\Rev,K)}$, we have that $\psi
\in K \Rev \phi_1$. Thus, we get that $\phi_1 \rimp \psi \in
K\Rev(\phi_1\lor\phi_2)$. If $\neg\phi_2 \not\in
K\Rev(\phi_1\lor\phi_2)$, by similar arguments we get that
$\phi_2 \rimp \psi \in  K\Rev(\phi_1\lor\phi_2)$.
This implies
that $(\phi_1 \lor \phi_2) \rimp \psi \in K\Rev(\phi_1\lor\phi_2)$,
and thus $\psi \in K\Rev(\phi_1\lor\phi_2)$. On the other hand, if
$\neg\phi_2 \in  K\Rev(\phi_1\lor\phi_2)$, then, since
$\phi_1\lor\phi_2 \in K\Rev(\phi_1\lor\phi_2)$, we get that $\phi_1 \in
K\Rev(\phi_1\lor\phi_2)$,
and thus $\psi \in K\Rev(\phi_1\lor\phi_2)$.

\item[CM] Assume that $\phi \Cond \psi_1, \phi \Cond \psi_2 \in
\Delta_{(\Rev,K)}$. If $K \Rev \phi$ is inconsistent, then using R5
we get that $\phi$ is inconsistent. Thus, $\phi\land\psi_1$ is
inconsistent, so $\psi_2 \in K\Rev(\phi\land\psi_1)$. Now assume that
$K \Rev \phi$ is consistent. Since $\phi\Cond\psi_1$, we have that
$\psi_1 \in K\Rev\phi$. Since $K \Rev \phi$ is consistent, we get
that $\neg \psi_1 \not\in K \Rev\phi$. Applying R8, we get that $K
\Rev \phi \subseteq K \Rev(\phi\land\psi_1)$. Since
$\phi\Cond\psi_2\in \Delta_{(\Rev,K)}$, we have that $\psi_2 \in K
\Rev \phi$. Thus, $\psi_2 \in K \Rev
(\phi\land\psi_1)$. This implies that $(\phi\land\psi_1) \Cond \psi_2
\in \Delta_{(\Rev,K)}$.
\end{itemize}

We now prove part (b). Let $\phi \in \Le$ be a consistent
formula. Then, using R5, we get that $K \Rev\phi$ is consistent. Thus,
$\phi \Cond \False \not\in \Delta_{(\Rev,K)}$.

Finally we prove part (c). Assume that $\phi\Cond \psi \in \Delta_{(\Rev,K)}$,
and $\phi\land\xi \Cond \psi \not\in \Delta_{(\Rev,K)}$. Since $\phi\Cond\psi
\in \Delta_{(\Rev,K)}$, we have that $\psi \in K \Rev \phi$. Now if $\neg\xi
\not \in K \Rev \phi$, then, using R8, we have that $Cl(K \Rev\phi
\union \{ \xi \}) \subseteq K \Rev (\phi\land\xi)$. This implies that
$\psi \in K \Rev (\phi\land\xi)$. However, since we assumed that
$\phi\land\xi \Cond \psi \not \in\Delta_{(\Rev,K)}$, we have that
$\psi \not \in K \Rev (\phi\land\xi)$; thus, we get a
contradiction. We conclude that $\neg\xi \in K \Rev \phi$. Thus, $\phi
\Cond \neg\xi \in
\Delta_{(\Rev,K)}$.
\eprf

We now use this result to show that there exists a plausibility
structure that corresponds to $\Rev$ applied to $K$.
\lem\label{lem:Exists-Ranking}
Let $\Rev$ be an AGM revision operator, and let $K \subseteq \Le$ be a
consistent belief set. Then there is a plausibility structure $\PL =
(W,\Pl,\pi)$ such that $\Pl$ is ranked, $\PL \sat \phi
\Cond \psi$ if and only if $\psi \in K \Rev \phi$, and
$\Pl(\intension{\phi}) > \bot$ for all $\vdash_\Le$-consistent
formulas $\phi \in \Le$.
\elem

\prf
We use the basic techniques described in the proof of
\cite[Theorem~8.2]{FrH5Full}.
Let $\Delta_{(\Rev,K)}$ be the set of defaults defined by
Lemma~\ref{lem:Delta_K}. We now construct a plausibility
space $\PL' = (W,\Pl',\pi)$ such that $\PL' \sat \phi\Cond\psi$ if and
only if $\phi \Cond \psi \in \Delta_{(\Rev,K)}$.
We define $\PL'$ as follows:
\begin{itemize}\denselist
\item $W = \{ w_V : V\subseteq \Le$ is a maximal
$\vdash_\Le$-consistent set$\}$,
\item $\pi(w_V)(p) = $ {\bf true} if $p \in
V$, and
\item $\Pl'(\intension{\phi}) \ge \Pl'(\intension{\psi})$ if and only if $(\phi \lor \psi) \Cond \phi \in \Delta_{(\Rev,K)}$.
\end{itemize}
Using \cite[Lemma~4.1]{FrH5Full}, we get that $\PL' \sat \phi\Cond\psi$
if and only if $\phi\Cond\psi \in \Delta_{(\Rev,K)}$. {From}
Lemma~\ref{lem:Delta_K}~(c) and
and results of \cite{FrH5Full}, it follows that there is a
ranked plausibility measure $\Pl$ that is {\em default-isomorphic\/}
to $\Pl'$, that is $(W,\Pl,\pi)$ satisfies
precisely the same defaults as  $(W,\Pl',\pi)$.
Let $\PL = (W,\Pl,\pi)$.

Since $\PL$ is default-isomorphic to $\PL'$, we have that $\PL \sat
\phi \Cond \psi$ if and only if $\phi \Cond \psi \in
\Delta_{(\Rev,K)}$. Moreover, using Lemma~\ref{lem:Delta_K}, we have that $\phi
\Cond \psi \in \Delta_{(\Rev,K)}$ if and only if $\psi \in K \Rev
\phi$. Thus, $\PL \sat \phi\Cond\psi$ if and only if $\psi \in K \Rev
\phi$. Finally, let $\phi$ be a $\vdash_\Le$-consistent formula. {From}
Lemma~\ref{lem:Delta_K}~(b), we get that $\phi \Cond \False \not\in
\Delta_{(\Rev,K)}$. Since $\Delta_{(\Rev,K)}$ is closed under the
rules of \SysP, we
conclude that $(\phi \lor \False) \Cond \False \not\in
\Delta_{(\Rev,K)}$. Thus, $\Pl'(\intension{\phi}) \not\le
\bot = \Pl'(\intension{\False})$, and thus $\Pl'(\intension{\phi}) >
\bot$. Since $\Pl$ is default-isomorphic to $\Pl'$, we conclude that
$\Pl(\intension{\phi}) > \bot$.
\eprf

We now prove the converse to Lemma~\ref{lem:Exists-Ranking}.
\lem\label{lem:Rank-to-Rev}
Let $\PL = (W,\Pl,\pi)$ be a ranked plausibility structure such that
$\pi(w)$ is $\vdash_\Le$-consistent for all worlds $w$, and $\PL
\not\sat \phi\Cond\False$ for all $\vdash_\Le$-consistent formulas
$\phi\in\Le$; let $K = \{ \phi\in\Le : \PL \sat \True \Cond
\phi\}$. Then there is an AGM revision operator $\Rev$ such that
$\psi \in K \Rev\phi$ if and only if $\PL \sat \phi\Cond\psi$.
\elem

\prf
Let $\Rev$ be some belief change operation such that $K \Rev \phi = \{
\psi : \PL \sat \phi\Cond\psi \}$.  Since
this requirement constrains only the result of applying $\Rev$ to
$K$, we can assume without loss of generality that $\Rev$
satisfies the AGM postulates when applied to belief sets other than
$K$.  Thus, we  need prove only that $\Rev$ satisfies the
AGM postulates for revision applied to $K$. (Note that the proofs for R3
and R4 follow from the proofs for R7 and R8, respectively.)

\begin{description}\denselist
\item[R1] Since $\PL$ is qualitative, we have that $\{ \psi : \PL
\sat\phi\Cond\psi\}$ is a belief set, that is, closed under logical
consequences.

\item[R2] Axiom C1 implies that $\PL \sat \phi \Cond
\phi$. Thus, $\phi \in K \Rev \phi$.

\item[R5] By our assumptions, if $\phi$ is $\vdash_{\Le}$-consistent,
then $\Pl(\intension{\phi}) > \bot$, and thus $\PL \not\sat \phi
\Cond \false$. On the
other hand, if $\phi$ is not $\vdash_{\Le}$-consistent, then
$\intension{\phi} = \emptyset$, and thus $\Pl(\intension{\phi}) =
\bot$. We conclude that $\Pl( \intension{\phi} ) = \bot$ if and only
if $\vdash_\Le \neg\phi$.  This implies that $\PL \sat \phi \Cond
\False$ if and only if $\vdash_\Le\neg\phi$. Thus, $K \Rev \phi =
Cl(\False)$ if and only if $\vdash_\Le\neg\phi$.

\item[R6] Assume that $\vdash_{\Le} \phi \dimp \phi'$. Then, by
our assumption, $\pi(w)(\phi) = \pi(w)(\phi')$. Thus,
$\intension{\phi\land\psi} =
\intension{\phi'\land\psi}$ for all formulas
$\psi \in \Le$. We conclude that $\PL \sat \phi \Cond \psi$ if
and only if $\PL \sat \phi' \Cond\psi$. This implies that
$K\Rev \phi = K \Rev \phi'$.

\item[R7] There are two cases: either $\Pl(\intension{\phi\land\psi}) =
\bot$ or $\Pl(\intension{\phi\land\psi}) > \bot$.
If   $\Pl(\intension{\phi\land\psi}) =
\bot$,  then $\phi\land\psi$ is inconsistent.
According to R2, we have that $\phi \in K \Rev \phi$. Thus, $\phi
\land \psi \in Cl(K \Rev \phi \union \{ \psi \} )$. This implies that
$Cl(K \Rev \phi \union \{ \psi \} )$ contains $\False$, and thus $K
\Rev(\phi\land\psi) \subseteq Cl(K \Rev \phi \union \{ \psi \} )$. If
$\Pl(\intension{\phi\land\psi}) > \bot$, let $\xi \in K \Rev (\phi
\land \psi)$. We now show that $\xi \in Cl(K \Rev \phi \union \{ \psi
\})$. This will show that $K \Rev(\phi\land\psi) \subseteq Cl(K \Rev
\phi \union \{ \psi \} )$. Since $\xi \in K \Rev (\phi \land \psi)$,
we get that $\PL \sat (\phi\land\psi) \Cond \xi$. Since
$\Pl(\intension{\phi\land\psi}) > \bot$, we get that
$\Pl(\intension{\phi\land\psi\land\xi}) >
\Pl(\intension{\phi\land\psi\land\neg\xi})$. Then we have that
$\Pl(\intension{\phi\land(\psi\rimp \xi)}) >
\Pl(\intension{\phi\land\neg(\psi\rimp\xi))})$, since
$(\phi\land\psi\land\xi) \rimp (\phi\land(\psi\rimp\xi))$ and
$(\phi\land\neg(\psi\rimp\xi)) \rimp
(\phi\land\psi\land\neg\xi)$. This also implies that
$\Pl(\intension{\phi}) > \bot$. Thus, $\PL \sat \phi \Cond (\psi
\rimp \xi)$.  So, $(\psi \rimp \xi) \in K \Rev \phi$, and thus $\xi
\in Cl(K \Rev \phi \union \{ \psi \})$.

\item[R8] Assume that $\neg\psi \not\in K \Rev \phi$. Let $\xi
\in Cl(K \Rev \phi \union \{\psi\})$. We now show that $\xi \in
  K \Rev (\phi\land\psi)$. This will show that $Cl(K \Rev \phi \union
  \{\psi\}) \subseteq K \Rev (\phi\land\psi)$.
Let $A =
  \intension{\phi\land\neg\psi}$, $B =
  \intension{\phi\land\psi\land\xi}$, and $C =
  \intension{\phi\land\psi\land\neg\xi}$. It is easy to verify that these
  sets are pairwise disjoint.  Since $\phi\land(\psi\rimp\xi) \equiv
  (\phi\land\neg\psi) \lor( \phi\land\psi\land\xi)$ and
  $(\phi\land\neg(\psi\rimp\xi)) \equiv (\phi\land\psi\land\neg\xi)$, we
  conclude that $\intension{\phi\land(\psi\rimp\xi)} = A \union B$,
  and $\intension{\phi\land\neg(\psi\rimp\xi)} = C$.  Since $\xi \in
  Cl( K\Rev\phi \union \{ \psi \})$, we have that $(\psi \rimp \xi) \in
  K\Rev\phi$. This means that $\PL \sat \phi \Cond (\psi \rimp
  \xi)$. Thus, either $\Pl(\intension{\phi}) = \bot$ or $\Pl(A
  \union B) > \Pl(C)$. If $\Pl(\intension{\phi}) = \bot$, then
  according to A1, we get that $\Pl(\intension{\phi\land\psi}) =
  \bot$. Thus, $\PL \sat (\phi\land\psi)\Cond\xi$ vacuously, and
  $\xi \in K \Rev (\phi\land\psi)$ as desired.

  Now assume that $\Pl(A \union B) > \Pl(C)$.  Since $\Pl$ is ranked, it
  satisfies A4$'$ and A5$'$. According to A5$'$, we get that either $\Pl(A)
  > \Pl(C)$ or $\Pl(B) > \Pl(C)$. Assume that $\Pl(A) > \Pl(C)$ and
  $\Pl(B) \not> \Pl(C)$. Then, using A4$'$, we get that $\Pl(A) >
  \Pl(B)$. Applying A2, we get that $\Pl(A) > \Pl(B \union
  C)$. However since $A = \intension{\phi\land\neg\psi}$ and $B \union
  C = \intension{\phi\land\psi}$, this implies that $\neg\psi \in
  K\Rev\phi$, which contradicts our assumption. Thus, we conclude that
  $\Pl(B) > \Pl(C)$. Since $B = \intension{\phi\land\psi\land\xi}$
  and $C = \intension{\phi\land\psi\land\neg\xi}$, we get that $\PL
  \sat (\phi\land\psi) \Cond \xi$, and thus $\xi \in K \Rev
  (\phi\land\psi)$.

\item[R3 {\rm and} R4] Our definition of $\Rev$ implies that $K \Rev \True
= K$. According to R6, we have that $K \Rev (\True\land  \phi)$ = $K \Rev
\phi$. Combining these two facts, we get that R3 and R4 are special
cases of R7 and R8, respectively.
\end{description}
\eprf

These results show how to map between ranked plausibility structures
and AGM revision operators. We now relate systems in $\SR$ and ranked
plausibility structures. Let $\Sys = (\R,\pi,\Plass) \in \SR$.
Recall that REV2 requires that the prior of $\Sys$ be a ranking. Thus,
we can construct a ranked plausibility structure where worlds are runs
in $\R$. We define the {\em characteristic
structure\/} of $\Sys$ to be
$\PL_\Sys = (\R,\Pl_a,\pi_{\sPl_\Sys})$, where $\Pl_a$ is the agent's
prior over runs and $\pi_{\sPl_\Sys}(r)(p) = \pi(r,0)(p)$ for all $p
\in \Phis$. Note that $\intension{\phi}_{\sPL_\Sys} =
\RE{\phi}$.

We now use $\PL_\Sys$ to describe the beliefs of the agent in each
local state.

\lem\label{lem:Bel-in-SR}
Let $\Sys \in \SR$ and let $s_a = \<o_1,\ldots,o_m\>$. Then $\phi \in
\BEL(\Sys,s_a)$ if and only if $\PL_\Sys \sat (\band_{i = 1}^m{o_i})
\Cond \phi$. (By convention, if $m = 0$, we take $(\band_{i =
1}^m{o_i})$ to be {\bf true}.)
\elem

\prf
Let $\Sys \in \SR$ and let $s_a = \<o_1,\ldots,o_m\>$.
There are two cases: either $s_a$ is a local state in
$\Sys$,  or it is not.

If $s_a$ is a local state in $\Sys$, suppose that $r_a(m) = s_a$. Note
that $\phi \in \BEL(\Sys,s_a)$ if and only if
$\Pl_{(r,m)}(\intension{\phi}_{(r,m)}) >
\Pl_{(r,m)}(\intension{\neg\phi}_{(r,m)})$.
Recall that,
according to the definition of conditioning, $\Pl_{(r,m)}(\cdot)$ is
isomorphic to $\Pl_a( \cdot | \RE{\cdot;
o_1,\ldots, o_m })$. Thus,
$\Pl_{(r,m)}(\intension{\phi}_{(r,m)}) >
\Pl_{(r,m)}(\intension{\neg\phi}_{(r,m)})$ if and only if $\Pl_a(
\RE{\phi} \mid \RE{ \cdot ; o_1,\ldots,o_m }) > \Pl_a( \RE{\neg\phi} \mid
\RE{ \cdot ; o_1,\ldots,o_m })$. Using C1, this is true if and only if
$\Pl_a(\RE{\phi ; o_1,\ldots,o_m }) > \Pl_a( \RE{\neg\phi;
o_1,\ldots,o_m })$. Using REV4, this is true if and only if $\Pl_a(
\RE{ \phi\land \band_{i = 1}^m{o_i}}) > \Pl_a(\RE{ \neg\phi
\land \band_{i = 1}^m{o_i}})$.  We get that $\phi \in
\BEL(\Sys,s_a)$ if and
only if $\Pl_a(\RE{\phi\land \band_{i = 1}^m{o_i}}) >
\Pl_a(\RE{\neg\phi\land \band_{i = 1}^m{o_i}})$. This implies that $\phi \in
\BEL(\Sys,s_a)$ if and only if $\PL_\Sys \sat (\band_{i = 1}^m{o_i})
\Cond \phi$.

If $s_a$ is not a  local state
in $\Sys$, then $\RE{\cdot ; o_1,\ldots, o_m} = \emptyset$, and by
definition $\Pl_a(\RE{\cdot ; o_1,\ldots, o_m}) = \bot$. Using C1 and
REV4, we get that $\PL_a(\RE{\band_{i = 1}^m{o_i}}) = \bot$, and
thus $\PL_\Sys \sat (\band_{i = 1}^m{o_i}) \Cond \phi$ for all
$\phi \in \Le$. Since $s_a$ is not a local state in $\Sys$,
by definition $\BEL(\Sys,s_a) = \Le$. Hence, we can conclude that
$\phi \in \BEL(\Sys,s_a)$ if and only if $\PL_\Sys \sat
(\band_{i = 1}^m{o_i}) \Cond \phi$.
\eprf

We now show that given a ranked plausibility structure $\PL$ we can
construct a system whose characteristic structure is
default-isomorphic to $\PL$.

\lem\label{lem:Exists-SR-Prior}
Let $\PL_K = (W_K,\Pl_K,\pi_K)$ be a plausibility space that satisfies
the conditions of Lemma~\ref{lem:Rank-to-Rev}. Then there is a system
$\Sys \in \SR$ such that $\PL_\Sys = \PL_K$.
\elem

\prf
Let $\PL_K = (W_K,\Pl_K,\pi_K)$ be a
plausibility space that satisfies the conditions of
Lemma~\ref{lem:Rank-to-Rev}.
For each world $w \in W_K$ and sequence of observations
$o_1,o_2, \ldots$, let $r^{w,o_1,o_2,\ldots}$ be the run
defined so that $r^{w,o_1,o_2,\ldots}_e(m) = w$ and
$r^{w,o_1,o_2,\ldots}_a(m) = \< o_1, \ldots, o_m \>$ for all $m$.
Let $\R = \{ r^{w,o_1,o_2,\ldots} : \pi_k(w)(o_i) = $ {\bf true} for
all $i \}$.
Define $\pi$
so that $\pi(r,m)(p) = \pi_K(r_e(m))(p)$ for $p \in
\Phis$, and so that $\pi(r,m)(\learn(\phi)) =$ {\bf true} if
$\obs{r,m} = \phi$ for $\phi \in \Le$. Finally, define the
prior plausibility $\Pl_a$ so that $\Pl_a( R ) = \Pl_K( \{ w :
\exists r \in R( w = r_e(0)) \}$. It is easy to check that this
definition implies that $\Pl_a(\RE{\phi}) =
\Pl_K(\intension{\phi}_{\sPL_K})$. Thus, $\PL_\Sys = \PL_K$.
Since
$\Pl_K$ is a ranking, $\Pl_a$ is also a ranking and thus
qualitative.

We now verify that the resulting interpreted system is indeed in
$\SR$. It is easy to check that $\Sys$ is a belief change system;
that is, it satisfies BCS1--BCS5. The construction is such that
$r_e(m) = r_e(0)$ for all runs $r$ and times $m$. Thus,
$\Sys$ satisfies REV1. Since the prior $\Pl_a$ is a ranking,
this system also satisfies REV2.
Lemma~\ref{lem:Exists-Ranking} implies that if $\phi$ is a consistent
formula, then $\Pl_K(\intension{\phi}_{\sPL_K}) > \bot$. This implies
that $\Pl_a(\RE{\phi}) > \bot$, and thus the system satisfies
REV3. Finally, it is easy to show that $\Pl_a( \RE{\phi; o_1,\ldots, o_m
} ) = \Pl_a( \RE{\phi\land o_1 \land \ldots \land o_m} ) = \Pl_K (
\intension{\phi\land o_1 \land \ldots \land o_m}_{\sPL_K})$. Thus, the
system satisfies REV4.
\eprf

We are finally ready to prove Theorem~\ref{rep1}.

\rethm{rep1}
Let $\Rev$ be an AGM revision operator and let $K \subseteq \Le$ be a
consistent belief set. Then there is a system $\Sys_{(\Rev,K)} \in
\SR$
such that
$\BEL(\Sys_{(\Rev,K)},\<\>) = K$ and
$$
\BEL(\Sys_{(\Rev,K)},\<\>) \Rev \phi = \BEL(\Sys_{(\Rev,K)}, \<\phi\>)
$$
for all
$\phi \in \Le$.
\erethm

\prf
Let $\Rev$ be an AGM revision operator and let $K \subseteq \Le$ be a
consistent belief set.  By Lemmas~\ref{lem:Exists-Ranking}
and~\ref{lem:Exists-SR-Prior}, there
is a system $\Sys_{(\Rev,K)} = (\R_{(\Rev,K)},\pi_{(\Rev,K)},
\P_{(\Rev,K)}) \in \SR$ such that $\PL_{\Sys_{(\Rev,K)}} \sat \phi
\Cond \psi$ if and only if $\psi \in K\Rev \phi$.
Our
construction is such that $\psi \in K \Rev \phi$ if and only if
$\PL_{\Sys_{(\Rev,K)}} \sat \phi \Cond \psi$. Using
Lemma~\ref{lem:Bel-in-SR}, we get that $\PL_{\Sys_{(\Rev,K)}} \sat
\phi \Cond \psi$ if and only
if $\psi \in \BEL(\Sys_{(\Rev,K)},\<\phi\>)$. Thus, $K \Rev \phi =
\BEL(\Sys_{(\Rev,K)},\<\phi\>)$.

Finally, we show $\BEL(\Sys_{(\Rev,K)},\<\>) = K$.
We start by showing that $K \Rev \True = K$.
Using R3, we get that $K \Rev \True \subseteq Cl(K \union \{ \True \}
) = K$. Since $K$ is consistent, by R4, $Cl(K \union \{ \True \})
\subseteq K \Rev \True$. Thus, $K \Rev \True = K$. By
Lemma~\ref{lem:Bel-in-SR}, we have that $\BEL(\Sys,\<\>) =
\BEL(\Sys,\<\True\>)$. Since
$\BEL(\Sys,\<\True\>) = K \Rev \True$, we conclude that
$\BEL(\Sys_{(\Rev,K)},\<\>) = K$.
\eprf

We next prove Theorem~\ref{rep2}.

\rethm{rep2}
Let $\Sys$ be a system in $\SR$.  Then there is an AGM revision
operator $\Rev_{\Sys}$ such that
$$
\BEL(\Sys,\<\>) \Rev_{\Sys} \phi = \BEL(\Sys, \<\phi\>)
$$
for all $\phi \in \Le$.
\erethm

\prf
Let $\Sys = (\R,\pi,\P)$ be a system in $\SR$.
It is easy to verify that $\PL_\Sys$ satisfies the conditions of
Lemma~\ref{lem:Rank-to-Rev} with $K = \BEL(\Sys,\<\>)$. This lemma
implies that
there is a revision operator $\Rev_\Sys$ such that $\psi \in K
\Rev_\Sys\phi$ if and only if $\PL_\Sys \sat \phi\Cond \psi$. Using
Lemma~\ref{lem:Bel-in-SR}, we have that $\psi \in
\BEL(\Sys,\<\phi\>)$ if and only if $\Pl_\Sys \sat \phi \Cond
\psi$. Thus, we have that $K \Rev_\Sys \phi = \BEL(\Sys,\<\phi\>)$
for all formulas $\phi$.
\eprf

\rethm{rep4}
Let $\Sys$ be a system in $\SR$ and $s_a = \<o_1, \ldots, o_m\>$
be a local state in $\Sys$.  Then there is an AGM revision operator
$\Rev_{\Sys,s_a}$ such that $$
\BEL(\Sys,s_a) \Rev_{\Sys,s_a} \phi = \BEL(\Sys, s_a \cdot \phi)
$$
for all formulas $\phi \in \Le$ such that $o_1 \land \ldots o_m
\land \phi$ is consistent.
\erethm

\prf
The structure of the proof is similar to that of
Theorem~\ref{rep2}. As in that proof, we construct a ranked
plausibility structure and use Lemma~\ref{lem:Rank-to-Rev} to find an
AGM revision operator. The main difference is that after observing
$\phi_1,\ldots,\phi_k$, some events are considered
impossible.  Lemma~\ref{lem:Rank-to-Rev}, however, requires that all
possible formulas are assigned a positive plausibility. We overcome
this problem by assigning a ``fictional'' positive plausibility to all
non-empty events that are ruled out by the previous observations.

We proceed as follows. Let  $d_0$
be a new plausibility value that is less plausible than all positive
plausibilities in $\Pl_a$; that is, if $\Pl_a(A) > \bot$,
then $\Pl_a(A) > d_0$. Let $\Sys = (\R,\pi,\P) \in \SR$; let $s_a =
\<o_1, \ldots, o_m\>$. We define $\PL = (\R,\Pl,\pi_{\sPL_\Sys})$,
where $\Pl$ is such that $\Pl(\intension{\phi}) = \max(
\Pl_a(\RE{\phi\land\band_{i=1}^{m}o_i}), d_0 )$  for all consistent
formulas $\phi$. This definition implies that if $\phi$ is consistent
with $\band_{i=1}^{m}o_i$, then $\Pl(\intension{\phi}_\sPL) =
\Pl_a(\RE{\phi_1\land\band_{i=1}^{m}o_i})$.

We now prove that if $\phi$ is consistent with $\band_{i=1}^{m}o_i$,
then $\PL \sat \phi\Cond\psi$ if and only if $\PL_\Sys \sat
(\phi\land\band_{i=1}^{m}o_i) \Cond \psi$.

For the ``if''
part, assume that $\PL_\Sys \sat
(\phi\land\band_{i=1}^{m}o_i) \Cond \psi$. Since $\phi$ is
consistent with $\band_{i=1}^{m}o_i$ it follows, from REV3, that
$\Pl_a(\RE{\phi\land(\band_{i=1}^{m}o_i})) > \bot$. Thus,
$\Pl_a(\RE{(\phi\land(\band_{i=1}^{m}o_i))\land\psi}) >
\Pl_a(\RE{(\phi\land(\band_{i=1}^{m}o_i))\land\neg\psi}) \ge
\bot$. Thus, $\phi\land\psi$ is consistent with $\band_{i=1}^{m}o_i$.
This implies that
$\Pl(\intension{\phi\land\psi}) =
\Pl_a(\RE{(\phi\land(\band_{i=1}^{m}o_i))\land\psi} >
\max(d_0,\Pl_a(\RE{(\phi\land(\band_{i=1}^{m}o_i))\land\neg\psi}) =
\Pl(\intension{\phi\land\neg\psi})$.
We conclude that $\PL \sat \phi\Cond\psi$.

For the ``only if'' part, assume that $\PL_\Sys \not\sat
(\phi\land(\band_{i=1}^{m}o_i)) \Cond \psi$. This implies that
$\Pl_a(\RE{(\phi\land(\band_{i=1}^{m}o_i))\land
\psi}) \not>
\Pl_a(\RE{(\phi\land(\band_{i=1}^{m}o_i))\land
\neg\psi})$. Since $\Pl_a$ is a ranking, it follows that
$\Pl_a(\RE{(\phi\land(\band_{i=1}^{m}o_i))\psi}) \le
\Pl_a(\RE{(\phi\land(\band_{i=1}^{m}o_i))\land\neg\psi})$.
Since $\bot < \Pl_a(\RE{\phi\land(\band_{i=1}^{m}o_i)}) =
\max(\Pl_a(\RE{(\phi\land(\band_{i=1}^{m}o_i))\land\psi}),
\Pl_a(\RE{(\phi\land(\band_{i=1}^{m}o_i))\land\neg\psi}))$, we have that
$\Pl_a(\RE{(\phi\land(\band_{i=1}^{m}o_i))\land\neg\psi}) > \bot$. We
conclude that
$\Pl(\intension{\phi\land\neg\psi}) \ge
\Pl(\intension{\phi\land\psi})$. Thus, $\PL \not \sat \phi \Cond
\psi$.

It is easy to verify that $\PL$ is ranked, and satisfies the
requirements of Lemma~\ref{lem:Rank-to-Rev}. Thus,
there exists a revision operator $\Rev_{\Sys,s_a}$ such that $\psi \in
K \Rev_{\Sys,s_a} \phi$ if and only if $\PL \sat \phi \Cond \psi$,
where $K = \{ \phi : \PL \sat \True \Cond \phi \}$. Moreover,
since for all $\phi$ consistent with
$\band_{i=1}^m{o_i}$ we have that$\PL \sat \phi\Cond\psi$ if and only if
$\PL_\Sys \sat (\phi \land (\band_{i=1}^m{o_i})) \Cond \psi$, then,
from Lemma~\ref{lem:Bel-in-SR}, it follows that $K = \BEL(\Sys,s_a)$
and that if $\phi$ is consistent with
$\band_{i=1}^{m}o_i$, then $\PL \sat \phi \Cond \psi$ if and only if
$\psi \in \BEL(\Sys,s_a\cdot \phi)$.
\eprf

\rethm{rep6}
Let $\Sys$ be a system in $\SR$ whose local states are $\E_{\Le}$.
There is a function
$\BEL_{\Sys}$
that maps epistemic states to belief states such that
\begin{itemize}\denselist
\item if $s_a$ is a local state of the agent in $\Sys$, then
$\BEL(\Sys,s_a) = \BEL_{\Sys}(s_a)$, and
\item $(\Rev,\BEL_{\Sys})$ satisfies R1$'$--R8$'$.
\end{itemize}
\erethm

\prf
As we said earlier,
roughly speaking, we define $\BEL_\Sys(s_a)
= \BEL(\Sys,s_a)$ when $s_a$ is a local state in $\Sys$. If $s_a$
is not in $\Sys$, then we set $\BEL_\Sys(s_a) = \BEL(\Sys,s')$,
where $s'$ is the longest consistent suffix of $s_a$. We now make
this definition precise, and show that the resulting $\BEL_\Sys$
satisfies R1$'$--R8$'$.

We proceed as follows.
We define a function $f(\cdot)$ that maps sequences of observations to
suffixes as follows:
$$
f(\<o_1,\ldots,o_m\>) = \left\{
\begin{array}{ll}
\<\> & \mbox{if $m = 0$,}\\
\<\False\> & \mbox{if $m > 0$ and $o_m$ is inconsistent,} \\
\<o_k, \ldots, o_m\> & \mbox{otherwise, with $k \le m$ the minimal
index}\\
& \mbox{s.~t.~$\not\vdash_\Le \neg(o_k\land \ldots \land o_m)$.}
\end{array}\right.
$$
Aside from the special case where $o_m$ is inconsistent, we simply
choose the longest suffix of $s_a$ that is still consistent.
We define $\BEL_\Sys(s_a) = \BEL(\Sys,f(s_a))$. Clearly, if $s_a$ is a
local state in $\Sys$, then $f(s_a) = s_a$, so $\BEL_\Sys(s_a) =
\BEL(\Sys,s_a)$.

We now have to show that $(\Rev,\BEL_\Sys)$ satisfies R1$'$--R8$'$. The proof
outline is as follows. Given a particular state $s_a$, we construct a
ranked plausibility structure that corresponds, in the sense of
Lemma~\ref{lem:Exists-Ranking}, to belief change from $s_a$. We then use
Lemma~\ref{lem:Rank-to-Rev} to show that belief changes from $s_a$ satisfies
the AGM postulates, \ie R1--R8. Since this is true from any $s_a$, we
get that $\BEL_\Sys$ satisfies R1$'$--R8$'$.

Let $s_a = \<o_1,\ldots,o_m\>$. We define a ranked plausibility space
that has the following structure. The most plausible events are the
ones consistent with $o_1\land\ldots\land o_m$. They are ordered
according to the prior ranking conditioned on
$o_1\land\ldots\land o_m$. The next tier of events are
those that are inconsistent with $o_1\land\ldots\land o_m$ but are
consistent $o_2\land\ldots\land o_m$.
Again, these are ordered according to the prior ranking conditioned on
$o_2\land\ldots\land o_m$. We continue this way; the last tier
consists of all events that are inconsistent with $o_m$.

Formally,
let $\PL = (\R,\Pl,\pi_{\sPL_\Sys})$, where $\Pl$ is such that
$\Pl(\intension{\phi}) \ge \Pl(\intension{\psi})$ if
$\Pl_a(\RE{\phi\land (\band_{i=k}^{m}o_i})) \ge
\Pl_a(\RE{\psi\land (\band_{i=k}^{m}o_i}))$ where $k \le m+1$  is the greatest
integer such that for all $j < k$, $\phi$ and $\psi$ are
both inconsistent with $\band_{i=j}^{m}o_i$.  It is easy to see that
$\PL$ is ranked, and that if $\phi$ is consistent, then
$\Pl(\intension{\phi}) > \bot$.

Let $\phi \in \Le$. We now show that $\PL \sat \phi \Cond\psi$ if and
only if $\psi \in \BEL_\Sys(s_a \cdot \phi)$.
If $\phi$ is inconsistent, then $\PL \sat
\phi \Cond \psi$ for all $\psi$. Moreover, since $\phi$ is
inconsistent, $f(s_a \cdot \phi) = \< \False \>$, and thus
$\BEL_\Sys(s_a \cdot \phi) = \Le$. We conclude that $\phi \Cond \psi$
if and only if $\psi \in \BEL_\Sys(s_a \cdot
\phi)$.
If $\phi$ is consistent, then let $k \le m+1$ be the
greatest integer such that for all $j < k$, $\phi$ is inconsistent with
$\band_{i=j}^{m}o_i$.  It is easy to verify that $f(s_a \cdot \phi) =
\< o_k,\ldots,o_m,\phi\>$. {From} Lemma~\ref{lem:Bel-in-SR}, it follows
that $\psi \in \BEL_\Sys(s_a \cdot \phi) =
\BEL(\Sys, \< o_k,\ldots,o_m,\phi\>)$ if and only if
$\Pl_a(\RE{(\phi\land(\band_{i=k}^{m}o_i))\land\psi}) >
\Pl_a(\RE{(\phi\land(\band_{i=k}^{m}o_i))\land\neg\psi})$. We now show
that this is the case if and only if $\PL \sat \phi\Cond\psi$.
Suppose that $\PL_a(\RE{(\phi\land\land(\band_{i=k}^{m}o_i))\land
\psi}) > \PL_a(\RE{(\phi \land(\band_{i=k}^{m}o_i))
\land\neg\psi})$. Then, clearly,
$\Pl_a(\RE{(\phi\land(\band_{i=k}^{m}o_i))\land \psi}) > \bot$, and thus
$\phi\land\psi$ is consistent with
$o_k, \ldots, o_m$. Since both $\phi\land\psi$ and $\phi\land\neg\psi$
are inconsistent with $o_j, \ldots, o_m$ for all $j < k$, we have that
$\Pl(\intension{\phi\land\psi}) >
\Pl(\intension{\phi\land\neg\psi})$.
On other hand, if
$\Pl_a(\RE{(\phi\land(\band_{i=k}^{m}o_i))\land\psi}) \not>
\Pl_a(\RE{(\phi\land(\band_{i=k}^{m}o_i))\land
\neg\psi})$, then since $\Pl_a$ is a ranking
$\PL_a(\RE{(\phi\land(\band_{i=k}^{m}o_i))\land \psi}) \le
\PL_a(\RE{(\phi \land(\band_{i=k}^{m}o_i))\land\neg\psi})$. Moreover,
since $\phi$ is consistent with $o_k\land \ldots \land o_m$, we have that
$\Pl_a(\RE{\phi\land(\band_{i=k}^{m}o_i)}) > \bot$. This implies that
$\Pl_a(\RE{(\phi\land(\band_{i=k}^{m}o_i)) \land \neg\psi}) > \bot$ and thus
$\Pl(\intension{\phi\land\psi}) \le
\Pl(\intension{\phi\land\neg\psi})$.
We conclude that $\PL \sat \phi
\Cond \psi$ if and only if $\psi \in \BEL_\Sys(s_a \cdot \phi)$.

By Lemma~\ref{lem:Rank-to-Rev}, there is a
revision operator $\Rev_{s_a}$ that satisfies R1--R8 such that $\psi
\in K \Rev \phi$ if and only if $\PL \sat \phi \Cond \psi$. It is not
hard to check that this
implies that the change from $\BEL_\Sys(s_a)$ to
$\BEL_\Sys(s_a\cdot \phi)$ satisfies R1$'$--R8$'$.
\eprf

\repro{pro:R9'}
Let $\Sys$ be a system in $\SR$ whose local states are $\E_{\Le}$.
There is a function
$\BEL_{\Sys}$
that maps epistemic states to belief states such that
\begin{itemize}\denselist
\item if $s_a$ is a local state of the agent in $\Sys$, then
$\BEL(\Sys,s_a) = \BEL_{\Sys}(s_a)$, and
\item $(\Rev,\BEL_{\Sys})$ satisfies R1$'$--R9$'$.
\end{itemize}
\erepro

\prf
As we said in the main text,
we show that the function $\BEL_\Sys$ defined in the proof of
Theorem~\ref{rep6} satisfies R9$'$.
Let $s_a = \<o_1,\ldots,o_m\>$, and let $\phi,
\psi \in \Le$ be formulas such that  $\not\vdash_\Le
\neg(\phi\land\psi)$. Since
$\phi$ is consistent with $\psi$, we get that $f(s_a \cdot \phi \cdot
\psi) = \<o_k, \ldots, o_m, \phi,\psi\>$, where $k\le m$ is the least
integer such that $\phi\land\psi$ is consistent with
$o_k,\ldots,o_m$. For the same reason, we get that $f(s_a \cdot
\phi\land\psi) = \<o_k, \ldots, o_m, \phi\land\psi\>$.  Using
Lemma~\ref{lem:Bel-in-SR} we immediately get that $\BEL(\Sys,\<o_k,
\ldots, o_m, \phi,\psi\>) = \BEL(\Sys,\<o_k,
\ldots, o_m, \phi \land\psi\>)$. Thus, we conclude that $\BEL_\Sys(s_a
\cdot \phi\cdot \psi) = \BEL_\Sys(s_a\cdot \phi\land\psi)$.
\eprf

\rethm{rep5}
Given a function $\BEL_{\Le}$ mapping epistemic states in $\E_{\Le}$ to
belief sets over $\Le$ such that $\BEL_{\Le}(\<\>)$ is consistent and
$(\BEL_{\Le},\Rev)$ satisfies R1$'$--R9$'$, there is a system
$\Sys \in \SR$ whose local states are in $\E_{\Le}$ such that
$\BEL_{\Le}(s_a) = \BEL(\Sys,s_a)$ for each local state $s_a$ in
$\Sys$.
\erethm

\prf
We show that $\BEL(\Sys,s_a) = \BEL_{\Le}(s_a)$ for local states $s_a$
in $\Sys$,
where $\Sys$ is the system guaranteed to exist by
Theorem~\ref{rep1} such that $\BEL(\Sys,\<\>) = \BEL_{\Le}(\<\>)$ and
$\BEL(\Sys,\<\phi\>) = \BEL_{\Le}(\<\phi\>)$ for all $\phi\in\Le$.
We prove this by induction on the
length $m$ of $s_a$. For $m \le 1$, this is true by our choice of
$\Sys$. For the induction case, let $s_a = \< o_1, \ldots, o_m \>$
be a local state in $\Sys$. Thus,, $o_1\land\ldots\land o_m$
is consistent. {From} R9$'$, it follows that
$\BEL_{\Le}(\<o_1,\ldots,o_m\>) =
\BEL_{\Le}(\<o_1,\ldots,o_{m-2},o_{m-1}\land o_m\>)$. Using the
induction hypothesis, we have that
$\BEL_{\Le}(\<o_1,\ldots,o_{m-2},o_{m-1}\land o_m\>) = \BEL(\Sys,
\<o_1,\ldots,o_{m-2},o_{m-1}\land o_m\>)$. Using
Lemma~\ref{lem:Bel-in-SR}, we get that
$\BEL(\Sys, \<o_1,\ldots,o_{m-2},o_{m-1}\land o_m\>) =
\BEL(\Sys, \<o_1,\ldots, o_m\>)$. Thus, we conclude that
$\BEL_{\Le}(\<o_1,\ldots, o_m\>) = \BEL(\Sys, \<o_1,\ldots, o_m\>)$.
\eprf

\subsection{Proofs for Section~\protect\ref{sec:update}}
\label{prf:update}

In this section we prove Theorem~\ref{thm:update-rep}.  We now show
that any system in $\SU$ corresponds to an update structure. Suppose
that $\Sys = (\R,\pi,\P) \in \SU$ is such that the set of environment
states is $\S_e$ and the prior of BCS5 is consistent with distance
function $d$. Define an update structure $U_\Sys = (S_e,\pi_e,d)$,
where for $p \in \Phis$, $\pi_e(s_e)(p) = \pi((s_e,s_a))(p)$ for some
choice of $s_a$. By BCS1, the choice of $s_a$ does not matter. It is
easy to see that UPD1 ensures that $S_e$ and $\pi_e$ satisfy the
requirements of the definition of update structures.  We want to show
that belief change in $\Sys$ corresponds to belief change in $U_\Sys$
in the sense of Theorem~\ref{thm:KM}. Since Theorem~\ref{thm:KM}
states that any belief change operation defined by an update structure
satisfies U1--U8, this will suffice to prove the ``if'' direction of
Theorem~\ref{thm:update-rep}. To prove the ``only if'' direction of
Theorem~\ref{thm:update-rep}, we show that that for any update
structure $U$, there is a system $\Sys \in \SU$ such that $U_\Sys =
U$.

We start with preliminary definitions and lemmas for the ``if''
direction of Theorem~\ref{thm:update-rep}.  Let $s_a =
\<o_1,\ldots,o_m\>$. We define $\States(\Sys,s_a) = \{ s \in
\S_e : s \sat \xi \mbox{\ for all\ } \xi \in \BEL(\Sys,s_a) \}$. Clearly, if
$\phi$ is such that $\BEL(\Sys,s_a) = Cl(\phi)$, then
$\States(\Sys,s_a) = \intension{\phi}_{U_\Sys}$. To show that belief
change in $\Sys$ corresponds to belief change in $U_\Sys$ we have to
show that
$$
\States(\Sys,s_a \cdot \psi) =
{\min}_{U_\Sys}(\States(\Sys,s_a),\intension{\psi}_{U_\Sys}).
$$
This is proved in Lemma~\ref{lem:Upd-States-Change}. To prove this
lemma, we need some preliminary lemmas.

\lem\label{lem:Bel-in-SU}
Let $\Sys \in \SU$, and let $s_a = \<o_1,\ldots,o_m\>$. Then $\phi \in
\BEL(\Sys,s_a)$ if and only if $(\Sys,r,0) \sat (\Next o_1\land\ldots\land
\Next^m o_m) \Cond \Next^m \phi$ for some run $r$ in $\R$.
\elem
\prf
The proof of this lemma is analogous to the proof of
Lemma~\ref{lem:Bel-in-SR}, using UPD3 and UPD4 instead of REV3 and
REV4.  We do not repeat the argument here.
\eprf

We now provide an alternative characterization of
$\States(\Sys,s_a)$ in terms of the agent's prior on run-prefixes.

\lem\label{lem:Upd-States-Def}
Let $\Sys \in \SU$ and let $s_a = \<o_1,\ldots,o_m\>$. Then $s_m \in
\States(\Sys,s_a)$ if and only if there is a sequence of states
$\RP{s_0,\ldots,s_m} \subseteq \RE{\True,o_1,\ldots,o_m}$ such that
$\Pl_a(\RP{s_0,\ldots,s_m}) \not< \Pl_a(\RE{\True,o_1,\ldots,o_m} -
\RP{s_0,\ldots,s_m})$.
\elem

\prf
{\sloppy
For the ``if'' direction,
assume that there is a sequence $s_0,\ldots,s_{m}$ such that
$\RP{s_0,\ldots,s_m} \subseteq \RE{\True,o_1,\ldots,o_m}$, and
$\Pl_a(\RP{s_0,\ldots,s_m}) \not< \Pl_a(\RE{\True,o_1,\ldots,o_m} -
\RP{s_0,\ldots,s_m})$. By way of contradiction, assume that $s_m
\not\in \States(\Sys,m)$. Thus, there is a formula $\xi \in
\BEL(\Sys,m)$ such that $s_m \sat \neg\xi$. {From}
Lemma~\ref{lem:Bel-in-SU} it follows that since $\xi \in
\BEL(\Sys,s_a)$, $(\Sys,r,0) \sat (\Next o_1\land\ldots\land
\Next^m o_m) \Cond \Next^m \xi$ for some run $r$ in $\R$. {From} the
definition of conditioning it follows that
$\Pl_a(\RE{\True,o_1,\ldots,o_{m-1},o_m \land \xi}) >
\Pl_a(\RE{\True,o_1,\ldots,o_{m-1},o_m \land \neg\xi})$.
Since $s_m \sat \neg\xi$, we get that $\RP{s_0,\ldots,s_m} \subseteq
\RE{\True,o_1,\ldots,o_{m-1},o_m \land \neg\xi}$ and that
$\RE{\True,o_1,\ldots,o_{m-1},o_m \land\xi} \subseteq
\RE{\True,o_1,\ldots,o_m} - \RP{s_0,\ldots,s_m}$. {From} A1, it follows
that $\Pl_a(\RP{s_0,\ldots,s_m}) < \Pl_a(\RE{\True,o_1,\ldots,o_m} -
\RP{s_0,\ldots,s_m})$, which contradicts our starting assumption. We conclude
that $s_m \in \States(\Sys,s_a)$.

For the ``only if'' direction, assume that $s_m \in
\States(\Sys,a)$. Since $\S_e$ is finite
and $\pi_e$ assigns a different truth assignment to each state in
$\S_e$, there is a formula $\xi \in \Le$ that characterizes $s_m$;
that is, $s \sat \xi$ if and only if $s = s_m$. Since $s_m \in
\States(\Sys,s_a)$, we have that $\neg\xi \not\in
\BEL(\Sys,s_a)$. Using Lemma~\ref{lem:Bel-in-SU}, we get that
$(\Sys,r,0) \not\sat (\Next o_1\land\ldots\land \Next^m o_m) \Cond
\Next^m\neg\xi$ for all runs $r \in \R$. By BCS5, this is true if and
only if $\Pl_a(\RE{\True,o_1,\ldots,o_m}) > \bot$ and
$\Pl_a(\RE{\True,o_1,\ldots,o_{m-1},o_m\land\xi}) \not<
\Pl_a(\RE{\True,o_1,\ldots,o_{m-1},o_m\land\neg\xi})$. By UPD2, there
is a sequence $\RP{s_0,\ldots,s_m} \subseteq
\RE{\True,o_1,\ldots,o_{m-1},o_m\land\xi}$ such that
$\Pl_a(\RP{s_0,\ldots,s_m}) \not< \Pl_a(\RP{s'_0,\ldots,s'_m})$ for
all $\RP{s'_0,\ldots,s'_m} \subseteq
\RE{\True,o_1,\ldots,o_{m-1},o_m\land\neg\xi}$. Moreover,  without
loss of generality, we can assume that $\Pl_a(\RP{s_0,\ldots,s_m})
\not< \Pl_a(\RP{s'_0,\ldots,s'_m})$ for all $\RP{s'_0,\ldots,s'_m}
\subseteq \RE{\True,o_1,\ldots,o_{m-1},o_m\land\xi}$, since there
are  only finitely many such sequences. Thus, by UPD2,
$\Pl_a(\RP{s_0,\ldots,s_m}) \not<  \Pl_a(\RE{\True,o_1,\ldots,o_m} -
\RP{s_0,\ldots,s_m})$.
}
\eprf

We can now prove that belief change in $\Sys$ corresponds to belief
change in $U_\Sys$.

\lem\label{lem:Upd-States-Change}
Let $\Sys = (\R,\pi,\P) \in \SU$
Then
$$\States(\Sys,s_a \cdot \psi) =
{\min}_{U_\Sys}(\States(\Sys,s_a),\intension{\psi}_{U_\Sys}) $$
for all local states $s_a$ and formulas $\psi \in \Le$.
\elem

\prf
{\sloppy
Let $\Pl_a$ be the prior in $\Sys$; assume that $\Pl_a$ consistent with
a distance function $d$.  Let $s_a = \<o_1,\ldots,o_m\>$.

To show that
$\min_{U_\Sys}(\States(\Sys,s_a),\intension{\psi}_{U_\Sys}) \subseteq
\States(\Sys,s_a \cdot \psi)$,
suppose that $s \in
\min_{U_\Sys}(\States(\Sys,s_a),\intension{\psi}_{U_\Sys})$. Thus,
there is a state $s_m \in \States(\Sys,s_a)$ such that $d(s_m,s')
\not < d(s_m,s)$ for all states $s'$ that satisfy $\psi$.
We want to show that $s \in \States(\Sys, s_a\cdot\psi)$.
{From} Lemma~\ref{lem:Upd-States-Def}, it follows that, since $s_m \in
\States(\Sys,s_a)$, there is a sequence
$s_0,\ldots,s_{m-1}$ such that $\RP{s_0,\ldots,s_m} \in
\RE{\True,o_1,\ldots,o_{m-1},o_{m}}$ and $\Pl_a(\RP{s_0,\ldots,s_m})
\not < \Pl_a(\RE{\True,o_1,\ldots,o_{m-1},o_{m}} -
\RP{s_0,\ldots,s_m})$.
We now show that $\Pl_a(\RP{s_0,\ldots,s_m,s}) \not <
\Pl_a(\RE{\True,o_1,\ldots,o_m,\psi} - \RP{s_0,\ldots,s_m,s})$.
By Lemma~\ref{lem:Upd-States-Def}, this suffices to show that $s \in
\States(\Sys,s_a \cdot \psi)$. Suppose that $\RP{s'_0,\ldots,s'_{m+1}}
\subseteq \RE{\True,o_1,\ldots,o_m,\psi} - \RP{s_0,\ldots,s_m,s}$.
If $\RP{s_0,\ldots,s_m} = \RP{s'_0,\ldots,s'_m}$, then we have that
$d(s'_m,s'_{m+1}) \not< d(s_m,s)$.  Since $\Pl_a$ is consistent with
$d$, it follows that $\Pl_a(\RP{s_0,\ldots,s_m,s}) \not <
\Pl_a(\RP{s'_0,\ldots,s'_m,s'_{m+1}})$.
If $\RP{s_0,\ldots,s_m} \neq
\RP{s'_0,\ldots,s'_m}$, then, since
$\Pl_a(\RP{s_0,\ldots,s_m}) \not<
\Pl_a(\RP{s'_0,\ldots,s'_m})$ and $\Pl_a$ is consistent with $d$, we
have that $\Pl_a(\RP{s_0,\ldots,s_m,s}) \not <
\Pl_a(\RP{s'_0,\ldots,s'_m,s'_{m+1}})$.

Since $\Pl_a(\RP{s_0,\ldots,s_m,s}) \not <
\Pl_a(\RP{s'_0,\ldots,s'_m,s'_{m+1}})$ for all
$\RP{s'_0,\ldots,s'_{m+1}} \subseteq \RE{\True,o_1,\ldots,o_m,\psi} -
\RP{s_0,\ldots,s_m,s}$ and $\Pl_a$ is prefix-defined, we have that
$\Pl_a(\RP{s_0,\ldots,s_m,s}) \not<
\Pl_a(\RE{\True,o_1,\ldots,o_m,\psi} -
\RP{s_0,\ldots,s_m,s}$. By Lemma~\ref{lem:Upd-States-Def}, $s \in
\States(\Sys,s_a \cdot \psi)$, as desired.

To show that $\States(\Sys,s_a \cdot \psi) \subseteq
\min_{U_\Sys}(\States(\Sys,s_a),\intension{\psi}_{U_\Sys})$,
suppose that $s \in \States(\Sys,s_a \cdot \psi)$.
By Lemma~\ref{lem:Upd-States-Def},
there is a sequence $s_0, \ldots, s_m$ such that
$\RP{s_0,\ldots,s_m,s}) \subseteq \RE{\True,o_1,\ldots,o_m,\psi}$ and
$\Pl_a(\RP{s_0,\ldots,s_m,s}) \not <
\Pl_a(\RE{\True,o_1,\ldots,o_m,\psi} - \RP{s_0,\ldots,s_m,s})$.
We want to show that $s_m \in \States(\Sys,s_a)$ and that $d(s_m,s')
\not< d(s_m,s)$ for all $s'$ that satisfy $\psi$. This suffices to
prove that $s \in
\min_{U_\Sys}(\States(\Sys,s_a),\intension{\psi}_{U_\Sys})$.

To show that $s_m \in \States(\Sys,s_a)$,
by Lemma~\ref{lem:Upd-States-Def}, it suffices to show that
$\Pl_a(\RP{s_0,\ldots,s_m}) \not < \Pl_a(\RE{\True,o_1,\ldots,o_m} -
\RP{s_0,\ldots,s_m})$.
Let $s'_0,\dots,s'_m$ be a sequence such that
$\RP{s'_0,\ldots,s'_m} \subseteq \RE{\True,o_1,\ldots,o_m}$. By
definition, $\RP{s'_0,\ldots,s'_m,s} \subseteq
\RE{\True,o_1,\ldots,o_m,\psi}$. Thus, from our choice of
$s_0,\ldots,s_m$, it follows that
$\Pl_a(\RP{s_0,\ldots,s_m,s}) \not<
\Pl_a(\RP{s'_0,\ldots,s'_m,s})$. Since $\Pl_a$ is consistent with $d$, it
follows that $\Pl_a(\RP{s_0,\ldots,s_m}) \not<
\Pl_a(\RP{s'_0,\ldots,s'_m})$. Thus, by
Lemma~\ref{lem:Upd-States-Def}, $s_m \in \States(\Sys,s_a)$. To see
that $d(s_m,s') \not< d(s_m,s)$ for all $s'$ that satisfy
$\psi$, let $s' \neq s$ be such that $s' \sat \psi$. Thus,
$\RP{s_0,\ldots,s_m, s'} \subseteq
\RP{\True,o_1,\ldots,o_m,\psi}$. {From} our choice   of
$s_0,\ldots,s_m$, it follows that
$\Pl_a(\RP{s_0,\ldots,s_m,s}) \not<
\Pl_a(\RP{s_0,\ldots,s_m,s'})$. Since $\Pl_a$ is consistent with $d$, it
follows that $d(s_m,s') \not< d(s_m,s)$. We conclude that $s \in
\min_{U_\Sys}(\States(\Sys,s_a),\intension{\psi}_{U_\Sys})$.
}
\eprf

We now have the tools to prove the ``if'' direction of
Theorem~\ref{thm:update-rep}.
\lem\label{lem:Update-if}
If $\Sys = (\R,\pi,\P) \in \SU$, then there is a belief change
operator $\Upd$ that satisfies U1--U8 such that
$$\BEL(\Sys,s_a) \Upd \psi = \BEL(\Sys, s_a \cdot \psi)$$
for all local states $s_a$ and formulas $\psi \in \Le$.
\elem

\prf
Let $\Sys \in \SU$. Using the arguments we presented above, it easy to
check that $U_\Sys$ is an update structure.  By Theorem~\ref{thm:KM},
there is a belief change operator $\Upd$ that satisfies U1--U8 such
that $\intension{\phi \Upd \psi}_{U_\Sys} =
{\min}_{U_\Sys}(\intension{\phi}_{U_\Sys},\intension{\psi}_{U_\Sys})$
for all $\phi,\psi \in \Le$. {From} Lemma~\ref{lem:Upd-States-Change},
it follows that $\BEL(\Sys,s_a) \Upd \psi = \BEL(\Sys, s_a \cdot
\psi).$
\eprf

We now prove the ``only if'' direction of Theorem~\ref{thm:update-rep}.
Suppose that $\Upd$ is a belief change operator that satisfies
U1--U8. According to Theorem~\ref{thm:KM}, there is an update
structure $U_\Upd$ that corresponds to $\Upd$. Thus, it suffices to
show that there is a system $\Sys$ such that $U_\Sys = U_\Upd$.

\lem\label{lem:Update-Exists}
Let $U = (W,d,\pi_U)$ be an update structure. Then there is a system
$\Sys \in \SU$ such that $U_\Sys = U$.
\elem

\prf
Given the sequences $w_0,w_1,\ldots \in W$ and $o_1,o_2,\ldots \in \Le$, let
$r^{w_0,w_1,\ldots;o_1,o_2,\ldots}$ be the run defined so that
$r^{w_0,w_1,\ldots;o_1,o_2,\ldots}_e(m) = w_m$ and
$r^{w_0,w_1,\ldots;o_1,o_2,\ldots}_a(m) = \<o_1,\ldots,o_m\>$.
Let $\R = \{ r^{w_0,w_1,\ldots;o_1,o_2,\ldots} : \pi_U(w_m)(o_m) = $
{\bf true} for all $m \}$. Define $\pi$
such that $\pi(r,m)(p) = \pi_U(r_e(m))(p)$ for $p \in
\Phis$ and $\pi(r,m)(\learn(\phi)) =$ {\bf true} if
$\obs{r,m} = \phi$ for $\phi \in \Le$.

It is clear that $(\R,\pi)$ satisfies BCS1--BCS4 and UPD1. Thus, all
that remains to show is that there is a prior plausibility measure
$\Pl_a$ that satisfies UPD2--UPD4. This will ensure that
$(\R,\pi,\Plass) \in \SU$.

We proceed as follows. We define a {\em preferential space\/}
$(\R,\prec)$ where $r \prec r'$ if and only if there is some $m$
such that $r_e(k) = r'_e(k)$ for all $0 \le k \le m$, $r_e(m+1) \neq
r'_e(m+1)$, and $d(r_e(m),r_e(m+1)) < d(r'_e(m),r'_e(m+1))$. Recall
that $r \prec r'$ denotes that $r$ is preferred over $r'$. Thus, this
ordering is consistent with the comparison of events of the form
$\RP{s_0,\ldots,s_n}$ according to UPD2.

Using the construction of Proposition~\ref{pro:prec}, there is a
plausibility space $(R,\Pl_a)$ such that $\Pl_a(A) \ge \Pl_a(B)$ if and
only if for all $r \in B - A$, there is a run $r' \in A$ such that $r'
\prec r$ and there is no $r'' \in B - A$ such that $r'' \prec r'$. By
\cite[Theorem~5.5]{FrH5Full}, $\Pl_a$ is a qualitative
plausibility measure. We now show that it satisfies UPD2--UPD4.

We start with UPD2. To show that $\Pl_A$ is consistent with $d$, we
need to show that $\Pl_a(\RP{s_0,\ldots,s_n}) < \Pl_a(\RP{s_0',
\ldots, s_n'})$ if and only if there is some $m < n$ such that $s_k =
s'_k$ for all $0 \le k \le m$, and
$d(s_{m},s_{m+1}) > d(s'_{m},s'_{m+1})$.   Suppose that
$\Pl_a(\RP{s_0,\ldots,s_n}) < \Pl_a(\RP{s'_0,\ldots,s'_n})$.
Let $r$ be some run in $\RP{s_0,\ldots,'_n}$. Without loss
of generality we can assume that $r_e(m) = r_e(n)$ for all $m >
n$. Since $\Pl_a(\RP{s_0,\ldots,s_n}) < \Pl_a(\RP{s'_0,\ldots,s'_n})$,
there is a run $r' \in
\RP{s'_0,\ldots,s'_n}$ such that $r' \prec r$. By definition,
this implies that there is an $m$ such that $r_e(k) = r'_e(k)$ for all
$0 \le k \le m$, and $d(r'_e(m),r'_e(m+1)) <
d(r_e(m),r_e(m+1))$. We claim that $m < n$. For if $m \ge n$, then
$r_e(m+1) = r_e(m)$ by construction, so $d(r_e(m),r_e(m+1)) =
d(r_e(m),r_e(m)) \le d(r'_e(m),r'_e(m+1))$ and $r' \not\prec r$, a
contradiction.
Thus, $s_k = s'_k$ for all $0 \le k \le m$, $d(s'_m,s'_{m+1}) <
d(s_{m},s_{m+1})$.

For the converse, suppose that there is an $m < n$ such that $s_k =
s'_k$ for all $0 \le k \le m$, and $d(s'_m,s'_{m+1}) <
d(s_{m},s_{m+1})$. Let $r'$ be the run where $r'_e(k) = s'_k$ for $k
\le n$, $r'_e(k) = s'_n$ for $k \ge n$, and $\obs{r',k} = \True$ for all
$k$. It follows $r' \prec r$ for all runs $r' \in
\RP{s_0,\ldots,s_n}$. Thus, $\Pl_a(\RP{s_0,\ldots,s_n}) <
\Pl_a(\RP{s'_0,\ldots,s'_n})$.

To show that $\Pl_a$ is prefix-defined, we must show that
$\Pl_a(\RE{\phi_0,\ldots,\phi_n}) \ge
\Pl_a(\RE{\psi_0,\ldots,\psi_n})$ if and only if for all
$\RP{s_0,\ldots,s_n} \subseteq \RE{\psi_0,\ldots,\psi_n} -
\RE{\phi_0,\ldots,\phi_n}$, there is some $\RP{s'_0,\ldots,s'_n}
\subseteq \RE{\phi_0,\ldots,\phi_n}$ such that
$\Pl_a(\RP{s'_0,\ldots,s'_n}) > \Pl_a(\RP{s_0,\ldots,s_n})$.
Suppose that $\Pl_a(\RE{\phi_0,\ldots,\phi_n}) \ge
\Pl_a(\RE{\psi_0,\ldots,\psi_n})$. Let $\RP{s_0,\ldots,s_n}
\subseteq \RE{\psi_0,\ldots,\psi_n} - \RE{\phi_0,\ldots,\phi_n}$. Let $r \in
\RP{s_0,\ldots,s_n}$ be a run such that $r_e(m) = r_e(n)$ for all $m \ge
n$. Since $\Pl_a(\RE{\phi_0,\ldots,\phi_n}) \ge
\Pl_a(\RE{\psi_0,\ldots,\psi_n})$ there is a run $r' \in
\RE{\phi_0,\ldots,\phi_n}$ such that $r' \prec r$. This implies that
there is an $m$ such that $r_e(k) = r'_e(k)$ for all $0 \le k \le m$,
and $d(r'_e(m),r'_e(m+1)) <
d(r_e(m),r_e(m+1))$. As before, we have that $m < n$, and thus
$\Pl_a(\RP{r'_e(0), \ldots, r'_e(n)}) >
\Pl_a(\RP{s_0,\ldots,s_n})$. Since $r' \in \RE{\phi_0,\ldots,\phi_n}$,
we also have that $\RP{r'_e(0), \ldots, r'_e(n)} \subseteq
\RE{\phi_0,\ldots,\phi_n}$, as desired.

For the converse, assume that  for all $\RP{s_0,\ldots,s_n} \subseteq
\RE{\psi_0,\ldots,\psi_n} - \RE{\phi_0,\ldots,\phi_n}$
there is  some $\RP{s'_0,\ldots,s'_n} \subseteq
\RE{\phi_0,\ldots,\phi_n}$ such that $\Pl_a(\RP{s'_0,\ldots,s'_n}) >
\Pl_a(\RP{s_0,\ldots,s_n})$. This implies that
$\Pl_a(\RE{\phi_0,\ldots,\phi_n}) >  \Pl_a(\RP{s_0,\ldots,s_n})$ for all
for all $\RP{s_0,\ldots,s_n} \subseteq
\RE{\psi_0,\ldots,\psi_n} - \RE{\phi_0,\ldots,\phi_n}$. Since there are
only finitely many sequences of states of length $m$, we can apply
A2, and conclude that $\Pl_a(\RE{\phi_0,\ldots,\phi_n}) >
\Pl_a(\RE{\psi_0,\ldots,\psi_n} - \RE{\phi_0,\ldots,\phi_n})$. Thus, $\Pl_a(\RE{\phi_0,\ldots,\phi_n}) \ge \Pl_a((\RE{\psi_0,\ldots,\psi_n}))$.

For UPD3, recall that the construction of
Proposition~\ref{pro:prec} is such that $\Pl_a(R) > \bot$ for
all non-empty $R \subseteq
\R$. Since, by our construction, the set $\RE{\phi_0,\ldots, \phi_n}$
is non-empty for all sequences $\phi_0,\ldots,\phi_n$ of consistent
formulas, UPD3 must hold.

{\sloppy
Finally, we consider UPD4. We have to show that
$\Pl_a(\RE{\phi_0,\ldots, \phi_{n+1};  o_1,\ldots,o_n})$ $\ge$ \linebreak
$\Pl_a(\RE{\psi_0,\ldots, \psi_{n+1};  o_1,\ldots,o_n})$ if and only if
$\Pl_a(\RE{\phi_0, \phi_1 \land o_1, \ldots, \phi_n \land
o_n,\phi_{n+1}}) \ge \Pl_a(\RE{\psi_0, \psi_1 \land o_1, \ldots, \psi_n \land
o_n,\psi_{n+1}})$.
By construction,
$\RE{\phi_0,\ldots, \phi_{n+1};  o_1,\ldots,o_n} \subseteq
\RE{\phi_0, \phi_1 \land o_1, \ldots, \phi_n \land
o_n,\phi_{n+1}}$. On the other hand, for each run $r \in \RE{\phi_0, \phi_1 \land o_1, \ldots, \phi_n \land
o_n,\phi_{n+1}}$ there is a run $r' \in \RE{\phi_0,\ldots, \phi_{n+1};
o_1,\ldots,o_n}$ such that $r'_e(m) = r_e(m)$ for
all $m$, and $\obs{r,m} = o_m$ for $1 \le m \le n$. Since the
preference ordering on runs is a function only of the environment
states, it is clear that $r$ and $r'$ are compared in the same manner;
that is for all $r''$, $r'' \prec r$ if and only if $r'' \prec r'$,
and $r \prec r''$ if and only if $r' \prec r''$. Thus, we conclude
that for the purposes of the preference ordering, both $\RE{\phi_0,
\phi_1 \land o_1, \ldots, \phi_n \land o_n,\phi_{n+1}}$ and
$\RE{\phi_0,\ldots, \phi_{n+1};o_1,\ldots,o_n}$ are compared in the
same manner to other sets. It easy to see that this suffices to show
that $\Pl_a$ satisfies UPD4.
}
\eprf

Finally, we can prove Theorem~\ref{thm:update-rep}.

\rethm{thm:update-rep}
A belief change operator $\Upd$ satisfies U1--U8 if and only if there
is a system $\Sys \in \SU$ such that
$$\BEL(\Sys,s_a) \Upd \psi = \BEL(\Sys, s_a \cdot \psi)$$
for all epistemic states $s_a$ and formulas $\psi \in \Le$.
\erethm

\prf
The ``if'' direction follows from Lemma~\ref{lem:Update-if}. For the
``only if'' direction, assume that $\Upd$ satisfies U1--U8. By
Theorem~\ref{thm:KM}, there is an update structure $U_\Upd$ such that
$\intension{\phi \Upd \psi}_{U_\Sys} =
{\min}_{U_\Sys}(\intension{\phi}_{U_\Sys},\intension{\psi}_{U_\Sys})$
for all $\phi,\psi \in \Le$. By Lemma~\ref{lem:Update-Exists}, there is a
system $\Sys \in \SU$ such that $U_\Sys = U_\Upd$. {From}
Lemma~\ref{lem:Upd-States-Change}, it follows that
$\BEL(\Sys,s_a) \Upd \psi = \BEL(\Sys, s_a \cdot \psi)$
for all local states $s_a$ and formulas $\psi \in \Le$.
\eprf

\end{document}